%% file: neurips_2026.tex
\documentclass{article}

\PassOptionsToPackage{numbers, compress}{natbib}
\usepackage[preprint]{neurips_2026}

\usepackage[utf8]{inputenc} % allow utf-8 input
\usepackage[T1]{fontenc}    % use 8-bit T1 fonts
\usepackage{xcolor}         % colors

\usepackage{hyperref}       % hyperlinks
\definecolor{mylinkcolor}{RGB}{0,114,178}  % 차분한 파랑
\usepackage[table]{xcolor}
\usepackage{colortbl}
\usepackage{url}            % simple URL typesetting
\usepackage{booktabs}       % professional-quality tables
\usepackage{amsfonts}       % blackboard math symbols
\usepackage{nicefrac}       % compact symbols for 1/2, etc.
\usepackage{microtype}      % microtypography
\usepackage{graphicx}
\usepackage{kotex}
\usepackage{array}
\usepackage{ragged2e}   % \RaggedRight 등 (선택)
\usepackage{longtable}
\usepackage{multirow}
\usepackage{makecell}
\usepackage{arydshln}
\usepackage{pifont}
\usepackage{amsmath}
\usepackage[most]{tcolorbox}
\usepackage{alltt}
\usepackage{wrapfig}
\usepackage{caption}
\usepackage{etoc}
\usepackage{appendix}
\usepackage{titletoc}
\definecolor{hansayellow}{rgb}{0.91, 0.84, 0.42}
\definecolor{checkingcolor}{HTML}{82C91E}
\newcommand{\myxmark}{\textcolor{red}{\ding{55}}}
\newcommand{\mychecking}{\textcolor{checkingcolor}{\ding{51}}}
\newcommand{\mytablecheck}{\ding{51}}

\newcommand{\Lbranch}{\rule[0.1ex]{0.5pt}{1.2ex}\rule[0.1ex]{0.7em}{0.5pt}\,}
\newcommand{\namefig}{Fig.}
\usepackage{enumitem}
\usepackage{setspace}
\tcbset{
  aibox/.style={
    width=\textwidth,
    top=10pt,
    colback=white,
    colframe=black,
    colbacktitle=black,
    enhanced,
    center,
    attach boxed title to top left={yshift=-2mm,xshift=2mm},
    boxed title style={boxrule=0pt,colframe=white,},
  }
}
\newtcolorbox{AIbox}[2][]{aibox,title=#2,#1}

\title{Towards Continuous Sign Language Conversation from Isolated Signs}

\author{
Youngmin Kim\textsuperscript{1}
\quad
Kyobin Choo\textsuperscript{1}
\quad
Jiwoo Park\textsuperscript{\scriptsize2}
\quad
Minseo Kim\textsuperscript{1}
\\
\textbf{Chanyoung Kim}\textsuperscript{3}
\quad
\textbf{Junhyeok Kim}\textsuperscript{1}
\quad
\textbf{Seong Jae Hwang}\textsuperscript{1}
\\[3pt]
\textsuperscript{1}Yonsei University
\quad
\textsuperscript{2}LG Electronics
\quad
\textsuperscript{3}Emory University
\\[3pt]
\texttt{winston1214@yonsei.ac.kr}
}

\begin{document}

\maketitle

\begin{abstract}
Sign language is the primary language for many Deaf and Hard-of-Hearing (DHH) signers, yet most conversational AI systems still mediate interaction through spoken or written language. This spoken-language-centered interface can limit access for signers for whom spoken or written language is not the most accessible medium, motivating direct sign-to-sign conversational modeling. However, sentence-level sign video data are expensive to collect and annotate, leaving existing sign translation and production models with limited vocabulary coverage and weak open-domain generalization. We address this bottleneck by constructing continuous sign conversations from isolated signs: large-scale labeled isolated clips are collected as lexically grounded motion primitives and recomposed into sign-language-ordered utterances derived from existing dialogue corpora. We introduce \textsc{SignaVox-W}, which provides, to our knowledge, the largest labeled isolated-sign vocabulary to date, and \textsc{SignaVox-U}, a continuous 3D sign conversation dataset built from \textsc{SignaVox-W}. To bridge structural mismatch between spoken and signed languages, we use a retrieval-guided spoken-to-gloss translator; to bridge independently collected isolated clips, we propose \emph{BRAID}, a diffusion Transformer that performs duration alignment and co-articulatory boundary inpainting. With the resulting data, we train \textsc{SignaVox}, a direct sign-to-sign conversational model that generates 3D body, hand, and facial motion responses from prior signing context without spoken-language text or externally provided glosses at inference time. Quantitative and qualitative evaluations show improved isolated-to-continuous motion quality, stronger response-level semantic alignment, and scalable signer-centered interaction that better supports visual-spatial articulation.% , non-manual cues, and conversational timing.
\end{abstract}

\input{sec/01introduction}
\input{sec/02Related_works}
\input{sec/03Dataset}
\input{sec/04ConversationModel}
\input{sec/06Experiments}
\input{sec/07Conclusion}

\begin{ack}
This work was supported in part by the IITP RS-2024-00457882 (AI Research Hub Project), IITP 2020-II201361, NRF RS-2024-00345806, NRF RS-2023-002620, NRF-2024S1A5C3A03046579, and RQT-25-120390. Affiliations: Department of Artificial Intelligence (Y.K, J.K, S.J.H), Department of Computer Science (K.C, M.K).
\end{ack}
% \begin{ack}
% Use unnumbered first level headings for the acknowledgments. All acknowledgments

% Do {\bf not} include this section in the anonymized submission, only in the final paper. You can use the \texttt{ack} environment provided in the style file to automatically hide this section in the anonymized submission.
% \end{ack}

\small
\bibliographystyle{plainnat}
\bibliography{main} 

%%%%%%%%%%%%%%%%%%%%%%%%%%%%%%%%%%%%%%%%%%%%%%%%%%%%%%%%%%%%
\clearpage
\appendix

\begin{center}
    {\LARGE \bfseries Appendix\par}
\end{center}
\vspace{1em}

% appendix 전용 TOC 시작점
\startcontents[appendix]

% \begin{center}
%     {\large \bfseries Contents\par}
% \end{center}
\vspace{0.5em}

% appendix 이후의 section/subsection만 출력
\printcontents[appendix]{}{1}{}

\vspace{1.5em}
\clearpage
\input{suppl/Limitations}
\input{suppl/Ethical_consideration}
\input{suppl/SignLanguage}

\input{suppl/Details_of_dataset}
\input{suppl/Details_frame_selection}
\input{suppl/Details_of_braid}

\input{suppl/Details_t2g}
\input{suppl/Details_sign2sign}
\input{suppl/Additional_Experiments}
\input{suppl/LLM_prompt}

%%%%%%%%%%%%%%%%%%%%%%%%%%%%%%%%%%%%%%%%%%%%%%%%%%%%%%%%%%%%
% \clearpage
% \newpage

% \input{checklist.tex}

\end{document}

%% file: sec/01Introduction.tex
\section{Introduction}
\label{sec:intro}
Sign language is an independent visual language and serves as the primary means of communication in DHH (Deaf and Hard-of-Hearing) communities~\cite{padden1988:deaf}, functioning as a first language for many of its users. Crucially, however, first-language proficiency in sign language does not imply proficiency in the surrounding spoken or written language. In practice, due to factors such as early language deprivation~\cite{sign_early1,parent,berger2024parent} and limited access to spoken language and literacy support~\cite{lederberg2013language, lederberg2014foundations}, many deaf signers acquire the spoken language as a second language~\cite{morford2011deaf,morford2014bilingual,goodwin2023deaf}. Therefore, conversational AI systems that rely on spoken or written language may be less accessible to many DHH signers, motivating sign-centered models that interact directly in sign language.

\begingroup
\renewcommand\thefootnote{}
\footnotetext{Definitions of sign language terms are provided in Appendix~\ref{sec:key_concept}.}
\endgroup

% A seemingly straightforward way is to combine sign language translation~\cite{slt_based}, text-based response generation, and sign language production (SLP)~\cite{slp, progressive_slp} in a cascade. However, this indirect pipeline can obscure visual-spatial structure and timing, and does not directly model how sign responses follow from prior signing context. This motivates direct sign-to-sign response generation, where both context and response are represented as sign motion.
A seemingly straightforward way is to combine sign language translation~\cite{slt_based}, text-based response generation, and sign language production (SLP)~\cite{slp, progressive_slp, signavatars} in a cascade. However, the SLP component in such a pipeline remains difficult to scale, since current SLP models are trained on limited sentence-level sign data and may generalize poorly to open-domain conversational text. Moreover, this indirect pipeline can obscure visual-spatial structure and timing, and does not directly model how sign responses follow from prior signing context. This motivates direct sign-to-sign response generation, where both context and response are represented as sign motion.
\begin{figure}[t]
    \centering
    \includegraphics[width=\linewidth]{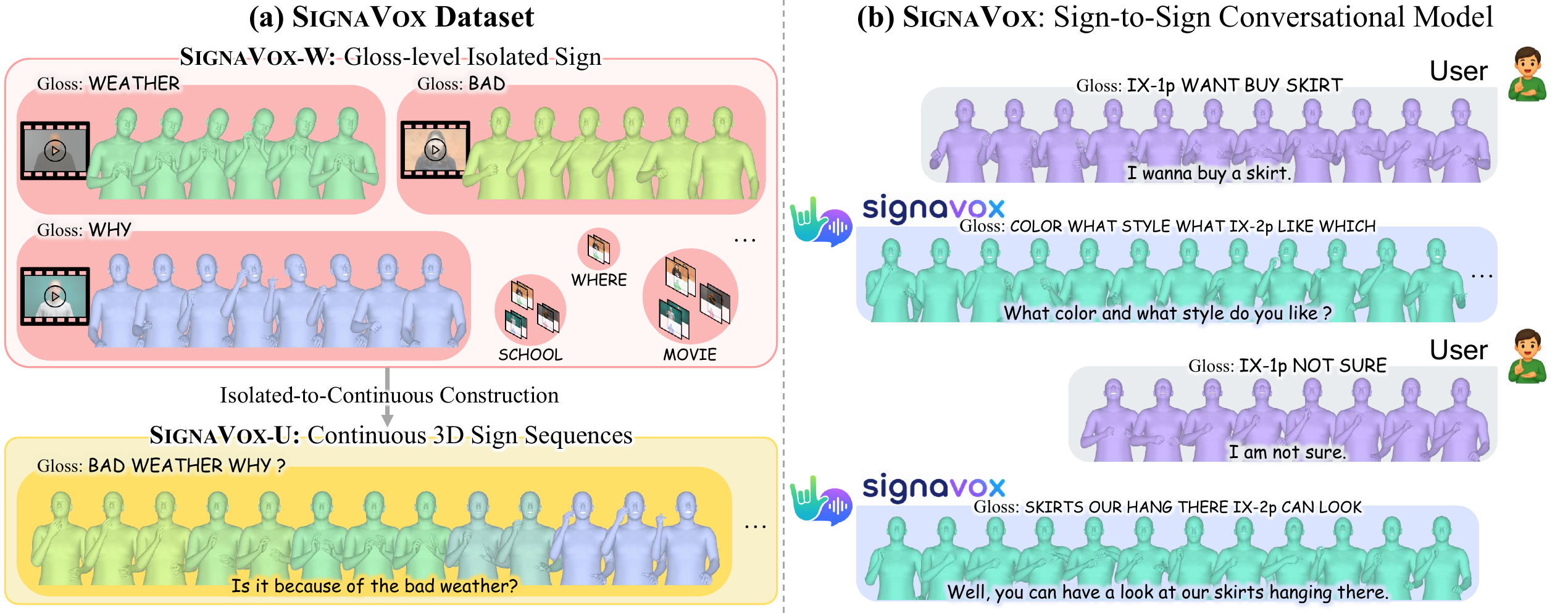}
    \caption{Overview of our proposed dataset and model. (a) We introduce 3D sign language datasets, extending from gloss-level isolated signs (\textsc{SignaVox-W}) to continuous sign sequences (\textsc{SignaVox-U}).
(b) We present \textsc{SignaVox}, which generates 3D sign responses from prior signing context.}
    \label{fig:teaser}
    \vspace{-10pt}
\end{figure}

Training sign-to-sign conversational models requires paired data in which prior signing context is followed by a sign response. However, large-scale sign-to-sign conversational datasets are not readily available; existing continuous sign datasets~\cite{phoenix14, how2sign, openasl} are mostly designed for translation or domain-specific instruction rather than open-ended response generation. We therefore scale data construction by combining abundant text-based dialogue corpora with relatively available labeled isolated-sign datasets: dialogue responses are converted into gloss sequences, and the corresponding isolated signs are composed as lexically grounded motion primitives.
% One option is to use SLP models~\cite{slp, progressive_slp, signavatars}, but current SLP models are trained on limited sentence-level sign data and may generalize poorly to open-domain conversational text. Instead, we leverage labeled isolated-sign datasets as lexically grounded motion primitives. This enables scalable construction of diverse pseudo-continuous sign responses, while providing a more controlled and reliable than generating all motion from a low-resource SLP model.

% To instantiate this strategy, we first construct \textsc{SignaVox-W}, a large-scale labeled isolated-sign dataset collected from diverse web sources and public datasets and represented shared 3D motion space. However, 
However, constructing sign-to-sign conversational data from isolated signs introduces two key challenges. First, at the visual-motion level, \emph{co-articulatory\footnote{Co-articulation refers to neighboring signs influencing each other's motion.} mismatch} arises when independently collected isolated clips are composed into continuous signing. To tackle this, we instantiate our strategy in two steps. We first construct \textsc{SignaVox-W}, a large-scale labeled isolated-sign dataset collected from diverse web sources and public datasets, represented in a shared 3D motion space. Although we retrieve a core-signing clip for each sign unit after trimming non-lexical preparation and retraction motions, adjacent clips remain misaligned in duration, boundary pose, and motion trajectory. Unlike prior works on prepared sign transitions~\cite{signd2c, g2pddm}, our setting requires composing continuous signing from independently collected clips, making both duration alignment and boundary refinement necessary. We therefore propose \emph{BRAID} (\textbf{\underline{B}}oundary \textbf{\underline{R}}efinement via co-\textbf{\underline{A}}rticulatory \textbf{\underline{I}}npainting \textbf{\underline{D}}iffusion Transformer), which aligns adjacent sign pairs with a predicted duration plan and refines boundary regions through co-articulatory inpainting to produce pseudo-continuous signing.

% First, at the visual-motion level, \emph{co-articulatory\footnote{Co-articulation refers to neighboring signs influencing each other's motion.} mismatch} arises when independently collected isolated clips are composed into continuous signing. Although we retrieve a core-signing clip for each sign unit after trimming non-lexical preparation and retraction motions, adjacent clips remain misaligned in duration, boundary pose, and motion trajectory. Unlike prior works on prepared sign transitions~\cite{signd2c, g2pddm}, our setting requires composing continuous signing from independently collected clips, making both duration alignment and boundary refinement necessary. We therefore propose \emph{BRAID} (\textbf{\underline{B}}oundary \textbf{\underline{R}}efinement via co-\textbf{\underline{A}}rticulatory \textbf{\underline{I}}npainting \textbf{\underline{D}}iffusion Transformer), which aligns adjacent sign pairs with a predicted duration plan and refines boundary regions through co-articulatory inpainting to produce pseudo-continuous signing.

Second, at the linguistic level, \emph{structural mismatch} arises because spoken and signed languages differ in word order and grammar. We therefore reorder spoken language turns into sign language ordered \textit{gloss}\footnote{Glosses are written labels for signs in sign language order. (e.g., I don’t have any dogs. $\rightarrow$ \texttt{DOG I HAVE NONE)}} sequences. To obtain these sequences, we use an LLM-based spoken-to-gloss translator with translation-memory-based RAG~\cite{rag}. We apply this translator to existing spoken language conversation datasets~\cite{realtalk, everydayconversations, dailydialog}, and transform the resulting gloss sequences into continuous signing with \emph{BRAID} and the isolated signs in \textsc{SignaVox-W}. This results in \textsc{SignaVox-U}, a large-scale sign language conversational dataset represented as sequences of 3D features for training sign-to-sign conversational models.

Building on \textsc{SignaVox-U}, we introduce \textsc{SignaVox}, a sign-to-sign conversational model for sign-centered interaction.
\textsc{SignaVox} generates continuous 3D sign motion responses directly from prior signing context and a target role prompt, without routing the interaction through spoken-language text or requiring externally provided gloss sequences at inference time.
It models responses autoregressively in motion blocks with conditional flow-matching heads~\cite{flowmatching, conditional_flowmatching}, providing a first step toward sign-centered conversational modeling.

An overview of the proposed dataset and sign-to-sign conversational model is shown in \namefig{}~\ref{fig:teaser}. 
We evaluate the framework with quantitative metrics and qualitative analyses. \emph{BRAID} improves isolated-to-continuous motion quality, while \textsc{SignaVox} outperforms motion-to-motion baselines in 3D response generation. Together, these results demonstrate that sign-centered conversational modeling can be scaled by combining sign-compatible linguistic planning with real sign motions.

\noindent \textbf{Contributions.} Our main contributions are as follows:
\vspace{-3pt}
\begin{itemize}[nosep, leftmargin=*]
\item We introduce \textsc{SignaVox-W} and \textsc{SignaVox-U}, a large-scale isolated-sign dataset and a sign-to-sign conversational dataset constructed via spoken-language-to-gloss translation and isolated-to-continuous 3D motion construction.
\item We develop \emph{BRAID}, an isolated-to-continuous sign generation model that composes independently collected isolated sign clips into continuous signing through duration alignment and boundary refinement.
\item We present, to the best of our knowledge, the first direct sign-to-sign conversational model that takes sign motion as input and generates continuous 3D sign motion responses without text or externally provided gloss sequences at inference time.
\end{itemize}
\vspace{-8pt}

%% file: sec/02Related_works.tex
\section{Related Works}
% \noindent \textbf{Sign Language.}
% Sign language is not a mere visual counterpart of spoken language, but a natural language with its own independent grammar and structure. It conveys meaning through \textbf{manual components} (\eg, handshape, location, movement, and orientation), together with \textbf{non-manual signals (NMS)} (\eg, facial expressions, gaze, head movements, and body leans). These elements combine to express not only word-level meanings but also grammatical functions, discourse structure, and the signer’s intent.

\noindent \textbf{Sign Language Dataset.}
% Sign language processing has long been hindered by the lack of large-scale annotated datasets~\cite{bragg2019:sign}, though recent efforts have expanded available resources.
% For ASL, many isolated-sign datasets have been introduced~\cite{msasl, wlasl, asllex, asllvd}, but their isolated sign clips  focus often misses key properties of continuous signing such as co-articulation~\cite{slt_based, coarticulation}.
% To address these limitations, continuous sign language video datasets have recently been proposed, covering not only ASL but also a wide range of languages such as German, Korean and Chinese Sign Languages (DGS, KSL, and CSL)~\cite{how2sign, ksl, phoenix12, phoenix14, csl}. These datasets enable sentence-level understanding and generation by providing continuous sign expressions that more closely reflect real-world signing scenarios.
% Nevertheless, continuous sign language datasets remain difficult to scale due to the high cost and expertise required for gloss annotation. To mitigate this issue, several web-scale sign language datasets~\cite{openasl, youtubeasl, multi_youtube} have been introduced; however, they rely on captions and do not provide explicit written representations such as gloss annotations.
Sign language processing has long been limited by the lack of large-scale annotated datasets~\cite{bragg2019:sign}. While many ASL (American Sign Language) isolated-sign datasets have been introduced~\cite{msasl, wlasl, asllex, asllvd}, isolated clips often fail to capture key properties of continuous signing, such as co-articulation~\cite{slt_based, coarticulation}. Continuous sign video datasets address this by providing sentence-level data across diverse sign languages~\cite{how2sign, ksl, phoenix12, phoenix14, csl}, but scaling them remains difficult because gloss annotation requires costly expertise. Web-scale datasets mitigate the scale issue~\cite{openasl, youtubeasl, multi_youtube}, yet they typically rely on captions and lack explicit gloss annotations.

\noindent \textbf{Sign Language Processing.}
% Sign language processing task has evolved around three main tasks: recognition, translation, and generation. The recognition task focuses on modeling the temporal structure of continuous signing to predict sign-internal representations using gloss annotations~\cite{twostream_slr, stmc}. In contrast, the translation task aims to map sign language videos directly into spoken language sentences, requiring higher-level semantic and linguistic reasoning~\cite{slt_based, better_stmc}. Recently, gloss-free translation methods have gained increasing attention to overcome the limitations and cost of gloss labeling~\cite{kym_slt, glossfree_slt, glossfree_slt2}.
% Finally, generation task synthesizes sign motions from text or gloss inputs. 
% Earlier methods often represented sign language using 2D or 3D keypoints~\cite{slp_keypoint,progressive_slp}, while recent work increasingly employs 3D human models or avatars to provide richer spatial representations and improved visual realism~\cite{signavatars, simple_baseline}.
% Recently, large language models (LLMs) have emerged as key backbones for sign language translation and generation.
% For translation, LLM-based approaches improve spoken-language sentence generation from sign inputs in a gloss-free manner~\cite{signgpt,zhang2025:large, llm_good_translator}.
% For generation, LLMs have been explored for sign language production from text prompts~\cite{signllm} and further extended to synthesize NMS alongside manual articulations~\cite{nms_generation}.
Sign language processing has developed largely around recognition, translation, and generation. Recognition predicts the sequence of signs or glosses from continuous signing videos by modeling their temporal progression, often with gloss supervision~\cite{twostream_slr, stmc}. Translation maps sign videos into spoken-language sentences, and recent gloss-free methods aim to reduce dependence on costly gloss labels~\cite{slt_based, better_stmc, kym_slt, glossfree_slt, glossfree_slt2, unisign}. Generation, or sign language production, synthesizes sign motions from text or gloss inputs, evolving from keypoint-based representations~\cite{slp_keypoint, progressive_slp} to 3D human models and avatars~\cite{signavatars, simple_baseline}. Recently, LLMs have also been explored for gloss-free translation~\cite{llm_good_translator, signgpt, zhang2025:large} and text-conditioned sign production~\cite{signllm, nms_generation}.

\noindent \textbf{Co-articulation-Aware Sign Language Modeling.}
Co-articulation is central to natural continuous signing, yet its influence on the timing and motion of adjacent signs often blurs temporal boundaries, making it a key challenge for modeling continuous sign language~\cite{bsl,coarticulation,recognition_coarticulation,recognition_coarticulation2}.
In sign language generation, this challenge is instead framed as the explicit synthesis of co-articulation, where keypoint or mesh-based models learn motion deformation and temporal continuity between adjacent signs~\cite{simple_baseline,signavatars,coarticulation}.
A more specific line of transition-generation methods focuses on transition poses between adjacent segments to alleviate abrupt motion, rather than regenerating entire sign segments~\cite{signd2c,g2pddm}.

%% file: sec/03Dataset.tex
% \begin{figure}[t]
%   \centering
%   % \fbox{\rule{0pt}{2in}\rule{0.9\linewidth}{0pt}}
  
%   \caption{Overall data collection and processing pipeline.}
%   \label{fig:data_pipeline}
% \end{figure}
\begin{figure}
    \centering
    \includegraphics[width=\linewidth]{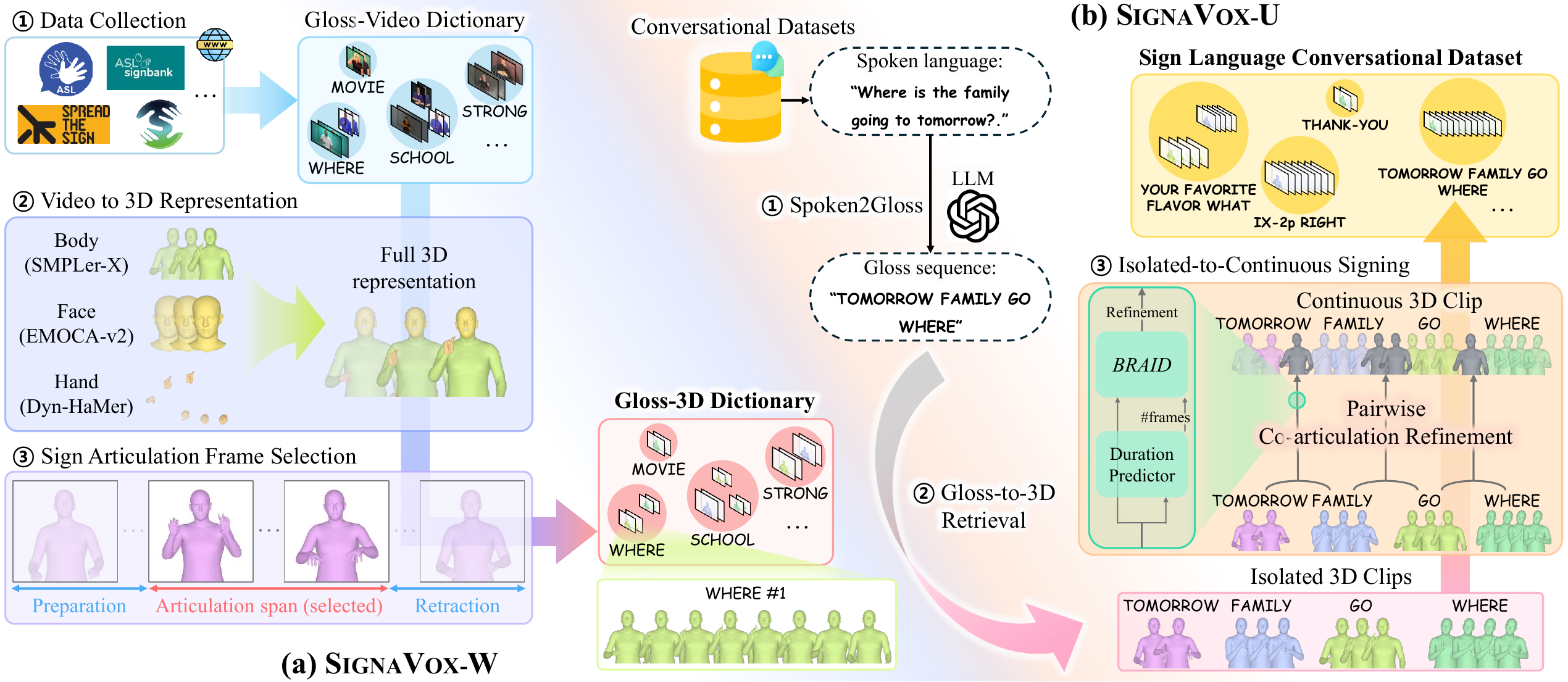}
    \vspace{-15pt}
    \caption{Overall data collection and processing pipeline. (a) illustrates the collection process for our isolated sign language dataset, \textsc{SignaVox-W}. (b) details the construction of \textsc{SignaVox-U}, a conversational dataset created by synthesizing continuous sentences from \textsc{SignaVox-W}.}
    \label{fig:data_pipeline}
    \vspace{-10pt}
\end{figure}

\input{tables/comparison_dataset}
\section{SIGNAVOX Dataset}
% Existing sign language datasets either cover a limited vocabulary~\cite{msasl, wlasl} or lack gloss annotations for continuous sign language videos~\cite{how2sign, openasl}. As a result, aligning video segments with gloss labels is difficult, limiting lexical-unit learning and co-articulation modeling for continuous signing. Consequently, large-scale and structured sign language conversation datasets remain scarce.

In this section, we introduce (1) the construction pipeline of \textsc{SignaVox-W}, a large-scale isolated sign clips collected from diverse sources (Sec.~\ref{sec:signavoxw}), (2) an overview of our isolated-to-continuous signing model, which converts isolated sign clips into continuous sign sequences with natural co-articulation (Sec.~\ref{sec:isolated_continuous}), and (3) the construction of \textsc{SignaVox-U}, an utterance-level continuous sign dataset built on these two components, enabling broad utterance coverage for improved generality (Sec.~\ref{sec:signavoxu}). 
% Together, these components provide the data foundation needed to train sign language dialogue models.
\subsection{Isolated Sign Dataset Construction}
\label{sec:signavoxw}
% To support scalable sign language conversation modeling, we construct \textsc{SignaVox-W}, a large-scale word-level video--gloss lexicon with canonical gloss labels curated from diverse sources.
% Since such in-the-wild data exhibits substantial variability in viewpoints, recording conditions, and signer localization, we further apply 3D normalization and systematic quality control (QC). Table~\ref{tab:comp_dataset} summarizes dataset statistics in comparison to prior ASL datasets, and Fig.~\ref{fig:data_pipeline} illustrates the overall collection and processing pipeline.
We construct \textsc{SignaVox-W}, a large-scale sign video lexicon with canonical gloss labels for scalable sign language conversation modeling.
Table~\ref{tab:comp_dataset} summarizes dataset statistics in comparison to existing ASL datasets, and \namefig{}~\ref{fig:data_pipeline}-(a) illustrates the overall collection and preprocessing pipeline.

% Existing word-level sign language datasets typically have limited vocabulary coverage. In contrast, scalable sign language conversation modeling requires a word-level lexical resource that spans a broad vocabulary and provides sufficient video--gloss mappings. To this end, we construct \textsc{SignaVox-W}, a word-level video--gloss lexicon with canonical gloss labels. Table~\ref{tab:comp_dataset} compares statistics of existing ASL datasets and shows that \textsc{SignaVox-W} offers richer annotations and substantially broader vocabulary coverage. The overall data collection and processing pipeline is illustrated in Fig~\ref{fig:data_pipeline}.

\noindent \textbf{Data Collection.}
Following previous studies~\cite{coarticulation, simple_baseline}, we construct a video–gloss dictionary. To ensure generality, we collect data from four web-based sign language dictionaries~\cite{signasl, signbank, signingsavvy, spreadthesign} that provide sign demonstrations performed by native signers, as well as two public datasets~\cite{msasl, wlasl}, rather than relying on a single dataset, thereby achieving broad vocabulary coverage. This results in a one-to-many mapping, where each gloss is associated with multiple video samples. Details of our dataset are provided in the Appendix~\ref{sec:supp_dataset}.

\noindent \textbf{3D Representation.}
% \textcolor{red}{Pelvis1 을 0으로 만듦. Supplementary 에 넣는게 좋을듯?}
To mitigate spatial misalignment and domain shifts inherent in multi-source videos, we adopt a normalized 3D representation. Following~\cite{signavatars, nonverbal}, we estimate part-specific 3D parameters for the body, hands, and face to explicitly capture fine-grained manual and non-manual cues.
We estimate full-body motion using SMPLer-X~\cite{smplerx} with SMPL-X parameters~\cite{smpl-x}, model hand articulations via Dyn-HaMR~\cite{dynhamr} based on MANO~\cite{MANO}, and estimate facial expressions using EMOCA-v2~\cite{emocav2} based on FLAME~\cite{flame}.
For each gloss clip {\small$g_k$}, we represent its $t$-th frame using a unified 3D motion feature vector that concatenates body, face, and hand parameters. Specifically,
{\small
\begin{equation}
    \mathbf{x}^{(k)}_t 
    = \big[\,\boldsymbol{\theta}^{\mathrm{body}}_t;\
            \boldsymbol{\psi}_{t}; \
            \boldsymbol{\theta}^{\mathrm{jaw}}_{t}; \
             \boldsymbol{\theta}^{\mathrm{rhand}}_t;\
            \boldsymbol{\theta}^{\mathrm{rhand}}_t;\ \big]
       \in \mathbb{R}^{D}.
\label{eq:def_feature}
\end{equation}
}
where {\small$D=206$}. Accordingly, the frame-wise motion representation of {\small$g_k$} is denoted as {\small$\mathbf{X}^{(k),\mathrm{raw}} = [\mathbf{x}^{(k),\mathrm{raw}}_{1}, \cdots, \mathbf{x}^{(k),\mathrm{raw}}_{T_k}] \in \mathbb{R}^{T_k \times D}$}, where {\small$T_k$} is the number of frames in {\small$g_k$}.

\noindent \textbf{Sign Articulation Frame Selection.}
Modeling natural co-articulation in sign language videos requires identifying frames that correspond to the core lexical articulation of each sign~\cite{coarticulation}.
We introduce a coarse-to-fine articulation-frame selection pipeline to isolate the core articulation while discarding preparation and retraction segments.
First, we perform feature-based coarse boundary estimation to remove long non-signing intervals and narrow each isolated sign clip to its effective signing region.
Given an isolated sign clip represented as motion features {\small $\mathbf{X}^{(k),\mathrm{raw}}$}, we estimate a frame-wise motion-energy signal from smoothed temporal derivatives and combine it with arm-posture gating to suppress low-activity arm-down segments.
The resulting signal is used to estimate coarse temporal boundaries {\small $(s_k,e_k)$}, which are further refined in the next stage.
% Details of the energy computation and boundary estimation are provided in Appendix~\ref{sec:suppl_engery_based}.

Based on the estimated coarse boundaries {\small $(s_k, e_k)$}, we crop the corresponding video segment and apply a two-stage VideoLLM-based~\cite{qwen3vl} refinement procedure to identify the core lexical articulation.
To facilitate more accurate temporal reasoning, we adopt a visual prompting strategy~\cite{vp, numberit} that explicitly overlays frame indices on each frame~\cite{vikey} (e.g., ``\textit{frame \#01}'').
In the first stage, the model predicts a contiguous articulation span ({\small $\hat{s}_k, \hat{e}_k$}), which provides an initial estimate of the articulation interval.
In the second stage, we present a candidate-focused clip centered around the predicted span and further refine its boundaries by removing residual non-signing segments. For fingerspelled cases such as alphabet letters, where a single canonical handshape is often sufficient, we select one representative frame. As a result, the final articulation-preserving representation is obtained as {\small$\mathbf{X}^{(k),\mathrm{raw}}_{\hat{s}_{k}:\hat{e}_{k}}$} or single-frame feature. In the following section, we notate {\small$\mathbf{X}^{(k),\mathrm{raw}}_{\hat{s}_{k}:\hat{e}_{k}} \in \mathbb{R}^{(\hat{e}_{k} - \hat{s}_{k}+1) \times D}$} as {\small$\mathbf{X}^{(k)} \in \mathbb{R}^{\hat{T}_k \times D}$}, and define the adjacent gloss-pair sequence as {\small $\mathbf{X}^{(k, k+1)}$} as {\small$[\mathbf{X}^{(k)};\mathbf{X}^{(k+1)}]$}.

\begin{figure}[t]
  \centering
  \includegraphics[width=\linewidth]{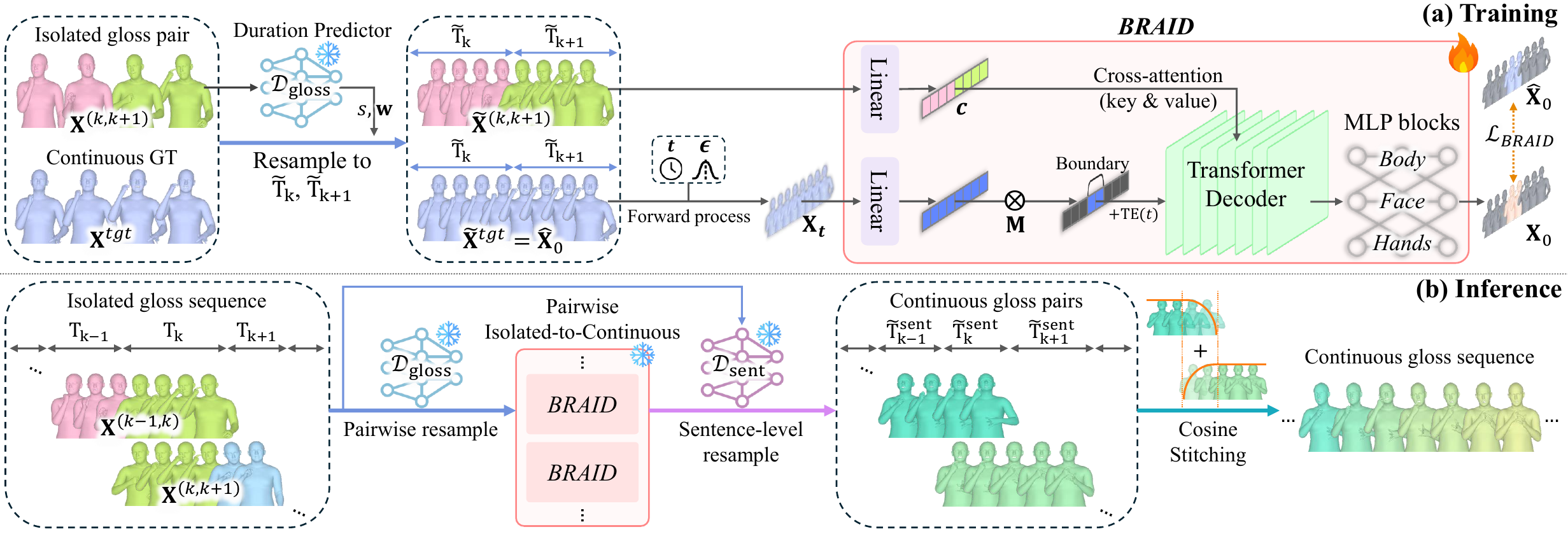}
  \caption{Overview of the \emph{BRAID} framework. (a) The training pipeline of our proposed model. (b) The inference process for generating the sentence-level continuous sign language.}
  \label{fig:model_pipeline}
  \vspace{-10pt}
\end{figure}

\subsection{Isolated-to-Continuous Signing} % 제목 수정 예정
\label{sec:isolated_continuous}
In this section, we describe our pipeline for composing sentence-level signing from isolated clips in \textsc{SignaVox-W}. Since directly concatenating isolated signs leads to temporal and articulatory mismatch at gloss boundaries, we process adjacent gloss pairs with \emph{BRAID} (\textbf{\underline{B}}oundary \textbf{\underline{R}}efinement via co-\textbf{\underline{A}}rticulatory \textbf{\underline{I}}npainting \textbf{\underline{D}}iffusion Transformer) and stitch the refined outputs into a continuous signing sequence. The overall pipeline is shown in \namefig{}~\ref{fig:model_pipeline}.
% In this section, we present a pipeline for composing natural sentence-level signing from isolated sign clips from \textsc{SignaVox-W}. This setting requires constructing continuous sign sequences from discrete gloss-level clips while restoring natural co-articulation and reducing temporal and articulatory mismatch around adjacent gloss boundaries. To simplify the problem, we model adjacent gloss pairs in a sliding-window manner and introduce \emph{BRAID} (\textbf{\underline{B}}oundary \textbf{\underline{R}}efinement via co-\textbf{\underline{A}}rticulatory \textbf{\underline{I}}npainting \textbf{\underline{D}}iffusion Transformer), which refines the boundary regions between $\mathbf{\hat{X}}^{(k)}$ and $\mathbf{\hat{X}}^{(k+1)}$. The refined gloss-pair outputs are finally stitched at the sentence level to form the final continuous signing sequence. The overall pipeline is shown in Figure~\ref{fig:model_pipeline}.

\noindent \textbf{Data Setting.}
% \noindent \textbf{Gloss-Guided Sentence Composition.}
% We construct training data for learning sequential dependency and co-articulation from isolated sign clips by leveraging ASLLRP~\cite{asllrp} as a source of continuous signing supervision. Since ASLLRP provides gloss-text-video annotations, we first organize the data into sentence-level utterances. During this process, gloss labels are normalized following the gloss conventions described in the Appendix~\ref{suppl:retrieval_rule1}. For each sentence, we retrieve gloss-aligned isolated sign clips from \textsc{SignaVox-W} and synthesize sentence-level clip compositions by sampling one candidate clip per gloss. We further augment the data by repeating this composition process $n$ times for each sentence to obtain diverse sentence-level combinations, where $n=14$. The resulting composed sequences are then decomposed into adjacent gloss pairs to define the training instances. 
We construct training pairs by using ASLLRP~\cite{asllrp} as continuous signing supervision. After organizing ASLLRP into sentence-level utterances with normalized gloss labels, we retrieve corresponding isolated clips from \textsc{SignaVox-W} and compose pseudo sentence-level sequences by sampling one clip per gloss. We repeat this process {\small $n=14$} times for diversity and decompose the composed sequences into adjacent gloss pairs for training.

% Let $g_k$ denote the retrieved gloss clip corresponding to the $k$-th gloss in the sentence. We represent its $t$-th frame as a feature vector $\mathbf{x}_t^{(k)}$ defined as
% \begin{equation}\small
%     \boldsymbol{\mathbf{x}}^{(k)}_t = \big[\,\mathbf{\boldsymbol{\theta}}^{\mathrm{body}}_{t};\, \mathbf{r}^{\mathrm{head}}_{t} \,;\, \boldsymbol{\mathbf{\psi}}_t \,;\, \mathbf{\boldsymbol{\theta}}^{\mathrm{jaw}}_{t} ;\, \boldsymbol{\theta}^{\mathrm{rhand}}_t ;\ \boldsymbol{\theta}^{\mathrm{lhand}}_t\,\big]\in\mathbb{R}^{D},
% \end{equation}
% where $D=206$. Accordingly, we denote the frame-wise motion representation of $g_k$ by $\mathbf{X}^{(k)} = \big[\mathbf{x}^{(k)}_{1}, \cdots \mathbf{x}^{(k)}_{T_{k}} \big] \in \mathbb{R}^{T_{k} \times D}$, where $T_{k}$ denotes the number of frames in $g_k$.

% In total, we retain 1,337 utterance videos.

% To help the model reason over temporal structure, we adopt a visual prompting strategy~\cite{vp,numberit} that explicitly provides frame index information (e.g., ``\textit{frame \#01}''). In practice, we use Qwen3-VL~\cite{qwen3vl} with the ViKey prompting scheme to extract $\mathcal{I}^{(k)}$. Finally, we retain only the selected feature subsequence,
% \begin{equation}
%     \hat{f} = \{f^{(k)}_{i} | i \in \mathcal{I}^{(k)} \},
% \end{equation}
% which is used as the articulation-preserving representation of the $k$-th word clip. This coarse-to-fine design allows us to efficiently discard obvious non-signing segments using motion cues, while relying on the VideoLLM for finer semantic selection.

\begin{wrapfigure}{r}{0.4\textwidth}
    \vspace{-15pt}
    \centering
    \includegraphics[width=\linewidth]{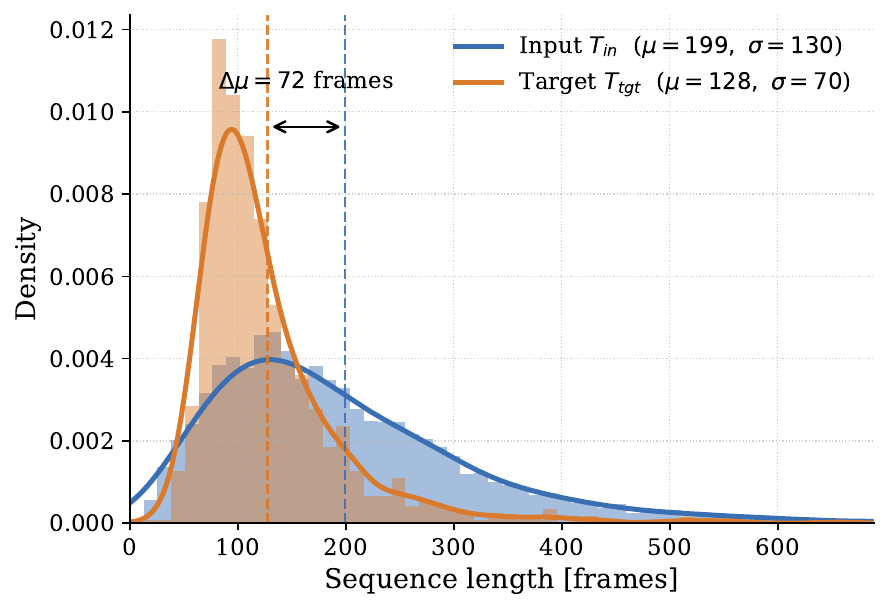}
    \captionsetup{skip=2pt}
    \caption{Distribution of sequence lengths for composed gloss-pair inputs and target continuous segments.}
    \label{fig:length_distribution}
    \vspace{-10pt}
\end{wrapfigure}

\noindent \textbf{Duration Prediction.}
As shown in \namefig{}~\ref{fig:length_distribution}, the gloss-pair inputs exhibit a substantial length 
mismatch with the target continuous segments. 
Specifically, the input sequences are not only longer on average than the targets, but also display considerably larger variance. This indicates that redundant temporal content still remains around gloss boundaries even after articulation-frame selection. 
To reduce this discrepancy, we use two duration predictors, denoted by {\small $\mathcal{D}_{\mathrm{gloss}}$} and {\small $\mathcal{D}_{\mathrm{sent}}$}. 
{\small $\mathcal{D}_{\mathrm{gloss}}$} reduces \emph{BRAID}'s refinement burden by adjusting local gloss-pair durations, while {\small $\mathcal{D}_{\mathrm{sent}}$} provides a consistent temporal plan for stitching pairwise outputs into full sentences.
% \begin{wrapfigure}{r}{0.4\textwidth}
%     \vspace{-15pt}
%     \centering
%     \includegraphics[width=\linewidth]{figures/length_mismatch_distribution.pdf}
%     \captionsetup{skip=2pt}
%     \caption{Distribution of sequence lengths for composed gloss-pair inputs and target continuous segments.}
%     \label{fig:length_distribution}
%     \vspace{-15pt}
% \end{wrapfigure}
Let {\small $T_{\mathrm{src}}$} denote the length of the input motion sequence and {\small $T_{\mathrm{tgt}}$} its corresponding target length. 
The predictor estimates a global scale {\small $s=\log(T_{\mathrm{tgt}}/T_{\mathrm{src}})$} and a gloss-wise duration ratio {\small $\mathbf{w}=[w_{1}, \cdots, w_{K}]$}, where {\small $K$} denotes the number of glosses. 
The length assigned to the {\small $k$}-th gloss is then computed as
{\small $\tilde{T}_{k}=w_{k}T_{\mathrm{src}}e^{s}$}.
We train the predictor to match the target global scale and gloss-wise duration allocation, using a quantile objective for the scale term to discourage excessive temporal compression.
Additional architecture and training details are provided in Appendix~\ref{sec:suppl_duration}.

\noindent \textbf{Architecture.}
% Our goal is to refine temporally adjusted gloss-pair sequences, $\hat{\mathbf{X}}^{(k,k+1)}$, into natural continuous signing by restoring boundary transitions and fine-grained motion details. To achieve this goal, we propose \emph{BRAID}, a conditional diffusion transformer for boundary refinement and co-articulatory inpainting, inspired by MDM~\cite{mdm} for human motion generation. Rather than predicting the denoiser to correct large temporal mismatch and boundary artifacts at the same time, \textit{BRAID} separates these two challenges by first aligning the pseudo sequence to a predicted target duration plan and then focusing diffusion-based refinement on boundary regions. 
% We propose \emph{BRAID}, a conditional diffusion Transformer that refines temporally adjusted gloss-pair sequences $\mathbf{X}^{(k,k+1)}$ into continuous signing. 
% We propose \emph{BRAID}, a conditional diffusion Transformer that refines gloss-pair motion sequences into continuous signing. By focusing on boundary regions, \emph{BRAID} restores natural transitions and fine-grained co-articulatory motion.
% Inspired by MDM~\cite{mdm}, it decouples temporal alignment from motion refinement: pseudo sequences are first aligned to a predicted duration plan, and then refined through diffusion.
Inspired by MDM~\cite{mdm}, we propose \emph{BRAID}, a conditional diffusion Transformer that refines gloss-pair motion sequences into continuous signing. By focusing on boundary regions, \emph{BRAID} restores natural transitions and fine-grained co-articulatory motion. We first perform duration-aware temporal alignment to reduce length mismatch and establish the temporal scale before diffusion-based boundary refinement. The frozen duration predictor {\small$\mathcal{D}_{\mathrm{gloss}}$} estimates target durations {\small$\tilde{T}_k$} and {\small$\tilde{T}_{k+1}$} for the gloss-pair sequence {\small$\mathbf{X}^{(k,k+1)}$}.
After predicting {\small$\tilde{T}_k$} and {\small$\tilde{T}_{k+1}$}, we linearly resample the two gloss-motion segments to the corresponding durations and concatenate them to obtain the duration-adjusted sequence {\small$\tilde{\mathbf{X}}^{(k,k+1)} \in \mathbb{R}^{(\tilde{T}_k+\tilde{T}_{k+1}) \times D}$}. The concatenation point at index {\small$\tilde{T}_k$} defines the boundary between the two resampled gloss segments. Around this boundary, we construct a binary temporal inpainting mask {\small$\mathbf{M} \in \{0, 1\}^{\tilde{T}_k+\tilde{T}_{k+1}}$} as follows:
\begin{equation}
\small
   M_i = \mathbf{1} \left[ |i - \tilde{T}_k| \le r \right] , \quad r \sim \mathcal{U}\{r_{\min}, \dots, r_{\max}\} . 
\end{equation}
Frames with {\small $M_i = 1$} form the co-articulatory inpainting region, where the denoiser output is used for refinement and receives supervision during training. Frames with {\small $M_i = 0$} serve as temporal context and are preserved from the duration-adjusted input sequence in the final composition.

% We first use the pre-trained and frozen $\mathcal{D}_{\mathrm{gloss}}$ to estimate the global scale and gloss-wise duration allocation for $\mathbf{X}^{(k,k+1)}$, from which the target durations $\tilde{T}_k$ and $\tilde{T}_{k+1}$ are derived. Each gloss feature, $\mathbf{X}^{(k)}$ and $\mathbf{X}^{(k+1)}$, is then linearly resampled according to the derived durations, and the resulting segments are concatenated to form a duration-adjusted input sequence $\tilde{\mathbf{X}}^{(k,k+1)} \in \mathbb{R}^{(\tilde{T}_k+\tilde{T}_{k+1}) \times D}$.
% The frozen duration predictor $\mathcal{D}_{\mathrm{gloss}}$ estimates target durations $\tilde{T}_k$ and $\tilde{T}_{k+1}$ for the gloss-pair sequence $\mathbf{X}^{(k,k+1)}$.
% We linearly resample the two gloss segments to these durations and concatenate them to obtain the duration-adjusted sequence $\tilde{\mathbf{X}}^{(k,k+1)} \in \mathbb{R}^{(\tilde{T}_k+\tilde{T}_{k+1}) \times D}$.
% To focus refinement on the boundary between the two resampled gloss segments, we define a binary temporal inpainting mask \(\mathbf{M}\in\{0,1\}^{\tilde{T}_k+\tilde{T}_{k+1}}\), where
% \begin{equation} \small 
% M_i = 
% \begin{cases} 
%     1, & \text{if } |i-\tilde{T}_k| \le r,\quad r\sim\mathcal{U}(r_{\min},r_{\max}),\\ 
%     0, & \text{otherwise}. 
% \end{cases} 
% \end{equation} 
% Here, $M_i=1$ marks frames where the denoiser prediction is applied and supervised, while $M_i=0$ marks frames preserved from the duration-adjusted input sequence; we refer to the $M_i=1$ region as the co-articulatory inpainting region. 

During training, we use the matched continuous ground-truth pair {\small$\mathbf{X}^{\mathrm{tgt}}$} as supervision. To match the length of the duration-adjusted pseudo sequence, we resample it to the same predicted durations, obtaining the duration-aligned target {\small$\tilde{\mathbf{X}}^{\mathrm{tgt}}$}, which we denote as the clean target {\small$\mathbf{X}_0$}. We then corrupt only $\mathbf{X}_0$ with the standard DDPM forward process~\cite{ddpm} at timestep {\small$t$}, producing the noisy target {\small$\mathbf{X}_t$}, while the duration-adjusted pseudo sequence {\small$\tilde{\mathbf{X}}^{(k,k+1)}$} remains clean. The noisy target $\mathbf{X}_t$ is linearly projected as the decoder target stream, while the clean pseudo sequence is projected into conditioning tokens {\small$\mathbf{c} = \phi_c(\tilde{\mathbf{X}}^{(k,k+1)})$}, which serve as the key and value memory for cross-attention. The denoiser {\small $G_\theta$} is implemented as a RoPE-based~\cite{rope} Transformer decoder. Given {\small $\mathbf{X}_t$}, timestep {\small $t$}, conditioning tokens {\small $\mathbf{c}$}, and the inpainting mask {\small $\mathbf{M}$}, it predicts the clean motion {\small $\hat{\mathbf{X}}_0 = G_\theta(\mathbf{X}_t, t, \mathbf{c}, \mathbf{M})$}.
The Transformer decoder output is passed to part-specific MLP heads for body, face, and hands, whose outputs are concatenated to produce the clean-motion prediction {\small $\mathbf{\hat{X}}_0$}.
% During training, we use the matched continuous ground-truth pair $\mathbf{X}^{\mathrm{tgt}}$ as supervision. To match the length of the duration-adjusted pseudo sequence, we resample it to the same predicted durations and denote the resulting clean target as $\mathbf{X}_0$. We then corrupt only $\mathbf{X}_0$ with the standard DDPM forward process~\cite{ddpm} at timestep $t$, producing the noisy target $\mathbf{X}_t$, while the duration-adjusted pseudo sequence $\tilde{\mathbf{X}}^{(k,k+1)}$ remains clean. The clean pseudo sequence is linearly projected into conditioning tokens $\mathbf{c} = \phi_c(\tilde{\mathbf{X}}^{(k,k+1)})$, which serve as the key and value memory for cross-attention. The denoiser $G_\theta$ is implemented as a RoPE-based~\cite{rope} Transformer decoder that takes $\mathbf{X}_t$, the timestep $t$, the conditioning tokens $\mathbf{c}$, and the inpainting mask $\mathbf{M}$, it predicts the clean motion $\hat{\mathbf{X}}_0 = G_\theta(\mathbf{X}_t, t, \mathbf{c}, \mathbf{M})$.
% Part-specific heads for body, face, and hands are concatenated to predict $\hat{\mathbf{X}}_0$. The final refined pair uses the denoised prediction within the mask-activated boundary region and preserves the duration-adjusted input sequence elsewhere.

\noindent \textbf{Training.} We train \emph{BRAID} with a masked objective consisting of a part-weighted reconstruction loss $\mathcal{L}_{\mathrm{recon}}$ and a velocity regularization $\mathcal{L}_{\mathrm{vel}}$ on first-order temporal differences: 
\begin{equation}
\small
    \mathcal{L}_{\mathrm{braid}} = w(t) \left( \mathcal{L}_{\mathrm{recon}}(\hat{\mathbf{X}}_0, \mathbf{X}_0; \mathbf{M}, \boldsymbol{\omega}_{\mathrm{part}}) + \lambda_{\mathrm{vel}} \mathcal{L}_{\mathrm{vel}}(\Delta \hat{\mathbf{X}}_0, \Delta \mathbf{X}_0; \mathbf{M}^{\Delta}, \boldsymbol{\omega}_{\mathrm{part}}) \right).
\end{equation}
Here, {\small $w(t)$} denotes the Min-SNR weighting~\cite{minsnr} at diffusion step {\small $t$}, {\small $\boldsymbol{\omega}_{\mathrm{part}}$} is a part-wise feature weight vector that emphasizes hand and face dimensions, and {\small $\mathbf{M}^{\Delta}$} denotes the temporal-difference mask for the velocity term. Detailed definitions of {\small $\mathcal{L}_{\mathrm{recon}}$ and $\mathcal{L}_{\mathrm{vel}}$} are provided in the Appendix.

\noindent \textbf{Inference.}
At inference, we first use {\small $\mathcal{D}_{\text{gloss}}$} to predict pair-level durations for each adjacent gloss pair. Each pair is resampled accordingly and refined through the DDIM reverse process~\cite{ddim} using a symmetric boundary inpainting mask, while the {\small $M_i = 0$} frames are preserved from the duration-adjusted input. In parallel, {\small $\mathcal{D}_{\mathrm{sent}}$} predicts a sentence-level duration plan {\small $\tilde{\mathbf{T}}^{\mathrm{sent}}$}. The refined pairwise outputs are split at their predicted gloss boundary, assembled into a sentence-level trajectory, and cosine-fused where intermediate glosses overlap. Finally, the stitched sequence is rescaled gloss-wise to match {\small $\tilde{\mathbf{T}}^{\mathrm{sent}}$}, resulting in {\small $\hat{\mathbf{X}}^{\mathrm{sent}} \in \mathbb{R}^{\sum_{k}\mathbf{\tilde{T}}^{\mathrm{sent}} \times D}$}.

% We first use \(\mathcal{D}_{\mathrm{gloss}}\) to predict pair-level durations for each adjacent gloss pair. Each pair is resampled accordingly and refined with the reverse process~\cite{ddim} under the symmetric inpainting mask \(\mathbf{M}\), while preserving the \(M_i=0\) frames from the resampled duration-adjusted input. Meanwhile, \(\mathcal{D}_{\mathrm{sent}}\) predicts a sentence-level duration plan \(\tilde{\mathbf{T}}^{\mathrm{sent}}\). The refined pairwise outputs are split into gloss segments, placed on the global timeline, and cosine-fused where intermediate glosses overlap. Finally, the stitched sequence is rescaled gloss-wise to match \(\tilde{\mathbf{T}}^{\mathrm{sent}}\), resulting in \(\hat{\mathbf{X}}^{\mathrm{sent}} \in \mathbb{R}^{\mathbf{T}^{\mathrm{sent}}\times D}\).

\subsection{\textsc{SignaVox-U}}
\label{sec:signavoxu}
\noindent \textbf{Spoken Language to Gloss.} 
% To build \textsc{SignaVox-U}, we require a spoken language-to-gloss converter that generalizes reliably to unseen utterances. To this end, we adopt an LLM-based spoken language-to-gloss approach following prior work~\cite{signllm, nms_generation, pseudogloss}. In addition, to ensure notation consistency across the dataset, we use the SignStream~\cite{signstream3} annotation convention adopted by ASLLRP~\cite{asllrp} as our canonical standard. Details are provided in the Appendix.
% To encourage the LLM to explicitly learn and adhere to this convention, we employ a retrieval-augmented generation (RAG) strategy with a translation-memory prompt. Specifically, we construct a translation memory by randomly sampling 100 examples from a retrieval split. Following prior work~\cite{nms_generation}, we further apply NER-based anonymization during retrieval to mitigate cases in which proper nouns distort similarity estimation.
To build \textsc{SignaVox-U}, we employ an LLM-based spoken language-to-gloss converter, following prior works~\cite{signllm, nms_generation, pseudogloss}. To ensure notation consistency, we adopt the SignStream annotation convention~\cite{signstream3} used in ASLLRP~\cite{asllrp} as the canonical standard, and guide the LLM with a translation-memory-based RAG strategy. During retrieval, we apply NER-based anonymization to reduce similarity distortions caused by proper nouns.

Retrieval is performed in two stages. First, we compute BM25~\cite{bm25} and SPLADE~\cite{splade} scores, and combine them into a hybrid retrieval score after min-max normalization, using weights {\small $\alpha$} and {\small $1-\alpha$}, respectively. Next, we rerank the top candidates using a cross-encoder~\cite{reranker}, and combine the reranking score with the hybrid retrieval score to obtain the final retrieval score. Finally, we select the top-6 examples and incorporate them into the translation-memory prompt, enabling the LLM to generate gloss sequences that better conform to the target annotation convention.

% Given an input sentence $S$ and a candidate example $C$ from the retrieval memory, we perform retrieval in two stages. In the first stage, we compute BM25 scores~\cite{bm25} over word-level tokens and SPLADE~\cite{splade} scores using a sparse masked language model encoder. We then min--max normalize the two scores and combine them into a hybrid retrieval score, where the BM25 and SPLADE scores are weighted by $\alpha$ and $1-\alpha$, respectively. In our experiments, we set $\alpha=0.65$.
% In the second stage, we rerank the top-$K$ candidates using a cross-encoder~\cite{reranker}. We then compute the final retrieval score by combining the cross-encoder score with the hybrid retrieval score using a weighting factor $\lambda$. Finally, we select the top-6 examples according to the final retrieval score and incorporate them into the translation-memory prompt, enabling the LLM to generate gloss sequences that better conform to the target annotation convention.

\noindent \textbf{Sign Language Conversational Dataset.}
To build \textsc{SignaVox-U}, our sign language conversational dataset, we convert existing spoken-language daily conversation datasets~\cite{realtalk, everydayconversations, dailydialog} into sign language. As illustrated in \namefig{}~\ref{fig:data_pipeline}-(b), this construction pipeline combines a spoken-language-to-gloss translation approach with our developed \emph{BRAID} to transform full conversations into sign sequences. Each conversation is ultimately represented as a sequence of 3D features, allowing us to construct a large-scale sign language conversational dataset. The overall statistics of \textsc{SignaVox-U} are summarized in Table~\ref{tab:comp_dataset}, and details can be found in Appendix~\ref{sec:supp_dataset}.

%% file: tables/comparison_dataset.tex
\begin{table*}[t]
    \centering
    \caption{Comparison of existing ASL datasets and our \textsc{SignaVox} datasets. In the ``Type'' column, ``I'' and ``C'' denote isolated signing and continuous signing, respectively. Durations marked with ${}^{*}$ are estimated by converting the reported total number of frames at 25 fps.}
    \label{tab:comp_dataset}
    \resizebox{\textwidth}{!}{%
    \begin{tabular}{cccccccc}
        \specialrule{1pt}{1pt}{1pt}
        \textbf{Dataset} & \textbf{Vocab.} & \textbf{\# Hours} & \textbf{Type} & \textbf{Gloss} & \textbf{Text} & \textbf{Pseudo Annotation} & \textbf{Level} \\
        \specialrule{0.5pt}{1pt}{1pt}
        MS-ASL~\cite{msasl} & 1K & - & I & \mychecking & - & - & Word \\
        WLASL~\cite{wlasl} & 2K & - & I & \mychecking & - & - & Word \\
        How2Sign~\cite{how2sign} & 16K & 79 & C & \myxmark & \mychecking & - & Sentence \\
        ASLLRP~\cite{asllrp} & 3,245 & 2.56 & C & \mychecking & \mychecking & - & Single Utterance\\
        OpenASL~\cite{openasl} & 3,228 & 288 & C & \myxmark & \mychecking & - & Sentence\\
        YoutubeASL~\cite{youtubeasl} & 60K & 984 & C & \myxmark & \mychecking & - & Sentence\\
        SignAvatar~\cite{signavatars} & - & 78.8${}^{*}$ & I / C & \myxmark & \mychecking & Body, Hand, 2D/3D Keypoints & Word / Sentence\\
        % \specialrule{0.5pt}{1pt}{1pt}
        \rowcolor{hansayellow!30} \textsc{SignaVox-W} & 42K & 79.27 & I & \mychecking & \mychecking & Body, Hand, Face & Word\\
        \rowcolor{hansayellow!30} \textsc{SignaVox-U} & 22.6K &  336.81${}^{*}$ & C & \mychecking & \mychecking & Body, Hand, Face & Dialogue\\
        \specialrule{0.8pt}{1pt}{1pt}
    \end{tabular}%
    }
    \vspace{-3pt}
\end{table*}

%% file: sec/04ConversationModel.tex
\section{Sign Language Conversational Model}
We introduce \textsc{SignaVox}, a sign-to-sign conversational model, trained on \textsc{SignaVox-U}. Details are provided in Appendix~\ref{sec:sign2sign_model}.

\noindent \textbf{Architecture.}
Inspired by MAR~\cite{mar}, \textsc{SignaVox} autoregressively generates sign responses as fixed-length 3D motion blocks.
Given a sign history {\small$H_t$} and a target role prompt {\small$r_t$}, the response {\small$Y_t=[B_1,\ldots,B_L]$}, with each {\small$B_\ell$} containing {\small$K$} frames, is factorized as
\begin{equation}\small
p(Y_t \mid H_t,r_t)
=
\prod_{\ell=1}^{L}
p(B_\ell \mid H_t,r_t,B_{<\ell}),
\end{equation}
where {\small$B_\ell$} denotes the {\small$\ell$}-th motion block.
The history {\small$H_t$} is serialized with role and boundary tokens in a ChatML-style layout.

A Transformer decoder maps the serialized history and previous blocks to a conditioning memory,
{\small$C_\ell = f_{\theta}^{\mathrm{dec}}(H_t, r_t, B_{<\ell})$}, which conditions the flow heads for the next block.
The next block is generated by anatomy-factorized conditional flow-matching heads for the body, face, and hands~\cite{flowmatching,conditional_flowmatching}, with a higher-capacity hand head to better model lexical articulation.

To model response structure, \textsc{SignaVox} predicts a boundary state {\small$b_\ell$} for each block, indicating whether it terminates a sentence, terminates a turn, or contains no boundary.
This prediction is distinct from the boundary tokens used for input serialization.
For gloss-level semantic planning, \textsc{SignaVox} includes an auxiliary branch that predicts a gloss distribution {\small$P_{\ell}^{\mathrm{plan}}$} from {\small$C_\ell$}, without taking gloss sequences as input.
We augment the decoder memory with these planning signals:
\begin{equation}\small
\tilde{C}_{\ell}
=
C_{\ell}
+
e_{\mathrm{bdry}}(b_{\ell})
+
\alpha_s \phi_{\mathrm{plan}}(P_{\ell}^{\mathrm{plan}}),
\end{equation}
where {\small$e_{\mathrm{bdry}}$} is a learned boundary-state embedding, {\small$\phi_{\mathrm{plan}}$} maps the soft gloss-plan distribution to the decoder hidden space, and {\small$\alpha_s$} is a curriculum weight.
The flow heads are conditioned on {\small$\tilde{C}_{\ell}$} to generate the next block.

\noindent \textbf{Training Objectives.}
We train \textsc{SignaVox} with objectives for motion generation, response structure, and gloss-level semantic planning.
The main objective is the conditional flow-matching loss {\small$\mathcal{L}_{\mathrm{FM}}$}, computed over the body, face, and hand heads.
For response structure, {\small$\mathcal{L}_{\mathrm{bdry}}$} supervises sentence-end and turn-end prediction, with a lightweight calibration term for boundary frequency.
For gloss-level semantic planning, {\small$\mathcal{L}_{\mathrm{plan}}$} provides auxiliary CTC supervision by predicting the training gloss sequence from the decoder state prior to motion generation~\cite{ctcloss,slt_ctc}.
The resulting soft gloss distribution is injected into the flow memory through a delayed plan-to-flow curriculum.
This lets the flow heads first learn stable motion dynamics and then gradually incorporate gloss-level guidance.
We further use weaker post-motion CTC and landmark-based gloss losses, {\small$\mathcal{L}_{\mathrm{post}}$} and {\small$\mathcal{L}_{\mathrm{lm}}$}, as auxiliary regularizers for lexical organization.
The full objective is
\begin{equation}\small
\mathcal{L}_{\mathrm{signavox}}
=
\mathcal{L}_{\mathrm{FM}}
+
\lambda_{\mathrm{bdry}}\mathcal{L}_{\mathrm{bdry}}
+
\lambda_{\mathrm{plan}}\mathcal{L}_{\mathrm{plan}}
+
\lambda_{\mathrm{post}}\mathcal{L}_{\mathrm{post}}
+
\lambda_{\mathrm{lm}}\mathcal{L}_{\mathrm{lm}}.
\end{equation}

\noindent \textbf{Inference.}
At inference time, \textsc{SignaVox} is conditioned only on the prior signing context and target role prompt, {\small$(H_t,r_t)$}, and generates the response autoregressively as motion blocks.
For each block {\small$\ell$}, the decoder produces {\small$C_\ell$}, the boundary head predicts {\small$b_\ell$}, and the gloss planner predicts an internal soft plan {\small$P_\ell^{\mathrm{plan}}$}.
The flow heads are conditioned on {\small $\tilde{C}_\ell=C_\ell+e_{\mathrm{bdry}}(b_\ell)+\alpha \phi(P_\ell^{\mathrm{plan}})$}, where {\small$\alpha$} is the final plan-to-flow weight. The gloss plan is therefore an internal prediction, rather than an externally supplied generation input.

%% file: sec/06Experiments.tex
% \input{tables/coarticulation}
% \begin{figure}
%     \centering
%     \includegraphics[width=0.7\linewidth]{figures/braid_qual3.pdf}
%     \caption{Qualitative results of \emph{BRAID}. We visualize the synthesized motions between two isolated glosses, "fs-AUDIO" (blue) and "fs-VOCAL" (green). The 'X' marks denote the absence of intermediate frames in the transition sequence.}
%     \label{fig:braid_qualitative_result}
% \end{figure}
\section{Experiments}
\input{tables/coarticulation}

\subsection{Isolated to Continuous Signing}

\noindent
\begin{wrapfigure}[18]{r}{0.5\linewidth}
    \vspace{-15pt}
    \centering
    \includegraphics[width=\linewidth]{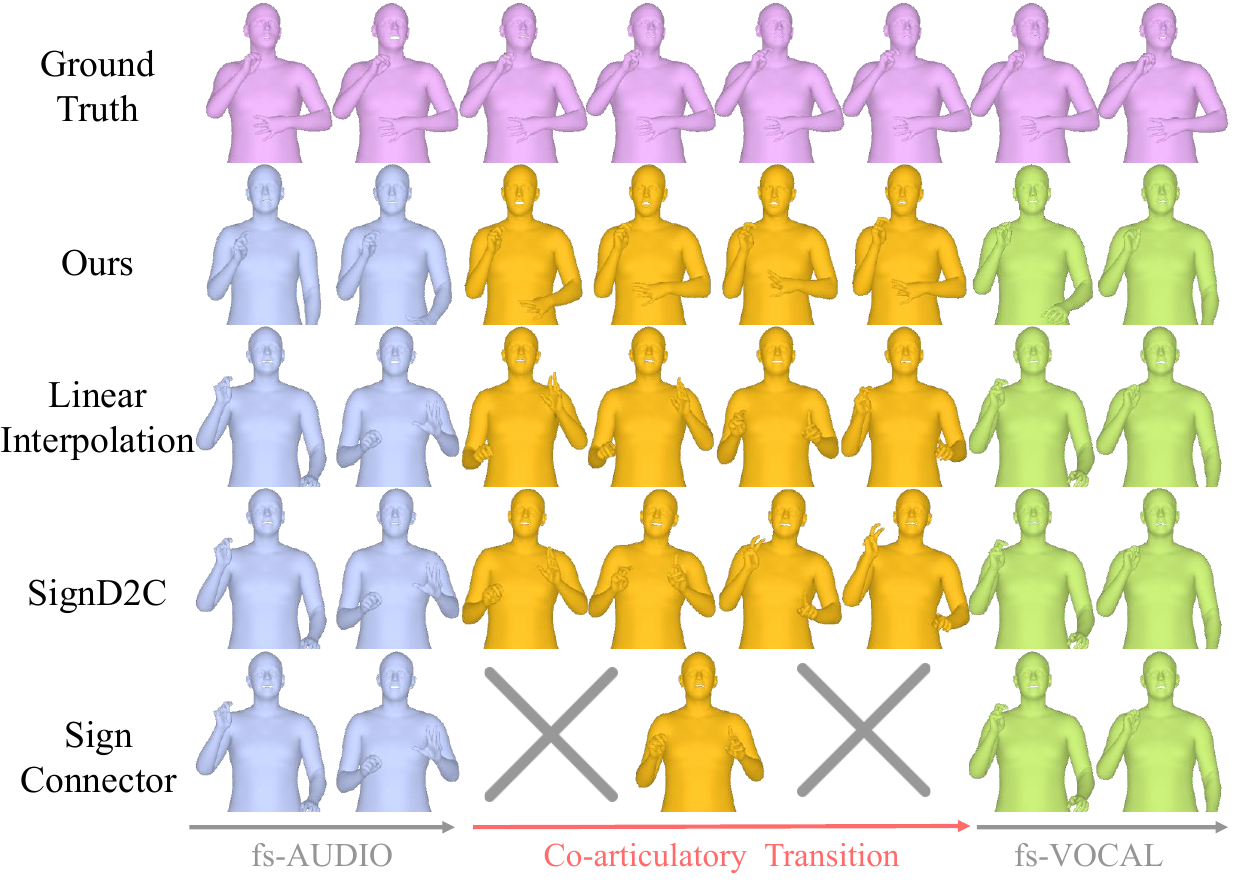}
    \captionsetup{skip=2pt}
    \caption{Qualitative results of \emph{BRAID}. We visualize the synthesized motions between two isolated glosses, "fs-AUDIO" (\textcolor{blue}{blue}) and "fs-VOCAL" (\textcolor{green}{green}). The 'X' marks denote the absence of intermediate frames in the transition sequence.}
    \label{fig:braid_qualitative_result}
    % \vspace{-10pt}
\end{wrapfigure}
\textbf{Setup.}
For gloss-level evaluation, we train the model on 78,316 samples and evaluate it on 8,274 test samples.
For sentence-level evaluation, we use 1,526 sentence sequences. We compare our method with a simple linear interpolation baseline and representative sign transition baselines~\cite{simple_baseline,signd2c}. For a controlled comparison, both the linear interpolation baseline and SignD2C~\cite{signd2c} generate only four transition frames at each gloss boundary. At the sentence-level, their gloss-pair outputs are stitched using the same procedure as ours, without the sentence-level duration predictor {\small $\mathcal{D}_{\mathrm{sent}}$}.
 
\noindent \textbf{Evaluation Metrics.}
For quantitative evaluation, we follow previous works~\cite{signavatars, slrtp_challenge} and apply dynamic time splitting (DTW) before computing pose and mesh errors. 
We report MPJPE/MPVPE and PA-MPJPE/PA-MPVPE, where Procrustes alignment accounts for global pose, scale, and signer-shape differences. Since DTW can absorb duration mismatches, we also report length ratio for sequence-length accuracy.
% We report MPJPE and MPVPE, as well as their Procrustes-aligned variants, PA-MPJPE and PA-MPVPE, to account for global pose, scale, and signer-shape differences. Since DTW can absorb duration mismatches, we also report length ratio to evaluate sequence-length accuracy.

\input{tables/frame_selection_braid}
\noindent \textbf{Results.}
Table~\ref{tab:braid_quan_results} shows that \emph{BRAID} achieves the best overall performance. It consistently reduces DTW-based motion errors over the baselines, indicating that boundary-aware refinement improves co-articulatory motion beyond simple local interpolation. Its length ratio is also closer to 1, suggesting better temporal-scale preservation.
\namefig{}~\ref{fig:braid_qualitative_result} shows that \emph{BRAID} generates more natural co-articulatory transitions than the baselines. Since linear interpolation simply connects the boundary poses of adjacent glosses, large pose discrepancies often produce unnatural articulation and static intermediate poses. SignD2C also generates short transition frames, but it tends to rely strongly on boundary anchors, pulling the motion toward the anchor poses. SignConnector is further limited to interpolating a short connector segment between adjacent signs, which restricts its ability to model broader transition dynamics. In contrast, \emph{BRAID} performs non-rigid boundary refinement after duration-aware resampling, softly adjusting frames around the boundary rather than treating anchors as hard constraints. This enables \emph{BRAID} to synthesize plausible intermediate articulation continuous motion closer to the ground truth.

\noindent \textbf{Ablation Study.} 
As shown in Table~\ref{tab:ablation_frame_selection_braid}, performance improves progressively from the raw input to the single-stage variants and further to the full pipeline.
This suggests that articulation-frame selection provides cleaner inputs for \emph{BRAID}, with motion-based selection and VideoLLM refinement playing complementary roles.

\input{tables/t2g_quan}
\input{tables/ablation_braid}
Table~\ref{tab:ablation_braid} further ablates the main components of \emph{BRAID}. At the gloss-pair level, removing boundary inpainting or gloss-level duration prediction degrades reconstruction accuracy, confirming the need for boundary-focused refinement and duration-aware resampling. At the sentence level, removing sentence-level duration prediction causes large length mismatch, and hard stitching increases motion error, showing that duration planning and smooth stitching are necessary for temporally aligned continuous sign motion.

\subsection{Spoken Language to Gloss}
% \noindent \textbf{Experimental Setup.}
% We evaluate the gloss translator on 1,261 spoken language--gloss pairs, using 100 examples as the retrieval set. 
% Finger-spelled expressions are normalized into single lexical tokens before evaluation, and the full evaluation protocol is provided in Appendix~\ref{app:gloss_eval}.

% \noindent \textbf{Experimental Setup.}
% To evaluate the performance of the gloss translator, we construct a retrieval set of 100 examples and conduct experiments on the remaining 1,261 spoken language–gloss pairs.
% For a more appropriate quantitative assessment of semantic fidelity and grammatical correctness, finger-spelled expressions are normalized into a single lexical token prior to evaluation (\textit{e.g.} ns-fs-PARIS $\rightarrow$ PARIS). Since the official implementation of the previous work~\cite{nms_generation} is not publicly available, we reproduced its methodology based on the descriptions provided in the work. We will publicly release both the evaluation set and the retrieval set.
% \input{tables/ablation_braid_data_chu} 

\noindent \textbf{Evaluation Metrics.}
We evaluate gloss translation using standard translation metrics~\cite{bleu, scarebleu, chrf} and GPT-5.2-based evaluation~\cite{openai_gpt52}. 
To complement overlap-based metrics, GPT-5.2 assesses semantic preservation and structural well-formedness. 
The evaluation prompt is provided in Appendix~\ref{sec:suppl_llm_prompt}.

\noindent \textbf{Results.}
Table~\ref{tab:quan_translator} shows that our translator outperforms the previous approach~\cite{nms_generation} across all metrics. 
Ablations confirm the importance of guidance, with retrieval and reranking providing complementary gains. 
As shown in \namefig{}~\ref{fig:qual_translator}, our translator better follows ASL-style structure, including dataset-style indices and appropriate \texttt{WH} placement. 
Some dataset-specific meta glosses, such as \texttt{ATTENTION-WAVE}, encode visual or discourse cues unavailable from text alone, which can lower n-gram scores despite reasonable outputs. 
More results are provided in Appendix~\ref{sec:suppl_experiments_t2g}.

\subsection{Sign Language Conversation}

\noindent 
\begin{wraptable}[8]{r}{0.48\linewidth}
\vspace{-14pt}
\centering
\caption{Performance comparison with~\cite{remos} and ablation studies.}
\label{tab:signavox_quan}
\renewcommand{\arraystretch}{1.0}
\vspace{-8pt}
\resizebox{\linewidth}{!}{%
\begin{tabular}{lcccc}
\specialrule{1pt}{1pt}{1pt}
Method 
& FGD $\downarrow$ 
& \shortstack{DTW\\MPJPE $\downarrow$} 
& \shortstack{DTW\\MPVPE $\downarrow$} 
& BLEU-4 $\uparrow$ \\
\specialrule{0.5pt}{1pt}{1pt}
ReMoS~\cite{remos} & 83.281 & 0.1105 & 0.0788 & 0.0919 \\ 
\hdashline
\rowcolor{hansayellow!30} 
\textsc{SignaVox} & \textbf{2.9497} & \textbf{0.0800} & \textbf{0.0550} & \textbf{0.0944} \\
\Lbranch w/o $\alpha_s\phi(P^{\mathrm{plan}})$ & 4.5312 & 0.0821 & 0.0568 & 0.0900 \\
\Lbranch w/o $\mathcal{L}_{\mathrm{plan}}$ & 2.9683 & 0.0818 & 0.0565 & 0.0850 \\
\Lbranch w/o $\mathcal{L}_{\mathrm{post}}+\mathcal{L}_{\mathrm{lm}}$ & 4.1140 & 0.0868 & 0.0565 & 0.0720 \\
\Lbranch w/o semantic grounding & 8.2454 & 0.0810 & 0.0565 & 0.0520 \\
\specialrule{0.8pt}{1pt}{1pt}
\end{tabular}%
}
% \vspace{-10pt}
\end{wraptable}
\textbf{Evaluation Metrics.} To evaluate our model, we compare generated response motions with the corresponding ground-truth responses using DTW-MPJPE and DTW-MPVPE. We additionally report Fréchet Gesture Distance (FGD)~\cite{fgd} for distributional motion quality.
For response-level semantic alignment, we propose a retrieval-based semantic evaluation method, detailed in Appendix~\ref{sec:retrieval_based_translation}.

\noindent \textbf{Results.}
Table~\ref{tab:signavox_quan} compares our model with ReMoS~\cite{remos}, a motion-conditioned response generation baseline, and reports ablation results.
\textsc{SignaVox} consistently outperforms ReMoS across motion quality and semantic alignment metrics.
The ablations show that each grounding component contributes to response generation: {\small $\alpha_s\phi(P_{\ell}^{\mathrm{plan}})$} links the learned gloss plan to the motion generator, {\small $\mathcal{L}_{\mathrm{plan}}$} supervises pre-motion gloss planning, and {\small $\mathcal{L}_{\mathrm{post}}+\mathcal{L}_{\mathrm{lm}}$} supports semantic consistency.
Removing all three components results in the largest BLEU-4 drop, indicating that semantic grounding is important for preserving the intended response semantics.

%% file: tables/coarticulation.tex
% \begin{table}[t]
% \centering
% \resizebox{\textwidth}{!}{%
% \begin{tabular}{@{}p{0.35cm} l|cccc@{}}
% \toprule
% & Method & DTW-MPJPE & DTW-PA-MPJPE & DTW-MPVPE & DTW-PA-MPVPE \\ 
% \midrule
% \multirow{4}{*}{\rotatebox{90}{\textbf{Gloss}}}
% & Linear ($r=4$) & 0.0997 & 0.0803 & 0.0713 & 0.0616 \\
% & SignConnector~\cite{simple_baseline} & 0.0956 & 0.0796 & 0.0673 & 0.0597 \\
% & SignD2C~\cite{signd2c} & 0.0980 & 0.0764 & 0.0739 & 0.0593 \\
% & Ours & \textbf{0.0787} & \textbf{0.0642} & \textbf{0.0569} & \textbf{0.0490} \\
% \hdashline
% \multirow{4}{*}[-0.3em]{\rotatebox{90}{\textbf{Sentence}}}
% & Linear ($r=4$) & 0.1068 & 0.0921 & 0.0750 & 0.0678 \\
% & SignConnector~\cite{simple_baseline} & 0.1032 & 0.0909 & 0.0723 & 0.0669 \\
% & SignD2C~\cite{signd2c} & 0.1003 & 0.0832 & 0.0723 & 0.0625 \\
% & Ours & \textbf{0.0863} & \textbf{0.0770} & \textbf{0.0600} & \textbf{0.0551} \\
% \bottomrule
% \end{tabular}%
% }
% \caption{Isolated-to-continuous results.}
% \end{table}
\begin{table}[t]
\centering
\caption{Quantitative comparison of gloss-level and sentence-level co-articulation generation results. For methods with {\small $r=4$}, four co-articulation frames are predicted. Lower is better for all metrics, while Length Ratio is better when closer to 1.}
\label{tab:braid_quan_results}
\resizebox{0.98\textwidth}{!}{%
\begin{tabular}{@{}c lcccccccccc@{}}
\specialrule{1pt}{1pt}{1pt}
&
& \multicolumn{3}{c}{DTW-MPJPE $\downarrow$}
& \multicolumn{4}{c}{DTW-MPVPE $\downarrow$}
& \multirow{2}{*}[-0.3em]{\makecell[c]{DTW\\PA-MPJPE $\downarrow$}}
& \multirow{2}{*}[-0.3em]{\makecell[c]{DTW\\PA-MPVPE $\downarrow$}}
& \multirow{2}{*}[-0.3em]{\makecell[c]{Length\\Ratio}} \\
\cmidrule(lr){3-5} \cmidrule(lr){6-9}
&
& Body & Hands & Overall
& Body & Hands & Face & Overall
& & & \\
\specialrule{0.5pt}{1pt}{1pt}

\multirow{4}{*}{\rotatebox{90}{\textbf{Gloss}}}
& Linear ($r=4$) &0.0460& 0.1841 & 0.0816 &0.0358&0.0231&0.0018& 0.0538& 0.0769 & 0.0543 &5.3693\\
& SignConnector~\cite{simple_baseline}
                 &0.0461 &0.1843& 0.0817 &0.0359&0.0231&0.0018& 0.0539 & 0.0770 & 0.0544 &5.0981\\
& SignD2C ($r=4$)~\cite{signd2c}
                 &0.0466&0.1827& 0.0819 &0.0368& 0.0233& 0.0018 & 0.0559 & 0.0766 & 0.0542 & 5.3693 \\
\rowcolor{hansayellow!20} \cellcolor{white} & \emph{BRAID} (Ours)    &\textbf{0.0376}&\textbf{0.1384} & \textbf{0.0653} &\textbf{0.0309} &\textbf{0.0171}&\textbf{0.0017}& \textbf{0.0439} & \textbf{0.0610} & \textbf{0.0427} & \textbf{1.2786}\\
% \hdashline
& Linear ($r=4$) &0.0504&0.2053& 0.1020 &0.0537&0.0494&0.0243& 0.0706 & 0.0898 & 0.0651 &1.6290\\
& SignConnector~\cite{simple_baseline}
                 &0.0508&0.2074& 0.1028 &0.0496&0.0246& 0.0227 & 0.0711 & 0.0906 & 0.0657 & 1.5339 \\
& SignD2C ($r=4$)~\cite{signd2c}
                 &0.0515&0.2055& 0.1028 &0.0504&0.0246&0.0226& 0.0728 & 0.0900 & 0.0655 &1.6290\\
\rowcolor{hansayellow!20} 
\cellcolor{white} \multirow{-4}{*}{\rotatebox{90}{\textbf{Sentence}}} &  \emph{BRAID} (Ours)           &\textbf{0.0422}&\textbf{0.1686}& \textbf{0.0757} & \textbf{0.0336} &\textbf{0.0190}&\textbf{0.0015}& \textbf{0.0498} & \textbf{0.0758} & \textbf{0.0507} & \textbf{1.0203} \\
\specialrule{0.8pt}{1pt}{1pt}
\end{tabular}%
}
% \captionsetup{skip=4pt}
\vspace{-13pt}
\end{table}

%% file: tables/frame_selection_braid.tex
% \newcommand{\Lbranch}{\rule[0.1ex]{0.5pt}{1.2ex}\rule[0.1ex]{0.7em}{0.5pt}\,}

% frame selection braid
\begin{wraptable}{r}{0.45\textwidth}
    \centering
    \vspace{-10pt}
    \caption{Ablation on articulation-frame selection. ``G'' and ``S'' denote gloss-pair and sentence levels, respectively.}
    \label{tab:ablation_frame_selection_braid}
    \resizebox{\linewidth}{!}{%

    \setlength{\aboverulesep}{0pt}
    \setlength{\belowrulesep}{0pt}
    \renewcommand{\arraystretch}{1.3}
    \vspace{-10pt}
    \begin{tabular}{lcccc}
    \specialrule{1pt}{1pt}{1pt}
    & \multicolumn{2}{c}{DTW-MPJPE $\downarrow$} & \multicolumn{2}{c}{DTW-MPVPE $\downarrow$} \\
    \cmidrule(lr){2-3} \cmidrule(lr){4-5}
    & G & S & G & S \\
    \specialrule{0.5pt}{1pt}{1pt}

    % \multicolumn{5}{>{\columncolor[gray]{0.9}[\tabcolsep]}l}{\textbf{\textit{Data curation}}} \\
    Raw & 0.0704 & 0.0790 & 0.0478 & 0.0526 \\
    Motion-only& 0.0693 & 0.0783 & 0.0472 & 0.0523 \\ % \Lbranch
    VideoLLM-only & 0.0689 & 0.0784 & 0.0471 & 0.0524 \\ % \Lbranch 
    \rowcolor{hansayellow!30} \textbf{Motion + VideoLLM} & \textbf{0.0653} & \textbf{0.0738} & \textbf{0.0439} & \textbf{0.0486} \\

    \specialrule{0.8pt}{1pt}{1pt}
    \end{tabular}
    }
    \vspace{-10pt}
\end{wraptable}

% 1. 수직선이 이어지는 중간 항목용 (|- 모양)
% \newcommand{\Ibranch}{%
%   \makebox[0pt][l]{\rule[-0.6em]{0.5pt}{1.7em}}% 아래로 뻗는 수직선 (길이 조절됨)
%   \rule[0.1ex]{0.5pt}{1.2ex}\rule[0.1ex]{0.7em}{0.5pt}\,% 가로 틱
% }

% \newcommand{\LLbranch}{%
%   \rule[0.1ex]{0.5pt}{1.9ex}\rule[0.1ex]{0.7em}{0.5pt}\,% 여기서 선이 끝남
% }

% % 2. 계층 구조의 마지막 항목용 (|_ 모양)
% \newcommand{\Lbranch}{%
%   \rule[0.1ex]{0.5pt}{1.2ex}\rule[0.1ex]{0.7em}{0.5pt}\,% 여기서 선이 끝남
% }

% \begin{wraptable}{r}{0.45\textwidth}
%     \centering
%     \vspace{-10pt}
%     \caption{Ablation on Gloss and Inpainting.}
%     \vspace{-4pt}
%     \resizebox{\linewidth}{!}{%
%     \setlength{\aboverulesep}{0pt}
%     \setlength{\belowrulesep}{0pt}
%     \renewcommand{\arraystretch}{1.2}
%     \begin{tabular}{llcc}
%     \specialrule{0.8pt}{1pt}{1pt}
%     &  & DTW-MPJPE $\downarrow$ & DTW-MPVPE $\downarrow$ \\
%     \specialrule{0.5pt}{1pt}{1pt}
    
%     \multirow{7}{*}{\rotatebox{90}{\textbf{Gloss}}} 
%     & \multicolumn{3}{>{\columncolor[gray]{0.9}[\tabcolsep]}l}{\textit{Data curation}} \\ 
%     & No F-S                       & 0.0704 / 0.0704 & 0.0478 / 0.0478 \\
%     & \Ibranch w/ motion           & 0.0693 & 0.0472 \\ % 이어지는 중간 선
%     & \LLbranch w/ llm              & 0.0689 & 0.0471 \\ % 마지막 선
%     \cmidrule(lr){2-4}
    
%     & \multicolumn{3}{>{\columncolor[gray]{0.9}[\tabcolsep]}l}{\textit{Training}} \\
%     & F-S (w/o inpaint)           & 0.0670 & 0.0462 \\
%     & \Lbranch w/ inpaint (ours)  & \textbf{0.0653} & \textbf{0.0486} \\
    
%     \specialrule{0.8pt}{1pt}{1pt}
%     \end{tabular}
%     }
%     \label{tab:ablation_gloss}
%     \vspace{-10pt}
% \end{wraptable}

%% file: tables/t2g_quan.tex
\begin{figure}[t]
    \centering
    \begin{minipage}[t]{0.43\columnwidth}
        \centering
        \captionof{table}{Quantitative results of spoken-language-to-gloss translation. We set {\small $N=100$} following previous work~\cite{nms_generation}.}
        \label{tab:quan_translator}
        \vspace{-7pt}
        \resizebox{\linewidth}{!}{%
        \begin{tabular}{c!{\vrule}ccccc}
            \specialrule{1pt}{1pt}{1pt}
            & BLEU-4 & SacreBLEU & chrF & \makecell{GPT\\Struct} & \makecell{GPT\\Semantic} \\
            % \hline\hline
            \specialrule{0.5pt}{1pt}{1pt}
            % \specialrule{0.5pt}{1pt}{1pt}

            $N$-shot~\cite{nms_generation}
            & 0.094
            & 15.583 & 51.701 & 3.674 & 4.308 \\ \hdashline
            No guidance
            & 0.021
            & 8.581 & 42.818 & 3.289 & 4.083 \\
            Retrieval-only % \textit{w/} 
            & 0.098
            & 16.512 & 55.176 & 3.692 & 4.331 \\
            Rerank-only % \textit{w/} 
            & 0.099
            & 16.506 & 54.912 & 3.707 & 4.332 \\
           \rowcolor{hansayellow!30}  Ours
            & \textbf{0.108}  & \textbf{17.062} & \textbf{55.645} & \textbf{3.751} & \textbf{4.351} \\
            \specialrule{0.8pt}{1pt}{1pt}
        \end{tabular}%
        }
    \end{minipage}\hfill
    \begin{minipage}[t]{0.55\columnwidth}
        \centering
        \vspace{0pt}
        \includegraphics[width=\linewidth]{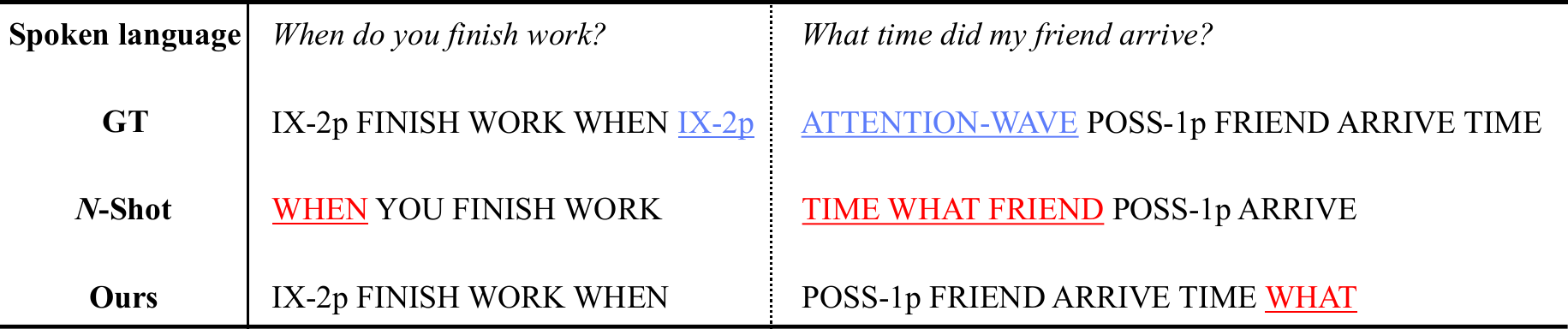}
        \captionof{figure}{Qualitative examples of spoken-language-to-gloss translation. \textcolor{blue}{Blue} omitted reference glosses, and \textcolor{red}{red} marks erroneous generated glosses.}
        \label{fig:qual_translator}
    \end{minipage}
    \vspace{-10pt}
\end{figure}

%% file: tables/ablation_braid.tex
\begin{wraptable}{r}{0.35\linewidth}
    % \vspace{-12pt}
    \centering
    \caption{Ablation study on duration prediction and boundary inpainting.}
    \label{tab:ablation_braid}
    \begingroup
    % \scriptsize
    % \tiny
    \fontsize{6pt}{7pt}\selectfont
    \setlength{\tabcolsep}{3.2pt}
    \setlength{\aboverulesep}{0pt}
    \setlength{\belowrulesep}{0pt}
    \renewcommand{\arraystretch}{0.92}

    \vspace{-5pt}
    \resizebox{0.98\linewidth}{!}{%
    \begin{tabular}{lccc}
    \specialrule{0.8pt}{0.5pt}{0.5pt}
    & \shortstack{DTW\\MPJPE$\downarrow$} 
    & \shortstack{DTW\\MPVPE$\downarrow$} 
    & \shortstack{Length\\Ratio} \\
    \specialrule{0.4pt}{0.5pt}{0.5pt}

    \multicolumn{4}{>{\columncolor[gray]{0.9}[\tabcolsep]}l}{\textbf{\textit{Training (Gloss-level)}}} \\
    w/o inpaint   & 0.0670 & 0.0881 & 1.2786 \\
    w/o {\tiny $\mathcal{D}_{\mathrm{gloss}}$}    & 0.0701 & 0.0480 & 1.5050 \\
    \rowcolor{hansayellow!30} Ours  & \textbf{0.0653} & \textbf{0.0439} & \textbf{1.2786} \\

    \multicolumn{4}{>{\columncolor[gray]{0.9}[\tabcolsep]}l}{\textbf{\textit{Stitching (Sentence-level)}}} \\
    w/o {\tiny $\mathcal{D}_{\mathrm{sent}}$}    & 0.0863 & 0.6000 & 0.4735 \\
    Hard Stitching         & 0.0745 & 0.0491 & 1.0203 \\
    \rowcolor{hansayellow!30} Ours             & \textbf{0.0738} & \textbf{0.0486} & \textbf{1.0203} \\

    \specialrule{0.6pt}{0.5pt}{0.5pt}
    \end{tabular}
    }
    \endgroup

    \vspace{-12pt}
\end{wraptable}

%% file: sec/07Conclusion.tex
\section{Conclusion}
We introduce a sign-centered approach for constructing and modeling continuous sign language motion. We introduced \textsc{SignaVox-W} and \textsc{SignaVox-U} as 3D signing resources, proposed \emph{BRAID} for co-articulation-aware composition of isolated signs, and further developed \textsc{SignaVox}, a sign-to-sign conversational model for generating continuous 3D sign responses. Together, these components establish an initial step toward direct sign-centered conversational modeling. Additional discussion on limitations and future work is provided in the Appendix~\ref{sec:limitation}.

%% file: suppl/Limitations.tex
\section{Limitations \& Future Works}
\label{sec:limitation}
While our framework provides a step toward sign-centered generation and interaction, several limitations remain. First, our dataset and models rely on 3D motion features extracted by existing body, hand, and face reconstruction models. As a result, the quality of the resulting motion representation is bounded by the accuracy of these estimators, particularly in challenging cases such as occlusion, fast hand motion, or low-resolution videos. Improving sign-specific 3D reconstruction remains an important direction for future work.

Second, although our representation includes facial parameters and captures limited non-manual signals, it does not fully cover the rich non-manual components of sign language, including gaze, head movement, eyebrow motion, mouthing, and other grammatical or affective cues. Future work should incorporate more expressive representations and explicit modeling of non-manual signals to better capture the linguistic structure of signing.

Third, our spoken-language-to-gloss conversion follows SignStream-style annotation conventions. This provides a consistent rule set for constructing gloss sequences, but also limits the framework to a particular glossing style. Extending the pipeline to accommodate different annotation conventions, glossing style. Extending the pipeline to accommodate different annotation conventions, sign languages, and community-specific variation is an important future direction.

Finally, evaluating generated sign conversations remains challenging. Our evaluation protocol provides useful proxies for motion quality and semantic recoverability, but it cannot fully measure linguistic correctness, non-manual expression, discourse-level coherence, or human-perceived naturalness. More comprehensive evaluation with expert signers and members of the signing community will be necessary for future sign-centered conversational systems.

%% file: suppl/Ethical_consideration.tex
\section{Ethical Consideration}
\label{sec:ethical}
Our dataset is constructed from publicly accessible sign language videos collected from the web. Since these materials contain identifiable human subjects, we carefully considered privacy, consent, and redistribution issues throughout the dataset construction process. First, we do not redistribute the original raw videos as part of the released dataset. Instead, whenever possible, we provide only links or references to the original source pages, so that access to the underlying content remains tied to the original hosting platform. This is intended to reduce unnecessary redistribution of personally identifiable visual data while preserving traceability to the original source. Second, for qualitative examples shown in the paper and appendix, we anonymize the signer by masking the face region. These visual examples are included only to illustrate dataset characteristics and model behavior, and we aim to minimize the exposure of identifiable appearance information in all presented figures. Third, because the dataset is collected from web sources, we acknowledge that public accessibility does not necessarily imply that all forms of downstream reuse carry the same expectation from the original signers. We therefore adopt a conservative release policy that focuses on processed annotations and source references rather than direct redistribution of raw visual content. We additionally emphasize that the dataset is intended for academic research purposes, and any future use should respect the terms, licenses, and access conditions associated with the original sources. Finally, we recognize that web-collected sign language data may reflect biases in source availability, recording conditions, signer demographics, and platform-specific content practices. Such factors may influence both dataset composition and downstream model behavior. We therefore view this dataset as a research resource and encourage its use with appropriate caution, particularly in settings involving human identity, privacy, or real-world deployment.

%% file: suppl/SignLanguage.tex
\section{Key Concepts in Sign Language}
\label{sec:key_concept}
\begin{itemize}
    \item \textit{Isolated sign.} A single lexical sign produced independently, typically containing handshape, hand motion, body pose, and, when relevant, facial expression.

    \item \textit{Continuous sign.} A sequence of signs produced as a continuous utterance, where adjacent signs are connected through natural transitions and co-articulation.
    
    \item \textit{Co-articulation.} The motion adaptation that occurs around the boundary between adjacent signs, reflecting how neighboring signs influence each other in timing and movement.

    \item \textit{Gloss.} A written label used to represent a sign, typically following sign language order and reflecting aspects of sign language grammar~\cite{stokoe1980:sign}.
    
    \item \textit{Articulation.} The core motion segment that realizes the lexical content of a sign, excluding preparation, retraction, or idle frames.

    \item \textit{Preparation.} The motion before the core articulation, where the signer moves toward the starting configuration of a sign.

    \item \textit{Retraction.} The motion after the core articulation, where the signer moves away from the completed sign.

    \item \textit{Gloss sequence.} An ordered sequence of gloss labels representing a continuous sign utterance. In our system, it serves as an intermediate representation between spoken language and 3D sign motion.

    \item \textit{Spoken language.} The text sentence associated with a sign utterance. It is not necessarily word-aligned with the gloss sequence, since sign languages and spoken languages can differ in word order and grammatical structure.
\end{itemize}

%% file: suppl/Details_of_dataset.tex
\section{Details of Dataset}
\label{sec:supp_dataset}
\subsection{Data Analysis}
\subsubsection{Statistics}
\input{tables/data_source_stat.tex}
\input{tables/data_source_signavoxu.tex}
The statistics of the processed source datasets used to construct \textsc{SignaVox-W} are presented in Table~\ref{tab:data_source_stat}. As shown in the table, the source datasets vary substantially in scale, annotation type, resolution, and video length. In particular, while SignASL~\cite{signasl} and Spreadthesign~\cite{spreadthesign} provide large vocabularies and broad lexical coverage, other resources such as MS-ASL~\cite{msasl} and WLASL~\cite{wlasl} offer complementary samples with different annotation schemes and video characteristics. This diversity allows us to collect lexical sign videos from heterogeneous sources, although it also necessitates careful quality control and normalization.

The statistics of the processed source datasets used to construct \textsc{SignaVox-U} are presented in Table~\ref{tab:dataset_stats}. As shown in the table, the source datasets differ in conversational scale, turn structure, sentence count, and resulting motion length. DailyDialog~\cite{dailydialog} provides the largest number of dialogues and sentences, offering broad coverage of everyday multi-turn interactions, while Everyday Conversations~\cite{everydayconversations} and RealTalk~\cite{realtalk} contribute complementary dialogue patterns with longer average sentence-level motion sequences. After spoken-language-to-gloss conversion and isolated-to-continuous construction, \textsc{SignaVox-U} contains 10,728 training dialogues, 81,663 training turns, and 133,222 training sentences, with an average of 191.8 frames per sentence. This scale and diversity allow us to construct a dialogue-style continuous signing dataset suitable for training sign-centered conversational models.

% \subsubsection{Data Distribution}
% \textcolor{red}{CHOO}
\subsubsection{Data Distribution of Conversational Glosses}

\begin{figure}[t]
    \centering
    \includegraphics[width=0.8\linewidth]{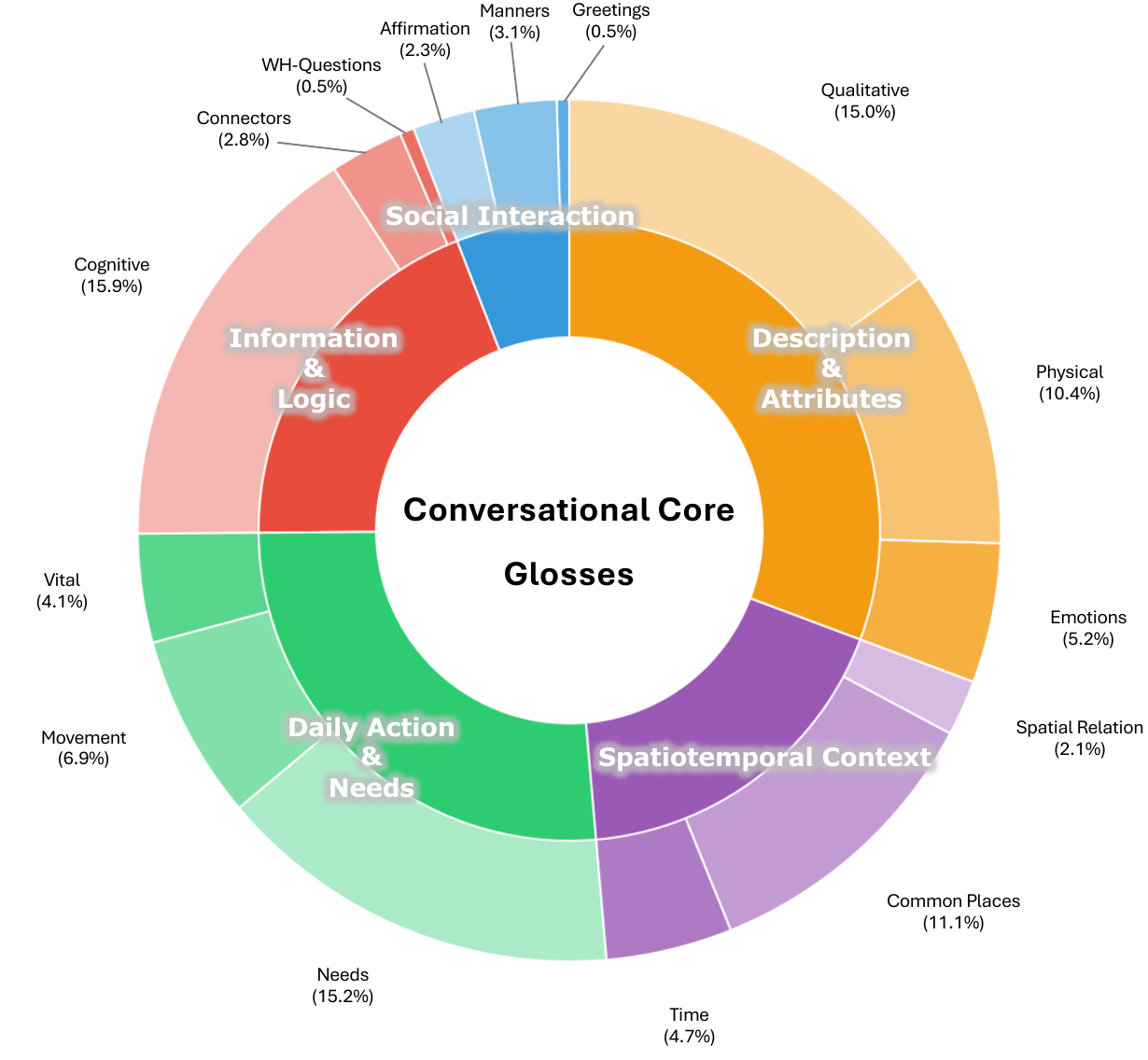}
    \caption{Semantic distribution of gloss-level videos in \textsc{SignaVox-W}.}
    \label{fig:gloss_distribution}
\end{figure}

% \begin{wrapfigure}{r}{0.45\linewidth}
%     \vspace{-10pt}
%     \centering
%     \includegraphics[width=\linewidth]{figures/gloss_distribution.pdf}
%     \vspace{-8pt}
%     \caption{Semantic distribution of gloss-level videos in \textsc{SignaVox-W}.}
%     \label{fig:gloss_distribution}
%     \vspace{-15pt}
% \end{wrapfigure}
To examine whether \textsc{SignaVox-W} provides sufficient lexical coverage for daily sign conversation, we analyze the semantic distribution of its gloss-level videos. For each video, we provide the gloss and its meaning label to GPT-4o and assign it to a taxonomy consisting of five high-level conversational groups and 15 fine-grained categories. We then count the video-level gloss instances in each category and visualize the resulting distribution in \namefig{}~\ref{fig:gloss_distribution}.

The resulting distribution shows that \textsc{SignaVox-W} covers a broad range of conversational functions 
% \begin{figure}{r}{0.45\linewidth}
%     \vspace{-1pt}
%     \centering
%     \includegraphics[width=\linewidth]{figures/gloss_distribution.pdf}
%     \vspace{-8pt}
%     \caption{Semantic distribution of gloss-level videos in \textsc{SignaVox-W}.}
%     \label{fig:gloss_distribution}
%     \vspace{-15pt}
% \end{figure}

rather than being concentrated in a narrow set of signs. The largest portion corresponds to \textit{Description \& Attributes} with 30.6\%, including qualitative descriptions, physical states, and emotions. This indicates that the dataset contains many glosses for describing people, objects, and states, which are essential for generating natural sign responses. \textit{Daily Actions \& Needs} accounts for 26.2\%, covering movement, vital actions, and requests. These signs directly support daily-life interactions and provide action-oriented expressions required in practical conversations.
The dataset also contains substantial coverage of contextual and logical expressions. \textit{Information \& Logic} accounts for 19.2\%, mainly including cognitive states, connectors, and question-related expressions, while \textit{Spatiotemporal Context} accounts for 17.9\%, including time references, common places, and spatial relations. These categories are important for constructing sentence-level sign videos, since they help combine individual glosses into coherent utterances with temporal, spatial, and causal structure.
Although \textit{Social Interaction} occupies a smaller portion at 5.9\%, this is expected given the nature of this category. Social-interaction glosses, such as greetings, manners, and affirmation or denial, form a comparatively small and closed set of conventional expressions. Once these basic interactional signs are covered, they can be repeatedly reused across different conversational contexts. In contrast, actions, attributes, places, and logical expressions are more open-ended and content-bearing, requiring broader lexical coverage to support diverse sentence construction.

Overall, this distribution suggests that \textsc{SignaVox-W} provides functionally diverse gloss-level building blocks for composing sentence-level sign videos and training a sign-to-sign conversational model.
\label{sec:data_qual_control}
\begin{figure}[b]
    \centering
    \includegraphics[width=\linewidth]{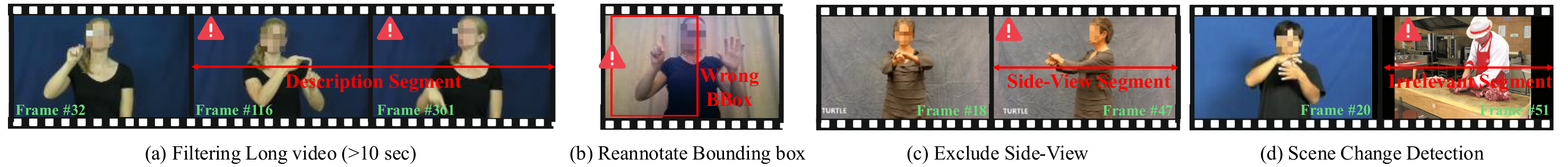}
    \caption{Examples of representative cases considered during data quality control.}
    \label{fig:quality_control}
\end{figure}

\subsection{Details of 3D Representation}
The main 3D motion representation used in our experiments follows
Equation~\ref{eq:def_feature}. 
For completeness, we also provide an augmented annotation that includes the
SMPL-X global body orientation and the FLAME neck rotation in addition to the
features in Equation~\ref{eq:def_feature}. 
For each frame \(t\), the augmented representation is defined as
\begin{equation}\small
    \boldsymbol{\Theta}^{\mathrm{body}}_t
    =
    \big[\,\mathbf{r}^{\mathrm{body}}_t,\ 
    \boldsymbol{\theta}^{\mathrm{body}}_{t}\,\big],
    \quad
    \boldsymbol{\Theta}^{\mathrm{face}}_t
    =
    \big[\,\mathbf{r}^{\mathrm{neck}}_t,\ 
    \boldsymbol{\theta}^{\mathrm{jaw}}_t,\ 
    \boldsymbol{\psi}_t\,\big],
    \quad
    \boldsymbol{\Theta}^{\mathrm{hands}}_t
    =
    \big[\,\boldsymbol{\theta}^{\mathrm{rhand}}_t,\ 
    \boldsymbol{\theta}^{\mathrm{lhand}}_t\,\big].
    \label{eq:data_features}
\end{equation}
Here, \(\mathbf{r}^{\mathrm{body}}_t\in\mathbb{R}^{3}\) denotes the global body orientation,
\(\boldsymbol{\theta}^{\mathrm{body}}_t\in\mathbb{R}^{63}\) denotes the local body joint rotations,
\(\mathbf{r}^{\mathrm{neck}}_t\in\mathbb{R}^{3}\) denotes the FLAME neck rotation,
\(\boldsymbol{\theta}^{\mathrm{jaw}}_t\in\mathbb{R}^{3}\) denotes the jaw rotation,
\(\boldsymbol{\psi}_t\in\mathbb{R}^{50}\) denotes the facial expression coefficients, and
\(\boldsymbol{\theta}^{\mathrm{rhand}}_t,\boldsymbol{\theta}^{\mathrm{lhand}}_t\in\mathbb{R}^{45}\)
denote the right and left hand joint rotations, respectively.
Thus, the augmented frame-wise feature dimension is $D=212$
Accordingly, the augmented motion sequence for gloss \(g_k\) is denoted as
\[
\mathbf{X}^{(k)}_{\mathrm{data}}
=
[\mathbf{x}^{(k)}_{1}, \cdots, \mathbf{x}^{(k)}_{T_k}]
\in \mathbb{R}^{T_k \times 212}.
\]
\subsection{Data Quality Control}
We describe in greater detail the data quality control procedure outlined in Section~\ref{sec:signavoxw}. \namefig{}~\ref{fig:quality_control} presents representative examples of the five filtering and preprocessing steps, highlighting the rationale behind each criterion and its role in the overall data curation process.
First, since each video is intended to correspond to a single lexical item, we discard samples longer than 10 seconds, as such videos often include irrelevant motions or unrelated content. As illustrated in \namefig{}~\ref{fig:quality_control}-(a), some videos contain an isolated signing of the target word followed by an explanatory segment. Since these samples do not align with our objective of collecting isolated sign clips, we remove them from the dataset.
Second, for datasets with existing signer annotations~\cite{msasl, wlasl}, we re-estimate signer bounding boxes using YOLOv8~\cite{yolov8} to replace the original YOLOv3-based annotations~\cite{yolov3} (see \namefig{}~\ref{fig:quality_control}. Samples for which reliable signer detection cannot be obtained are discarded.
Third, to ensure accurate estimation of facial expressions and hand movements, we retain only front-view frames while excluding side-view segments. (see \namefig{}~\ref{fig:quality_control}-(c). Specifically, using the notation defined in Eq.~\ref{eq:def_feature}, we discard frames satisfying both $|\mathbf{r}^{\mathrm{body}}_t|>\tau$ and $|\mathbf{r}^{\mathrm{neck}}_t|>\tau$, where $\mathbf{r}^{\mathrm{body}}_t$ and $\mathbf{r}^{\mathrm{neck}}_t$ denote the yaw angles extracted from the root body and neck rotations, respectively, via Euler-angle decomposition; we set $\tau=0.7$.
Next, we remove non-signing segments and frames in which the signer changes within a video. As shown in \namefig{}~\ref{fig:quality_control}-(d), some videos contain a trailing explanatory segment for the target word, which we exclude together with any signer-switching portions. We extract identity features using OSNet~\cite{osnet} and compute the cosine similarity with respect to a reference frame. Frames with similarity values below a predefined threshold are treated as scene transitions and discarded.
Finally, to enforce structural consistency within each gloss, we split videos into dominant and non-dominant samples. We extract SMPL-X~\cite{smpl-x} based motion sequences and compute pairwise similarity using subsequence dynamic time warping (DTW)~\cite{sub-dtw} by aligning video prefixes to candidate segments from other samples. Based on KNN average similarity scores, videos with low similarity to their neighbors are classified as non-dominant, while highly similar videos are regarded as dominant.

% Third, to ensure stable acquisition of facial expression cues critical for NMS, we retain only front-view frames and exclude side-view segments. Specifically, frames satisfying both $|\mathbf{r}^{\mathrm{body}}_t|>\tau$ and $|\mathbf{r}^{\mathrm{neck}}_t|>\tau$ are discarded, where $\mathbf{r}^{\mathrm{body}}_t$ and $\mathbf{r}^{\mathrm{head}}_t$ denote the yaw angles obtained from the root body rotation $\mathbf{r}^{\mathrm{body}}_t$ and the head rotation $\mathbf{r}^{\mathrm{neck}}_t$ via Euler-angle decomposition; we set $\tau=0.7$.
% Next, we remove frames corresponding to non-signing segments or signer identity changes within a video. We extract identity features using OSNet~\cite{osnet} and compute the cosine similarity with respect to a reference frame. Frames with similarity values below a predefined threshold are treated as scene transitions and discarded.
% Finally, to enforce structural consistency within each gloss, we split videos into dominant and non-dominant samples. We extract SMPL-X~\cite{smpl-x} based motion sequences and compute pairwise similarity using subsequence dynamic time warping (DTW)~\cite{sub-dtw} by aligning video prefixes to candidate segments from other samples. Based on KNN average similarity scores, videos with low similarity to their neighbors are classified as non-dominant, while highly similar videos are regarded as dominant.
\begin{figure}[t]
    \centering
    \includegraphics[width=\linewidth]{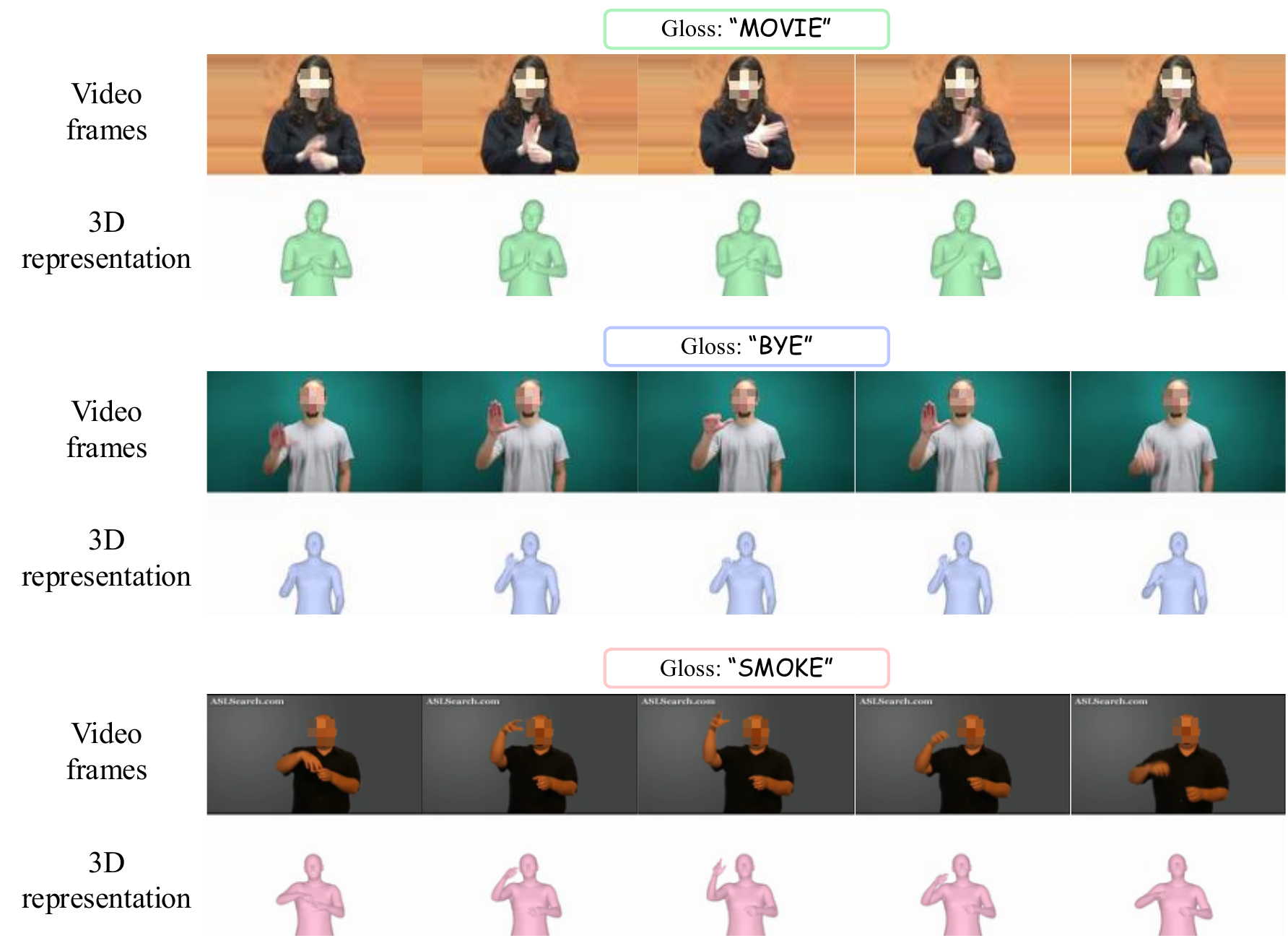}
    \caption{Visualized examples of \textsc{SignaVox-W}.}
    \label{fig:signavoxw_vis}
\end{figure}
\subsection{Data Visualization}
This section provides representative visualizations to qualitatively illustrate the structure and annotation format of the constructed dataset. \namefig{}~\ref{fig:signavoxw_vis} shows gloss-level examples of isolated sign clips in \textsc{SignaVox-W}, presenting both the original video frames and their corresponding 3D representations. For visualization, we map the frame-level motion annotations to SMPL-X meshes and render them as 3D representations. These examples show that sign clips collected from diverse sources are aligned under a unified 3D motion representation, while preserving gloss-specific hand trajectories, arm configurations, body postures, and temporal articulation patterns.

\begin{figure}[!t]
    \centering
    \includegraphics[width=\linewidth]{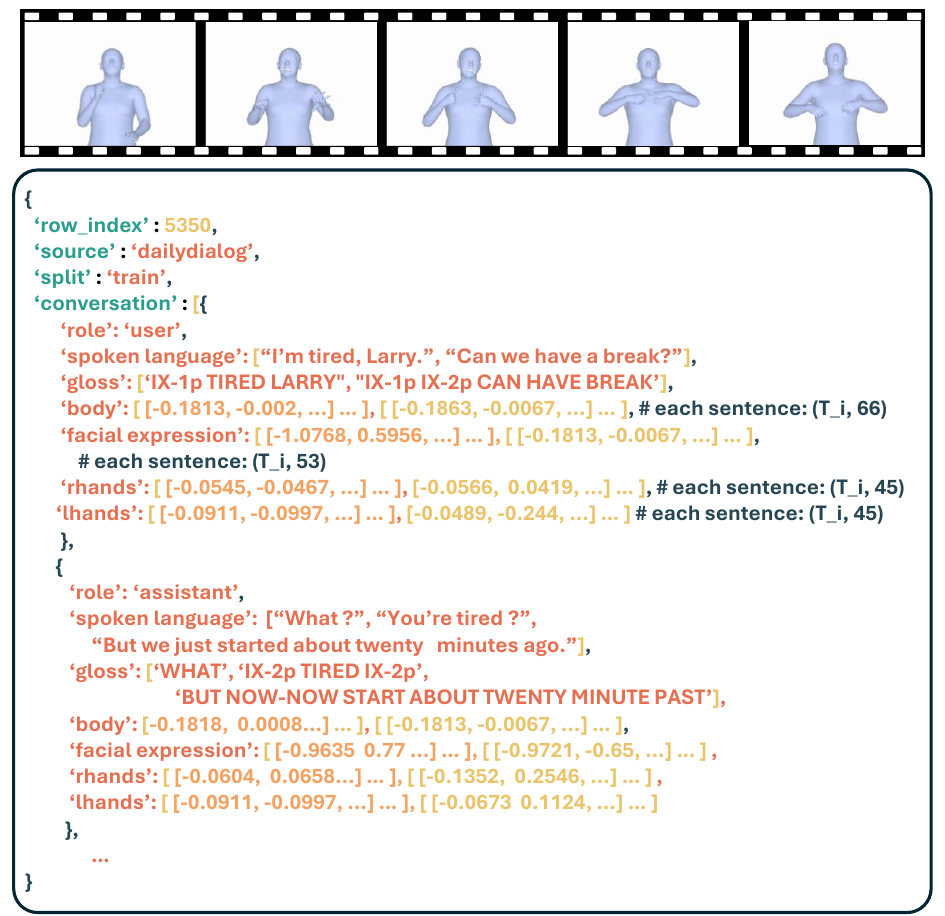}
    \caption{The \textsc{SignaVox-U} annotation format. We provide the \textsc{SignaVox-U} dataset in JSON format. $\mathrm{T}\_{\mathrm{i}}$ is a number of sentence's frames.}
    \label{fig:signavoxu_annotation}
\end{figure}

\namefig{}~\ref{fig:signavoxu_annotation} shows a sentence-level annotation example from \textsc{SignaVox-U}. A single conversation turn in \textsc{SignaVox-U} may contain multiple sentences, where each sentence consists of a spoken-language utterance, its corresponding gloss sequence, and the generated 3D sign motion. In the figure, user and assistant turns are shown separately, and the text span of each sentence is linked to the generated sign motion frames using the same color. This illustrates that \textsc{SignaVox-U} is not merely a continuous motion dataset, but provides structured alignment among conversation turns, sentences, glosses, and 3D sign motion.

\begin{figure}[t]
    \centering
    \includegraphics[width=0.95\linewidth]{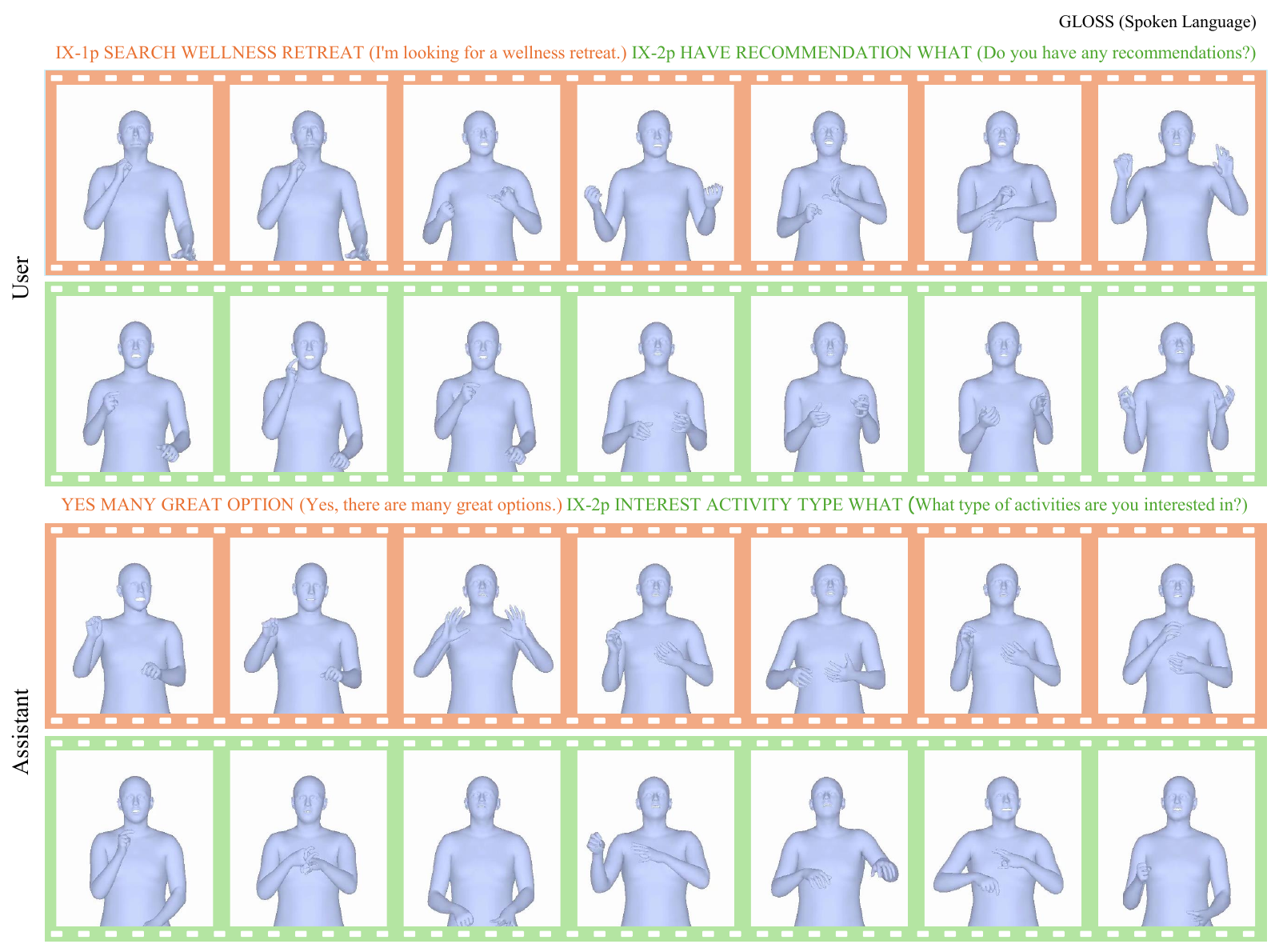}
    \caption{Visualized examples of \textsc{SignaVox-U}. The colors of the generated frames indicate their direct correspondence with the highlighted text.}
    \label{fig:signavoxu_vis}
\end{figure}
\subsection{\textsc{SignaVox} Annotation}
\begin{figure}[t]
    \centering
    \includegraphics[width=\linewidth]{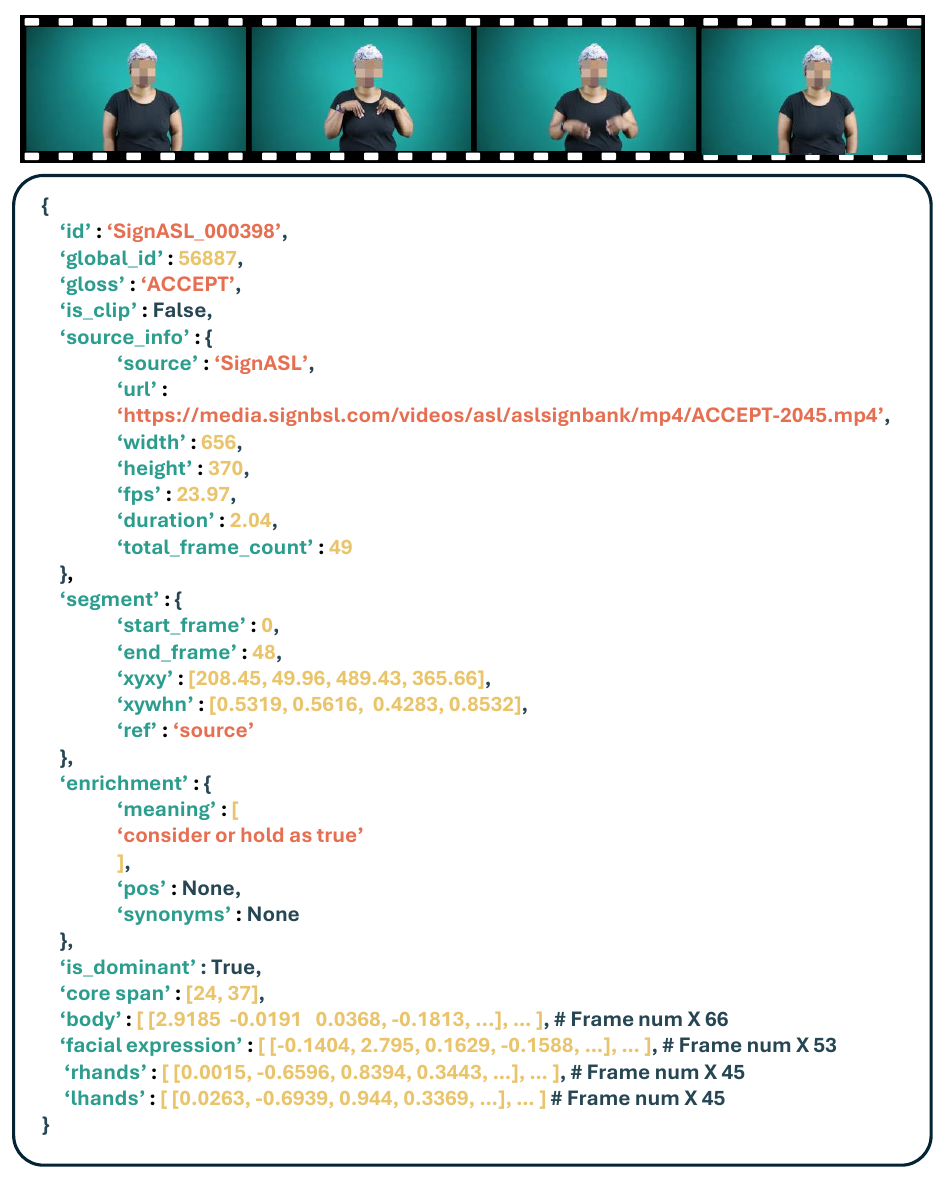}
    \caption{The \textsc{SignaVox-W} annotation format. We provide the \textsc{SignaVox-W} dataset in JSON format.}
    \label{fig:signavoxw_annotation}
\end{figure}

In this section, we describe the annotation structure of the \textsc{SignaVox}, as illustrated in \namefig{}~\ref{fig:signavoxw_annotation}.
The primary keys in \textsc{SignaVox-W} include ``gloss'', ``source info'', ``segment'', ``enrichment'', ``is dominant'', ``core span'', ``facial expression'', ``body'', ``rhands'' and ``lhands''. 
The ``gloss'' key specifies the gloss label of each video, enabling retrieval when constructing sentence-level sequences. The ``source info'' key contains metadata about the original source video.
The ``segment'' key records information about the cropped interval extracted from the source video, together with the signer bounding box. Since datasets such as MS-ASL are built from YouTube videos, the source-level information and the segment-level information may differ; we retain both for clarity and traceability. In addition, following the method proposed in Section~\ref{sec:braid_data}, we explicitly annotate the span corresponding to the core articulation frames of each sign. Finally, the ``enrichment'' key provides the linguistic meaning associated with the sign video. 
Each video-related 3D annotation is flattened into a one-dimensional array. Following the notation introduced in Section~\ref{sec:signavoxw}. 
The ``body'', ``facial expression'', ``rhands'', and ``lhands'' keys corresponding to $\boldsymbol{\theta}^{\mathrm{body}}$, $[\boldsymbol{\psi}, \theta^{\mathrm{jaw}}]$, $\mathbf{\theta}^{\mathrm{rhand}}$ and $\mathbf{\theta}^{\mathrm{lhand}}$, respectively.

\textsc{SignaVox-U} is organized at the dialogue-turn level. The ``conversation'' key contains entries structured into ``user'' and ``assistant'' turns. For each turn, we provide the spoken language and gloss annotations at the sentence level. In addition, the corresponding 3D sign-language parameters for each gloss sequence are also organized at the sentence level.

\subsection{Licensing}
Our dataset will first be released under the CC BY-NC-SA (Attribution-NonCommercial-Share-Alike) license for research purposes. Specifically, we will release the SMPL-X/FLAME/MANO annotation and provide the instruction to extract the data instead of distributing the raw videos. We also elaborate on the license of the data source we used in our dataset collection:
\begin{itemize}
    \item \textbf{MS-ASL}~\cite{msasl}. Microsoft Research dataset license terms (dataset-specific; research use).
    \item \textbf{WLASL}~\cite{wlasl}. Computational Use of Data Agreement (C-UDA-1.0).
    \item \textbf{SigningSavvy}~\cite{signingsavvy}. Accessed under the website Terms of Service; we do not redistribute original content.
    \item \textbf{Signbank}~\cite{signbank}. Creative Commons Attribution–NonCommercial–NoDerivatives 4.0 International (CC BY-NC-ND 4.0)
    \item \textbf{Spreadthesign}~\cite{spreadthesign}. Administered by the non-profit European Sign Language Centre; accessed under the website Terms of Service, and we do not redistribute original media.
    \item \textbf{SignASL}~\cite{signasl}. An online ASL dictionary with copyright notice (``ASL Sign Dictionary \textcopyright 2013–2026''); the site aggregates sign videos from multiple sources.
    \item \textbf{ASLLRP}~\cite{asllrp}. Provided by Boston University; Terms of Use allow research/educational use only, prohibit commercial use without permission, and restrict redistribution.
    \item \textbf{How2Sign}~\cite{how2sign}. Creative Commons Attribution-NonCommercial 4.0 International License.
\end{itemize}

%% file: tables/data_source_stat.tex
\begin{table}[t]
\centering
\caption{Statistics of the data sources used to construct the video–gloss dictionary after preprocessing.}
\label{tab:data_source_stat}
\small
\setlength{\tabcolsep}{3pt}
\resizebox{\columnwidth}{!}{%
\begin{tabular}{@{}ccccccc@{}}
\specialrule{0.8pt}{1pt}{1pt}
             & \# Videos & \# Gloss & Duration (h) & Annotations & Resolution & \# Frames (Min, Max, Mean) \\ \specialrule{0.5pt}{1pt}{1pt}
% Handspeak    &  $16,101$  & $11,066$   &   $9.63$     & Gloss, Meaning  &  Various    &   5 / 278 / 61.78               \\
Signbank~\cite{signbank}&   $3,402$   & $2,836$    &  $2.34$   &  Gloss, Synonym   &  Various   &       33 / 381 / 76.12                     \\
SignASL~\cite{signasl}     &  $35,833$   &  $28,269$ &   $34.40$    &  Gloss   & Various   &    16 / 457 / 101.50           \\
Spreadthesign~\cite{spreadthesign} &  $9,955$  &  $9,531$    & $10.24$      &  Gloss, POS    &  $320 \times 240$  &   44 / 379 / 100.75    \\ 
SigningSavvy~\cite{signingsavvy} & $13,461$   &  $11,493$   &  $8.75$        &  Gloss     &  $640 \times 360$    &   19 / 322 / 71.43                 \\ \hdashline
MS-ASL~\cite{msasl}       & $17,822$   &  $1,000$   &   $15.57$       &  Gloss, Synonym   & Various    &   9 / 271 / 89.11                         \\
WLASL~\cite{wlasl}        &  $11,836$   &  $1,992$  &   $7.97$     &   Gloss    & Various     &     15 / 233 / 69.11         \\ \specialrule{0.8pt}{1pt}{1pt}
\end{tabular}%
}

\end{table}

%% file: tables/data_source_signavoxu.tex
\begin{table*}[t]
\centering
\vspace{-4pt}
\setlength{\tabcolsep}{3.5pt} % 열 사이의 간격을 좁혀 가로폭 최적화
\renewcommand{\arraystretch}{1.1}
\caption{Detailed dataset statistics split by Train, Validation, and Test sets.}
\label{tab:dataset_stats}
\resizebox{\linewidth}{!}{%
\begin{tabular}{lcccccccccccc}
\specialrule{0.8pt}{1pt}{1pt}
\multirow{2}{*}{Source} & \multicolumn{3}{c}{\# Dialogues} & \multicolumn{3}{c}{\# Turns} & \multicolumn{3}{c}{\# Sentences} & \multicolumn{3}{c}{\# Avg. Frames/Sent} \\
\cmidrule(lr){2-4} \cmidrule(lr){5-7} \cmidrule(lr){8-10} \cmidrule(lr){11-13}
 & Train & Val & Test & Train & Val & Test & Train & Val & Test & Train & Val & Test \\
\specialrule{0.5pt}{1pt}{1pt}
Dailydialog~\cite{dailydialog} & 8,776 & 1,095 & 1,098 & 66,279 & 8,156 & 8,465 & 106,154 & 13,313 & 13,590 & 174.4 & 174.3 & 172.8 \\
Everyday~\cite{everydayconversations}    & 1,786 & 225   & 225   & 10,150 & 1,268 & 1,276 & 18,597  & 2,256  & 2,257  & 254.7  & 260.0   & 253.1   \\
RealTalk~\cite{realtalk}    & 166   & 21    & 21    & 5,234  & 833   & 641   & 8,471   & 1,522  & 1,038  & 258.9  & 268.1   & 284.5   \\
\specialrule{0.5pt}{1pt}{1pt}
Total       & 10,728 & 1,341 & 1,344 & 81,663 & 10,257 & 10,382 & 133,222 & 17,091 & 16,885 & 191.8 & 194.8 & 191.2 \\
\specialrule{0.8pt}{1pt}{1pt}
\end{tabular}%
}

\end{table*}

%% file: suppl/Details_frame_selection.tex
\section{Details of Frame Selection}
\label{sec:suppl_frame_selection}
\begin{figure}[t]
    \centering
    \includegraphics[width=\linewidth]{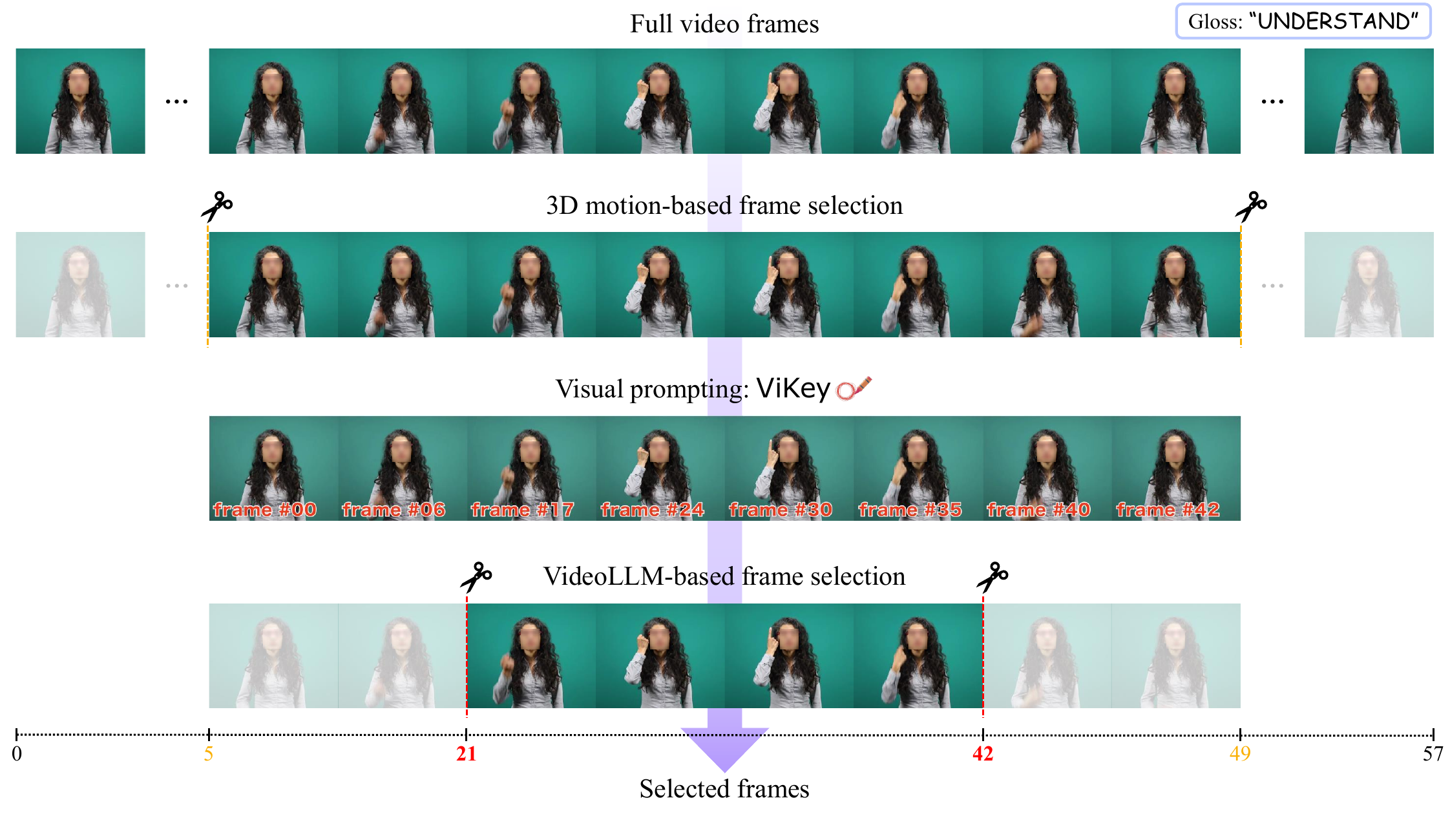}
    \caption{Overview of our frame selection pipeline. A coarse 3D motion-based stage first narrows the candidate range, and a VideoLLM with visual frame-index prompting further refines the start and end boundaries of the core articulation}
    \label{fig:frame_selection_pipeline}
\end{figure}

In this section, we describe in detail our method for selecting frames corresponding to the core articulation. The overall pipeline, which combines signer motion and a VideoLLM-based approach, is illustrated in \namefig{}~\ref{fig:frame_selection_pipeline}. Further details of the prompts used for the VideoLLM are provided in Sec.~\ref{sec:videollm_prompt}.

\begin{figure}
    \centering
    \includegraphics[width=\linewidth]{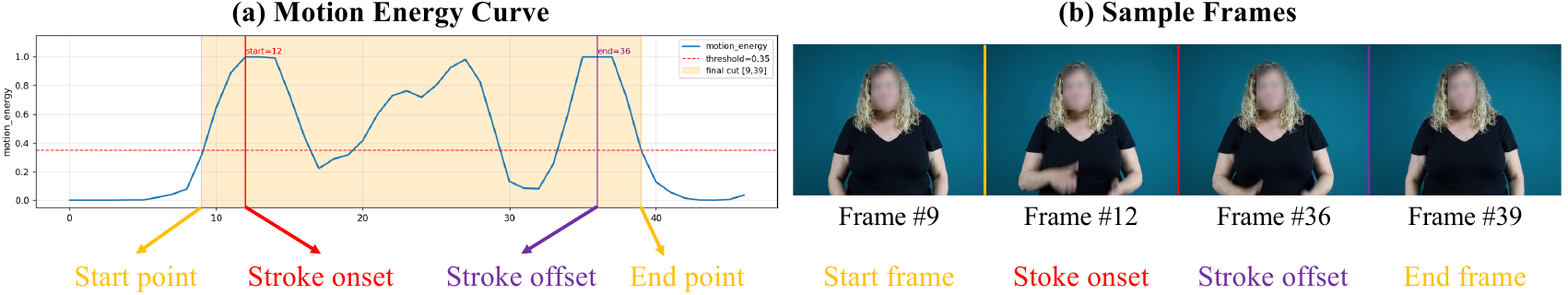}
    \caption{Example of motion-based frame selection. (a) The motion energy curve used to estimate the temporal boundaries of the core articulation. (b) Sample frames corresponding to the estimated start point ($t^{\mathrm{start}}$), stroke onset ($t^{\mathrm{on}}$), stroke offset ($t^{\mathrm{off}}$), and end point ($t^{\mathrm{end}}$). This video represents the sign for ``\textit{ACADEMICS}''}
    \label{fig:motion_frame_selection}
\end{figure}

\subsection{Energy-Based Preparation and Retraction Trimming}
\label{sec:suppl_energy_based}

To initially remove preparatory and ending motions from sign clips, we employ a bidirectional trimming method that combines arm/hand-centered motion energy with arm-posture gating.
Specifically, we first apply Savitzky--Golay smoothing~\cite{savagol} to the upper-body, arm, and hand features, and then compute motion energy from the first- and second-order temporal differences.
Given a smoothed motion feature sequence $\mathbf{X}=\{\mathbf{x}_t\}_{t=1}^{T}$, we define the raw frame-wise motion-energy score as
\begin{equation}\small
    E_t = \frac{\lambda_v}{D} \Vert \Delta \mathbf{x}_t \Vert_2^2 
    + \frac{\lambda_a}{D} \Vert \Delta^2 \mathbf{x}_t \Vert_2^2 ,
\end{equation}
where $D$ is the feature dimension, and $\lambda_v=1.0$ and $\lambda_a=0.5$ weight the velocity and acceleration terms.
For boundary frames where temporal differences are not directly defined, we pad the nearest valid difference values.

We then robustly normalize $E_t$ into $\tilde{E}_t \in [0,1]$ using quantile-based scaling and compute an arm-posture gate $h_t \in [0,1]$ from the normalized wrist-to-shoulder height.
The final trimming score, $\bar{E}_t = \tilde{E}_t \cdot h_t$, suppresses spurious motion responses from low-arm postures while preserving high-energy signing regions.

We determine the trimming boundaries using a continuity-confirmed threshold criterion.
Instead of selecting the first frame that crosses the threshold, we define the onset boundary as the end of the first stable high-energy interval.
This choice avoids retaining early arm-raising frames that may exceed the threshold before the signer reaches a sufficiently articulated signing posture.
Formally, let $\bar{E}_t$ denote the normalized and gated motion-energy score, $\theta$ the threshold, and $n=3$ the minimum number of consecutive frames required by the continuity condition.
The stroke onset $t^{\mathrm{on}}$ is defined as
\begin{equation}
t^{\mathrm{on}}
=
\min
\left\{
t+n-1
\;\middle|\;
t \in \{1,\ldots,T-n+1\},\;
\forall i \in \{0,\ldots,n-1\},\;
\bar{E}_{t+i} \ge \theta
\right\}.
\end{equation}
Thus, $t^{\mathrm{on}}$ corresponds to the confirmed onset boundary after the motion energy has remained above the threshold for $n$ consecutive frames.

Similarly, the stroke offset $t^{\mathrm{off}}$ is estimated by applying the same continuity-confirmed criterion in the reverse temporal direction.
Equivalently, in the original temporal order, it is defined as the beginning of the last stable high-energy interval:
\begin{equation}
t^{\mathrm{off}}
=
\max
\left\{
t-n+1
\;\middle|\;
t \in \{n,\ldots,T\},\;
\forall i \in \{0,\ldots,n-1\},\;
\bar{E}_{t-i} \ge \theta
\right\}.
\end{equation}

Finally, to avoid overly aggressive trimming, we add a temporal margin $m$ before and after the estimated onset and offset boundaries.
We set $m=3$ in all experiments.
The start and end points of the retained interval are defined as
\begin{equation}
t^{\mathrm{start}} = \max(1,\, t^{\mathrm{on}} - m), \qquad
t^{\mathrm{end}} = \min(T,\, t^{\mathrm{off}} + m),
\end{equation}
where $T$ denotes the total number of frames.
\namefig{}~\ref{fig:motion_frame_selection} illustrates these boundary definitions on the motion-energy curve.

\subsection{Posture Cue and Hyperparameters}
\label{sec:suppl_trimming_hyperparams}

For posture-based trimming, we compute a wrist-height cue from the SMPL-X joint sequence.
Let \(a \in \{L,R\}\) denote the left or right side, and let 
\(\mathbf{j}_t^{\mathrm{pelvis}}\), \(\mathbf{j}_t^{\mathrm{neck}}\),
\(\mathbf{j}_t^{\mathrm{shoulder},a}\), and \(\mathbf{j}_t^{\mathrm{wrist},a}\)
denote the pelvis, neck, shoulder, and wrist joints at frame \(t\), respectively.
Here, the \(y\)-axis denotes the vertical direction.
We first define the torso height as
\begin{equation}
    H_t = \max\left(
    \left|j_{t,y}^{\mathrm{neck}} - j_{t,y}^{\mathrm{pelvis}}\right|,
    10^{-6}
    \right),
\end{equation}
and normalize the wrist-to-shoulder height by this value:
\begin{equation}
    r_t^{a} =
    \frac{
    j_{t,y}^{\mathrm{wrist},a} - j_{t,y}^{\mathrm{shoulder},a}
    }{H_t}.
\end{equation}
We then use \(r_t^{\max}=\max(r_t^L,r_t^R)\) as a scale-normalized estimate of whether at least one hand is raised into an active signing posture.
For boundary detection, we form a direction-specific posture gate
\(g_t^{d}=\mathbb{1}[r_t^{\max}\ge \tau_{\mathrm{post}}^{d}]\), where \(d\in\{\mathrm{on},\mathrm{off}\}\) denotes the onset and offset scans.
This gate is applied to the normalized motion energy to suppress spurious responses from arm-down postures.

The trimming procedure uses a small set of fixed hyperparameters.
We set the activity threshold to \(\theta_{\mathrm{act}}=0.35\) and use a low-motion cutoff \(\theta_{\mathrm{low}}=0.4\) for auxiliary boundary protection.
The posture thresholds are set to \(\tau_{\mathrm{post}}^{\mathrm{on}}=0.6\) for the onset scan and \(\tau_{\mathrm{post}}^{\mathrm{off}}=0.25\) for the offset scan, reflecting the different posture patterns of preparation and retraction motions.
To avoid overly short or unstable clips, we require the retained segment to contain at least \(T_{\min}=8\) frames and remove boundary regions only when they are longer than \(B_{\min}=5\) frames.
We use a temporal margin of \(m=3\) frames on both sides of the estimated onset and offset.
For smoothing and energy computation, we use a Savitzky--Golay filter with window size \(w_{\mathrm{SG}}=7\) and polynomial order \(p_{\mathrm{SG}}=2\), and set the velocity and acceleration weights to \(\lambda_v=1.0\) and \(\lambda_a=0.5\), respectively.
Middle inactive-region removal is disabled, and the continuity criterion is fixed to \(n=3\) consecutive frames for both onset and offset detection.

For the subsequent Video-LLM refinement stage, we use the pretrained Qwen3-VL model~\cite{qwen3vl} without additional fine-tuning.
Given the coarsely trimmed clip, the model is asked to identify the contiguous frame interval corresponding to the core lexical articulation.
To make frame-level selection explicit, we overlay frame indices on the input video frames.
All queries are performed with greedy decoding by setting the temperature to \(0.0\); top-\(p\) is kept at \(0.95\) in the decoding configuration.

%% file: suppl/Details_of_braid.tex
\section{Details of Isolated to Continuous}
\subsection{Details of Data Setup.}
\input{tables/gloss_convention}
This section details the gloss normalization and filtering rules used to construct the training pairs described in Sec.~\ref{sec:isolated_continuous}.
We use ASLLRP~\cite{asllrp} as continuous signing supervision and organize its annotations into sentence-level utterances.
Following the SignStream conventions used in ASLLRP, we normalize gloss labels and summarize the filtering rules in Table~\ref{tab:gloss_convention}.
To improve consistency with \textsc{SignaVox-W} retrieval and downstream gloss-pair training, we normalize the ASLLRP gloss sequences before retrieving corresponding isolated clips.

We remove translation-like or annotative tokens, such as quoted forms and bracketed tags (\textit{e.g.}, \texttt{5"wow"}, \texttt{[false-start]}).
In contrast, we retain grammatical markers that are important for sign language structure (\textit{e.g.}, \texttt{IX-*}, \texttt{POSS-*}, \texttt{fs-*}, \texttt{\#*}, \texttt{ns-*}, and \texttt{ns-fs-*}).
We further remove classifier-related tokens (\textit{e.g.}, \texttt{DCL}, \texttt{TCL}, \texttt{PCL}, \texttt{SCL}, \texttt{BCL}, and their variants), since they exhibit high variability and limited consistency across annotations.
To improve annotation consistency, we also simplify deictic, possessive, and locus-indexed forms by removing locus suffixes where possible (\textit{e.g.}, \texttt{IX-3p:i} \(\rightarrow\) \texttt{IX-3p}, \texttt{IX-loc:j} \(\rightarrow\) \texttt{IX-loc}, and \texttt{POSS-3p:i} \(\rightarrow\) \texttt{POSS-3p}).
Finally, we remove discourse-management or meta-level markers (\textit{e.g.}, \texttt{NEXT-TOPIC}, \texttt{CURRENT-TOPIC}).

After normalization, we filter noisy English--gloss pairs using several criteria.
We discard samples whose gloss sequence becomes empty after preprocessing, as well as cases where the gloss sequence is disproportionately short relative to the English sentence (\textit{e.g.}, more than 20 English words but at most 3 gloss tokens).
We further exclude samples with very low character n-gram TF--IDF similarity between English and gloss, which often indicates weak surface correspondence.
To complement these heuristic filters, we also use GPT-4o~\cite{gpt4o} to assess semantic alignment and discard pairs judged to have substantially mismatched meanings.

\subsection{Consistency Analysis}
\begin{figure}
    \centering
    \includegraphics[width=0.5\linewidth]{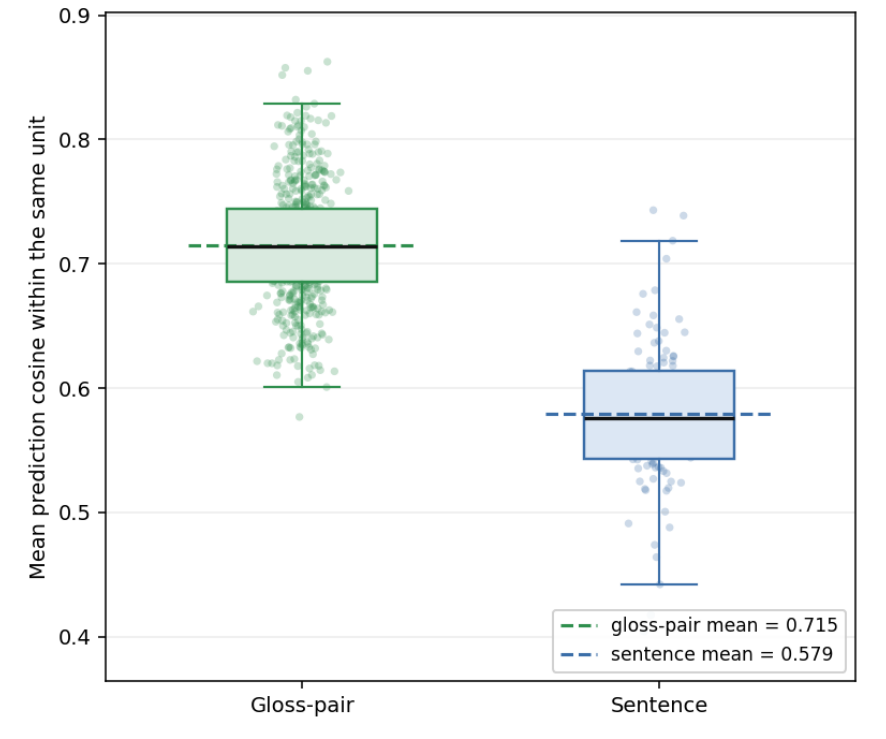}
    \caption{Consistency of \emph{BRAID} predictions across varying pseudo-input seeds. The box-plots illustrate the distribution of mean pairwise cosine similarities for motions generated from different random seeds, evaluated at both the gloss-pair and sentence levels.}
    \label{fig:consistency}
\end{figure}
During \emph{BRAID} training, we construct 14 different pseudo inputs for the same gloss pair and train the model to recover a natural co-articulatory transition between the two glosses. Although these pseudo inputs vary across seeds, the underlying transition required by the same gloss pair should remain similar. Therefore, if the refined outputs generated from different pseudo inputs exhibit consistent motion patterns, it suggests that the model captures the transition structure associated with the gloss pair rather than overfitting to seed-specific input noise.

To quantify this behavior, we measure the cosine similarity between refined motions generated from different pseudo-input seeds. For each gloss pair or sentence, we compute all pairwise cosine similarities among the 14 seed outputs and summarize them by their mean similarity. \namefig{}~\ref{fig:consistency} shows the distribution of these per-instance consistency scores for both gloss-pair-level and sentence-level predictions.
The gloss-pair predictions show high consistency across pseudo seeds, with a mean similarity of 0.715. This indicates that, despite receiving different pseudo inputs, \emph{BRAID} produces mutually similar refined transitions for the same gloss pair. Sentence-level predictions also exhibit a consistent trend, with a mean similarity of 0.579, although the score is lower due to the longer temporal span and greater motion variability at the sentence level. Overall, these results support that \emph{BRAID} learns stable co-articulatory motion patterns conditioned on the underlying gloss pair or sentence, rather than producing outputs that vary arbitrarily with the pseudo-input seed.

\subsection{Duration Predictor}
\label{sec:suppl_duration}

We provide additional details for the two duration predictors,
$\mathcal{D}_{\mathrm{gloss}}$ and $\mathcal{D}_{\mathrm{sent}}$, used in
Sec.~\ref{sec:isolated_continuous}. Both predictors estimate a global temporal scale
$\hat{s}$ and a gloss-wise duration allocation $\hat{\mathbf{w}}$, from which
integer target lengths are derived for resampling and downstream diffusion
refinement.

\paragraph{Predictor architectures.}
The gloss-pair predictor $\mathcal{D}_{\mathrm{gloss}}$ is implemented as a
4-layer MLP with hidden dimension 256. Its input consists of segment-level
motion summaries from the two isolated gloss clips, local boundary features
around the gloss transition, a boundary-jump feature, and simple length
statistics. The model predicts the global scale $\hat{s}$ and a two-way split
$\hat{\mathbf{w}}\in\mathbb{R}^{2}$ for the pair.

The sentence-level predictor $\mathcal{D}_{\mathrm{sent}}$ uses a token-based
Transformer encoder, where each gloss segment is represented as one token.
Each token encodes segment-level motion statistics and length-related
features, and valid gloss tokens are processed with padding masks. The
encoded tokens are used to predict both the sentence-level scale $\hat{s}$
and the gloss-wise allocation $\hat{\mathbf{w}}\in\mathbb{R}^{K}$ over the
valid \(K\) glosses. We set the maximum number of glosses to
\(K_{\max}=32\), which covers almost all utterances in our training data.

For both predictors, the scale prediction is bounded by clamping
\(\hat{s}\in[-3,3]\), which limits the multiplicative length rescaling
\(\exp(\hat{s})\). We also initialize the output heads to produce near-identity
rescaling and approximately uniform duration allocation at the beginning of
training.

\paragraph{Training targets.}
For both predictors, the target scale is defined as
\[
    s = \log(T_{\mathrm{tgt}} / T_{\mathrm{src}}),
\]
where \(T_{\mathrm{src}}\) and \(T_{\mathrm{tgt}}\) denote the input and target
motion lengths. For gloss pairs, the target split is directly obtained from
the target boundary position. For sentence-level prediction, we derive
per-gloss target lengths by assigning inter-gloss gaps to neighboring glosses
using midpoint boundaries, which yields a length allocation
\(\mathbf{w}^{\mathrm{GT}}\) whose entries sum to one.

Both predictors are trained with a scale loss and a split-allocation loss:
\[
\mathcal{L}_{\mathrm{dur}}
=
\rho_{\tau}(s-\hat{s})
+
\lambda_{\mathrm{split}}
\mathrm{CE}(\hat{\mathbf{w}}, \mathbf{w}^{\mathrm{GT}}),
\]
where \(\rho_{\tau}\) is the pinball quantile loss~\cite{quantile_loss}. We use a slightly larger quantile for the sentence-level predictor to reduce under-prediction, since sentence-level length errors propagate to all downstream stitched segments.

\paragraph{Inference-time duration plan.}
At inference time, the predicted scale determines the total output length,
while the predicted allocation determines the per-gloss lengths. For a
gloss pair, $\mathcal{D}_{\mathrm{gloss}}$ directly predicts the total pair
length and the boundary position between the two glosses. For a full sentence,
$\mathcal{D}_{\mathrm{sent}}$ first predicts the total sentence length and
then distributes it across the \(K\) glosses according to
\(\hat{\mathbf{w}}\). The resulting real-valued allocations are rounded with
a sum-preserving correction and constrained by a minimum length per gloss.
The final integer plan \(\{\tilde{T}_k\}_{k=1}^{K}\) is used consistently for
pairwise diffusion refinement and cosine-window stitching, ensuring that the
stitched sentence length matches the predicted sentence-level plan.

\paragraph{Hyperparameters.}
Table~\ref{tab:dur_hparam} summarizes the main hyperparameters used for both
duration predictors.

\begin{table}[h]
\centering
\small
\caption{Duration predictor hyperparameters.}
\begin{tabular}{lcc}
\specialrule{0.8pt}{1pt}{1pt}
& $\mathcal{D}_{\mathrm{gloss}}$ & $\mathcal{D}_{\mathrm{sent}}$ \\
\specialrule{0.5pt}{1pt}{1pt}
Architecture & 4-layer MLP & 3-layer Transformer \\
Hidden / model dim & 256 & 256 \\
FFN dim & -- & 512 \\
Attention heads & -- & 4 \\
Boundary window & 5 frames & 5 frames \\
Max glosses $K_{\max}$ & 2 & 32 \\
Motion dim $D$ & 206 & 206 \\
Optimizer & AdamW & AdamW \\
Learning rate & $10^{-3}$ & $10^{-3}$ \\
Weight decay & $10^{-4}$ & $10^{-4}$ \\
Batch size & 64 & 64 \\
Epochs & 50 & 60 \\
LR schedule & cosine & cosine \\
Grad. clipping & 1.0 & 1.0 \\
Quantile $\tau$ & 0.55 & 0.60 \\
$\lambda_{\mathrm{split}}$ & 1.0 & 1.0 \\
Scale clamp & $[-3,3]$ & $[-3,3]$ \\
\specialrule{0.8pt}{1pt}{1pt}
\end{tabular}

\label{tab:dur_hparam}
\end{table}

\subsection{Details of \emph{BRAID}}

\paragraph{Hyperparameters}
We trained our model using H100 GPUs, and the detailed hyperparameters are provided in Table~\ref{tab:braid_hparams}.
\begin{table}[t]
\centering
\caption{Training hyperparameters of the gloss-level diffusion model.}
\label{tab:braid_hparams}
\begin{tabular}{ll}
\toprule
Hyperparameter & Value \\
\midrule
\multicolumn{2}{l}{\textit{Architecture}} \\
\midrule
Latent / model dimension         & 512 \\
Transformer layers               & 6 \\
Attention heads                  & 8 \\
Feed-forward size                & 2048 \\
Dropout                          & 0.1 \\
Positional encoding              & RoPE \\
Hand head depth                  & 4 \\
Hand cross-attention layers / heads & 2 / 4 \\
\midrule
\multicolumn{2}{l}{\textit{Diffusion}} \\
\midrule
Diffusion steps $T$              & 1000 \\
Loss weighting                   & Min-SNR-$\gamma$ \\
Min-SNR $\gamma$                 & 5.0 \\
EMA decay                        & 0.9999 \\
Inpaint radius (min / max)       & 10 / 30 \\
Inference inpaint radius         & 10 \\
\midrule
\multicolumn{2}{l}{\textit{Optimization}} \\
\midrule
Optimizer                        & AdamW \\
Learning rate                    & $1 \times 10^{-4}$ \\
Weight decay                     & $1 \times 10^{-4}$ \\
Global batch size                & 256 \\
Epochs                           & 100 \\
Gradient clipping (max-norm)     & 1.0 \\
\midrule
\multicolumn{2}{l}{\textit{Loss weights}} \\
\midrule
Latent reconstruction $\lambda_{\text{latent}}$ (L2)   & 1.0 \\
Parameter reconstruction $\lambda_{\text{param}}$ (smooth-L1) & 1.0 \\
Velocity $\lambda_{\text{vel}}$                        & 0.5 \\
Body part weight                                       & 1.0 \\
Face part weight                                       & 3.0 \\
Hand part weight                                       & 15.0 \\
\bottomrule
\end{tabular}
\end{table}

We expand on the components introduced in the main text: the boundary-inpainting forward process, the masked part-weighted objective $\mathcal{L}_{\mathrm{braid}}$, and the inference-time composition. All notation matches the main paper.

\paragraph{Part-Wise Feature Weight Vector $\boldsymbol{\omega}_{\mathrm{part}}$}
\label{app:braid_partweight}

Sign language is dominated by hand articulation and modulated by facial expression, so we assign a non-uniform per-feature weight $\boldsymbol{\omega}_{\mathrm{part}} \in \mathbb{R}^{D}_{>0}$ that emphasizes hand and face dimensions. Let $D = D_{\mathrm{body}} + D_{\mathrm{face}} + D_{\mathrm{hand}}$ partition the feature axis into body, face (expression $+$ jaw), and hand (left $+$ right) sub-blocks. We define
\begin{equation}\small
    \omega_{\mathrm{part}}[d] =
    \begin{cases}
        \omega_{\mathrm{body}} & \text{if } d \in \mathcal{I}_{\mathrm{body}}, \\
        \omega_{\mathrm{face}} & \text{if } d \in \mathcal{I}_{\mathrm{face}}, \\
        \omega_{\mathrm{hand}} & \text{if } d \in \mathcal{I}_{\mathrm{hand}}, \\
    \end{cases}
\end{equation}
with $(\omega_{\mathrm{body}}, \omega_{\mathrm{face}}, \omega_{\mathrm{hand}}) = (1.0, 3.0, 15.0)$ in our experiments

\paragraph{Masked Reconstruction Loss $\mathcal{L}_{\mathrm{recon}}$}
\label{app:braid_recon}

$\mathcal{L}_{\mathrm{recon}}$ is a Smooth-L1 loss applied only on the supervised inpainting region $\{i: M_i = 1\}$, with per-feature weighting by $\boldsymbol{\omega}_{\mathrm{part}}$:
\begin{equation}\small
    \mathcal{L}_{\mathrm{recon}}(\hat{\mathbf{X}}_0, \mathbf{X}_0; \mathbf{M}, \boldsymbol{\omega}_{\mathrm{part}})
    =
    \frac{
        \sum_{i=1}^{\tilde{T}} \sum_{d=1}^{D}
        M_i \,\omega_{\mathrm{part}}[d] \,
        \rho_\beta\!\bigl(\hat{X}_{0,i,d} - X_{0,i,d}\bigr)
    }{
        \sum_{i=1}^{\tilde{T}} \sum_{d=1}^{D}
        M_i \,\omega_{\mathrm{part}}[d]
    },
\end{equation}
where $\rho_\beta(\cdot)$ is the elementwise Smooth-L1 (Huber) function with transition $\beta$:
\begin{equation}\small
    \rho_\beta(u) =
    \begin{cases}
        \tfrac{1}{2\beta}\, u^{2}, & |u| < \beta, \\[2pt]
        |u| - \tfrac{\beta}{2}, & |u| \ge \beta.
    \end{cases}
\end{equation}
We use $\beta = 1.0$.

\paragraph{Masked Velocity Regularization $\mathcal{L}_{\mathrm{vel}}$}
\label{app:braid_vel}

The velocity regularization is computed on first-order temporal differences. Let $\Delta \mathbf{X}_{0,i} = \mathbf{X}_{0,i+1} - \mathbf{X}_{0,i}$ (and similarly $\Delta \hat{\mathbf{X}}_{0,i}$) for $i = 1, \ldots, \tilde{T}-1$. The velocity-domain mask retains only differences between two supervised frames:
\begin{equation}\small
    M^{\Delta}_{i} \;=\; M_{i} \cdot M_{i+1},
    \qquad i = 1, \ldots, \tilde{T}-1.
\end{equation}
Then
\begin{equation}\small
    \mathcal{L}_{\mathrm{vel}}(\Delta \hat{\mathbf{X}}_0, \Delta \mathbf{X}_0; \mathbf{M}^{\Delta}, \boldsymbol{\omega}_{\mathrm{part}})
    =
    \frac{
        \sum_{i=1}^{\tilde{T}-1} \sum_{d=1}^{D}
        M^{\Delta}_i \,\omega_{\mathrm{part}}[d] \,
        \rho_\beta\!\bigl(\Delta \hat{X}_{0,i,d} - \Delta X_{0,i,d}\bigr)
    }{
        \sum_{i=1}^{\tilde{T}-1} \sum_{d=1}^{D}
        M^{\Delta}_i \,\omega_{\mathrm{part}}[d]
    }.
\end{equation}
This penalizes per-frame jerk specifically on the boundary region, where co-articulatory transitions matter most.

\paragraph{Min-SNR-$\gamma$ Timestep Weighting $w(t)$}
\label{app:braid_minsnr}

We adopt the Min-SNR-$\gamma$ scheme of Hang \emph{et al.}~\cite{minsnr} to prevent low-SNR diffusion steps from dominating the gradient. With signal-to-noise ratio $\mathrm{SNR}(t) = \bar{\alpha}_t / (1 - \bar{\alpha}_t)$,
\begin{equation}\small
    w(t) \;=\; \frac{\min\!\bigl(\mathrm{SNR}(t),\,\gamma\bigr)}{\mathrm{SNR}(t)}.
\end{equation}
We set $\gamma = 5$, sample $t \sim \mathcal{U}\{1,\ldots,T\}$ with $T = 1000$, and apply $w(t)$ as a per-sample multiplier on $\mathcal{L}_{\mathrm{recon}}$ and $\mathcal{L}_{\mathrm{vel}}$, recovering the full objective in the main text.

\subsection{Inference-Time Composition}
\label{app:braid_inference}

At inference we use DDIM sampling~\cite{ddim} with $\eta = 0$ on a respaced subset of $S \ll T$ timesteps. The reverse process is initialized only inside the boundary region:
\begin{equation}\small
    \mathbf{X}_{t_S} \;=\; \mathbf{M} \odot \boldsymbol{\eta} \;+\; (1 - \mathbf{M}) \odot \tilde{\mathbf{X}}^{(k,k+1)},
    \qquad \boldsymbol{\eta} \sim \mathcal{N}(\mathbf{0}, \mathbf{I}).
\end{equation}
At each respaced step $t_s \to t_{s-1}$, given $\hat{\mathbf{X}}_0 = G_\theta(\mathbf{X}_{t_s}, t_s, \mathbf{c}, \mathbf{M})$, the prediction is first composed with the temporal context to keep $M_i = 0$ frames pinned to the duration-adjusted pseudo sequence,
\begin{equation}\small
    \hat{\mathbf{X}}_0 \;\leftarrow\; \mathbf{M} \odot \hat{\mathbf{X}}_0 \;+\; (1 - \mathbf{M}) \odot \tilde{\mathbf{X}}^{(k,k+1)},
\end{equation}
and the deterministic DDIM update is then applied with $\boldsymbol{\epsilon}_\theta = (\mathbf{X}_{t_s} - \sqrt{\bar{\alpha}_{t_s}}\,\hat{\mathbf{X}}_0)/\sqrt{1-\bar{\alpha}_{t_s}}$,
\begin{equation}\small
    \mathbf{X}_{t_{s-1}} \;=\; \sqrt{\bar{\alpha}_{t_{s-1}}}\,\hat{\mathbf{X}}_0 \;+\; \sqrt{1 - \bar{\alpha}_{t_{s-1}}}\,\boldsymbol{\epsilon}_\theta.
\end{equation}
The final refined gloss-pair motion is the converged sample $\hat{\mathbf{X}}_0^{\mathrm{comp}}$, which is used as the building block for sentence-level stitching across all consecutive gloss pairs.

%% file: tables/gloss_convention.tex
\newcommand{\TblFont}{\fontsize{7.2}{8.6}\selectfont}
\newcommand{\TblTabColSep}{4pt}
\newcommand{\TblRowStretch}{1.3}

% Column widths
\newlength{\Wcat}\setlength{\Wcat}{0.20\textwidth}
\newlength{\Wgls}\setlength{\Wgls}{0.16\textwidth}
\newlength{\Wex}\setlength{\Wex}{0.16\textwidth}
\newlength{\Wexp}\setlength{\Wexp}{0.39\textwidth}

% Rules
\newcommand{\TopBottomRule}{\specialrule{0.9pt}{0pt}{0pt}}
\newcommand{\SectionRule}{\specialrule{0.45pt}{0pt}{0pt}}
\newcommand{\RowLineExExp}{\cline{3-4}}
\newcommand{\RowLineGlExExp}{\cline{2-4}}

\newcolumntype{M}[1]{>{\raggedright\arraybackslash\hspace{0pt}}m{#1}}

\begin{table}[t]
\centering
\label{tab:gloss_convention}

\TblFont
\setlength{\tabcolsep}{\TblTabColSep}
\caption{Gloss conventions adopted for dataset construction.}
\renewcommand{\arraystretch}{\TblRowStretch}

\begin{tabular}{M{\Wcat} M{\Wgls} M{\Wex} M{\Wexp}}
\TopBottomRule
\textbf{Category} & \textbf{Gloss} & \textbf{Example} & \textbf{Explanation} \\
\SectionRule

English-based glosses
& -
& THANK-YOU
& Used to separate words if the English translation of a single sign requires more than one. \\
\SectionRule

\multirow[c]{2}{\Wcat}[-1.3ex]{Fingerspelling}
& fs-
& fs-J-O-H-N
& Fingerspelled word. \\
\RowLineGlExExp
& \#
& \#-E-A-R-L-Y
& Fingerspelled loan sign. Represented with fingerspelling. \\
\SectionRule

Name Signs
& ns-
& ns-P-A-R-I-S
& Used for proper nouns (\textit{e.g.} people, places, etc.) \\
\SectionRule

\multirow[c]{3}{\Wcat}[-1.5ex]{Pronoun}
& \multirow[c]{3}{\Wgls}[-1.5ex]{IX-[person]}
& IX-1p
& First-person pronoun (I / WE). \\
\RowLineExExp
&  & IX-2p
& Second-person pronoun referring to the addressee (YOU). \\
\RowLineExExp
&  & IX-3p
& Third-person pronoun (HE / SHE / THEY). \\
\SectionRule

\multirow{3}{\Wcat}{Possessive}
& \multirow{3}{\Wgls}{POSS-[person]}
& POSS-1p
& First-person possessive marker (MY / OUR). \\
\RowLineExExp
&  & POSS-2p
& Second-person possessive marker (YOUR). \\
\RowLineExExp
&  & POSS-3p
& Third-person possessive marker (HIS / HER / THEIR). \\
\SectionRule

\multirow[c]{3}{\Wcat}[-2.6ex]{Emphatic reflexive}
& \multirow[c]{3}{\Wgls}[-2.6ex]{SELF-[person]}
& SELF-1p
& First-person emphatic reflexive marker (MYSELF / OURSELVES). \\
\RowLineExExp
&  & SELF-2p
& Second-person emphatic reflexive marker referring to the addressee (YOURSELF / YOURSELVES). \\
\RowLineExExp
&  & SELF-3p
& Third-person emphatic reflexive marker (HIMSELF / HERSELF / THEMSELVES). \\
\TopBottomRule
\end{tabular}
\captionsetup{skip=4pt}

\label{tab:gloss_convention}
\end{table}

%% file: suppl/Details_t2g.tex
\section{Details of Spoken Language to Gloss}
\subsection{Hyperparameters}
Anonymization is applied only to the retrieval index and retrieval queries; the original English utterances, retrieved demonstrations, and target gloss sequences are kept unchanged in the LLM prompt.
Named entities are detected using spaCy~\cite{spacy} and replaced with coarse placeholders: \texttt{PERSON} $\rightarrow$ ``someone,'' \texttt{ORG} $\rightarrow$ ``some organization,'' \texttt{GPE}/\texttt{LOC} $\rightarrow$ ``some place,'' \texttt{FAC} $\rightarrow$ ``some facility,'' and \texttt{NORP} $\rightarrow$ ``some group.''
For overlapping entity spans, we retain the longest span, and apply the same rule when multiple spans share the same start position.

For first-stage retrieval, BM25 uses $k_1=1.5$ and $b=0.75$.
SPLADE representations are computed with a maximum input length of 256 tokens, retaining up to 128 document terms and 64 query terms.
BM25 and SPLADE scores are normalized independently with per-query min--max normalization and combined as
\[
s_{\mathrm{first}} = \alpha s_{\mathrm{BM25}} + (1-\alpha)s_{\mathrm{SPLADE}},
\]
where $\alpha=0.35$.
We pass the top 30 first-stage candidates to the second-stage reranker, and compute the final score as
\[
s_{\mathrm{final}} = 0.85 \cdot s_{\mathrm{reranker}} + 0.15 \cdot s_{\mathrm{first}}.
\]
Finally, we deduplicate candidates using normalized English sentences and select the top 6 examples for the translation-memory prompt.

\subsection{Experimental Details}
To evaluate the performance of the gloss translator, we construct a retrieval set of 100 examples and conduct experiments on the remaining 1,261 spoken language–gloss pairs.
For a more appropriate quantitative assessment of semantic fidelity and grammatical correctness, finger-spelled expressions are normalized into a single lexical token prior to evaluation (\textit{e.g.} ns-fs-PARIS $\rightarrow$ PARIS). Since the official implementation of the previous work~\cite{nms_generation} is not publicly available, we reproduced its methodology based on the descriptions provided in the work. We will publicly release both the evaluation set and the retrieval set.

%% file: suppl/Details_sign2sign.tex
\section{Details of Sign Language Conversational Model}
\label{sec:sign2sign_model}

\subsection{Retrieval-Based Semantic Evaluation}
\label{sec:retrieval_based_translation}
We propose a hybrid retrieval-based evaluation protocol to assess the semantic recoverability of generated sign motions. Instead of directly decoding the generated motion with a learned sign-to-text model, we compare the input motion against two complementary memory banks: the sentence-level \textsc{SignaVox-U} memory and the gloss-level \textsc{SignaVox-W} memory.

Given an input sentence-level motion sequence \(x\), we first extract a motion feature embedding using an encoder \(\phi(\cdot)\). We then retrieve sentence-level candidates from the \textsc{SignaVox-U} memory by computing cosine similarity between the input motion and each sentence motion \(u_i\):
\begin{equation}\small
s_U(x,u_i) = \phi(x)^\top \phi(u_i),
\end{equation}
where \(u_i\) denotes a sentence-level motion sequence in \textsc{SignaVox-U}. Since each \textsc{SignaVox-U} sentence is paired with spoken-language text and gloss annotations, the retrieved candidates provide pseudo translation candidates for the input motion.

In parallel, we perform gloss-level retrieval using \textsc{SignaVox-W}. For each retrieved sentence candidate, we divide the input motion \(x\) into \(K\) temporal segments according to the length of the candidate gloss sequence. Each segment is then compared with the \textsc{SignaVox-W} gloss prototype memory, producing gloss-level evidence \(\hat{g}_W(x)\) that indicates which gloss sequence the input motion locally resembles.

For each \textsc{SignaVox-U} candidate \(u_i\), we compute a final hybrid score by combining sentence-level motion similarity, gloss-level consistency, and gloss retrieval confidence:
\begin{equation}\small
S(x,u_i)
=
\lambda_U \tilde{s}_U(x,u_i)
+
\lambda_G \mathrm{F1}(\hat{g}_W(x), g_i)
+
\lambda_C c_W(x),
\end{equation}
where \(\tilde{s}_U(x,u_i)\) is the min--max normalized sentence-level retrieval score among the top-\(k\) candidates, \(g_i\) is the reference gloss sequence of candidate \(u_i\), \(\mathrm{F1}(\hat{g}_W(x), g_i)\) measures token-level overlap between the gloss evidence retrieved from \textsc{SignaVox-W} and the candidate gloss sequence, and \(c_W(x)\) denotes the gloss retrieval confidence. We use \(\lambda_U=0.55\), \(\lambda_G=0.40\), and \(\lambda_C=0.05\) by default.

The candidate with the highest hybrid score is selected as the hybrid pseudo translation for the input motion.
\begin{table}[t]
\centering
\caption{Upper-bound calibration of the hybrid retrieval-based semantic evaluator on ground-truth assistant test motions.}
\label{tab:hybrid_retrieval_upper_bound}
\resizebox{\linewidth}{!}{%
\begin{tabular}{lccccccc}
\specialrule{1pt}{1pt}{1pt}
Setting & MRR & R@1 & R@5 & R@10 & Gloss F1 & Gloss BLEU-4 & Gloss chrF \\
\specialrule{0.5pt}{1pt}{1pt}
GT motion query
& 0.495 & 0.415 & 0.584 & 0.644 & 0.479 & 0.305 & 0.498 \\
\specialrule{0.8pt}{1pt}{1pt}
\end{tabular}
}
\captionsetup{skip=4pt}
\end{table}

\subsection{Implementation Details}
\label{app:implementation}

\noindent \textbf{Model Configuration.}
Table~\ref{tab:signavox_arch_hparams} summarizes the architecture hyperparameters used for \textsc{SignaVox}.
The model uses block-wise autoregressive generation with a Qwen-style causal Transformer decoder and anatomy-factorized flow heads.

\begin{table}[h]
\centering
\caption{Architecture hyperparameters of \textsc{SignaVox}.}
\label{tab:signavox_arch_hparams}
\small
\begin{tabular}{l c}
\specialrule{1pt}{1pt}{1pt}
\textbf{Component} & \textbf{Value} \\
\specialrule{0.5pt}{1pt}{1pt}
Motion block size \(K\) & 8 \\
Body / face / hand embedding chunks & 256 / 256 / 256 \\
Transformer hidden size & 768 \\
Transformer layers & 14 \\
Attention heads & 12 \\
Key-value heads & 4 \\
FFN intermediate size & 2048 \\
Maximum sequence length & 2048 \\
Body flow head hidden size & 256 \\
Face flow head hidden size & 256 \\
Hand flow head hidden size & 320 \\
Body / face flow head depth & 3 \\
Hand flow head depth & 4 \\
Flow head attention heads & 4 \\
Sampling steps per block & 8 \\
Maximum speakers & 8 \\
Maximum turns & 32 \\
\specialrule{0.8pt}{1pt}{1pt}
\end{tabular}
\end{table}

\noindent \textbf{Training Setup.}
We train \textsc{SignaVox} in the single-turn response generation setting.
Each training example consists of one source sign turn as context and the corresponding target response.
The model is trained to generate assistant responses, and context dropout is disabled in this setting.
Table~\ref{tab:signavox_training_hparams} summarizes the optimization and data settings.

\begin{table}[h]
\centering
\caption{Training hyperparameters for \textsc{SignaVox}.}
\label{tab:signavox_training_hparams}
\small
\begin{tabular}{l c}
\specialrule{1pt}{1pt}{1pt}
\textbf{Hyperparameter} & \textbf{Value} \\
\specialrule{0.5pt}{1pt}{1pt}
Target role & assistant \\
Maximum context turns & 1 \\
Maximum context slots & 1536 \\
Maximum target slots & 384 \\
Context dropout probability & 0.0 \\
Batch size per GPU & 16 \\
Number of GPUs & 2 \\
Gradient accumulation & 1 \\
Effective batch size & 32 \\
Optimizer & AdamW \\
Learning rate & \(3\times 10^{-4}\) \\
Adam betas & \((0.9, 0.95)\) \\
Weight decay & 0.01 \\
LR warmup steps & 2000 \\
LR schedule & cosine decay \\
Gradient clipping & 1.0 \\
Precision & bfloat16 mixed precision \\
Maximum training steps & 200k \\
Maximum epochs & 120 \\
Selected checkpoint & epoch 119 \\
\specialrule{0.8pt}{1pt}{1pt}
\end{tabular}
\end{table}

\noindent \textbf{Loss Weights and Curriculum.}
Table~\ref{tab:signavox_loss_hparams} reports the weights used for the auxiliary losses and curriculum schedules.
The model is trained with conditional flow matching as the main motion generation objective, together with boundary prediction, CTC-based gloss planning, post-motion CTC, and landmark gloss supervision.

\begin{table}[h]
\centering
\caption{Loss weights and curriculum schedules for \textsc{SignaVox}.}
\label{tab:signavox_loss_hparams}
\small
\begin{tabular}{l c}
\specialrule{1pt}{1pt}{1pt}
\textbf{Term} & \textbf{Value} \\
\specialrule{0.5pt}{1pt}{1pt}
Boundary loss weight \(\lambda_{\mathrm{bdry}}\) & 0.3 \\
Sentence boundary positive weight & 20.0 \\
Turn boundary positive weight & 12.0 \\
Boundary-rate calibration weight & 0.05 \\
Plan CTC weight \(\lambda_{\mathrm{plan}}\) & 0.35 \\
Post-motion CTC weight \(\lambda_{\mathrm{post}}\) & 0.05 \\
Landmark gloss weight \(\lambda_{\mathrm{lm}}\) & 0.02 \\
Plan-to-flow final weight & 0.4 \\
Plan-to-flow warmup & 2000--10000 steps \\
Plan CTC warmup & 500--5000 steps \\
Post CTC / landmark warmup & 3000--8000 steps \\
Hand flow loss weight & 1.0 \\
\specialrule{0.8pt}{1pt}{1pt}
\end{tabular}
\end{table}

The boundary-rate calibration term aligns the average predicted boundary frequency with the empirical target frequency within each batch:
\begin{equation}\small
\mathcal{L}_{\mathrm{rate}}
=
(\bar{p}_{\mathrm{sent}}-\bar{y}_{\mathrm{sent}})^2
+
(\bar{p}_{\mathrm{turn}}-\bar{y}_{\mathrm{turn}})^2,
\end{equation}
where \(\bar{p}_{\mathrm{sent}}\) and \(\bar{p}_{\mathrm{turn}}\) denote the mean predicted sentence-end and turn-end probabilities over valid blocks, and \(\bar{y}_{\mathrm{sent}}\) and \(\bar{y}_{\mathrm{turn}}\) denote the corresponding target rates.

For plan-to-flow conditioning, the pre-motion gloss planner outputs log probabilities over the gloss vocabulary.
We convert them into a soft gloss embedding by taking the expectation over a learned gloss embedding table.
The resulting embedding is passed through a projection MLP and added to the flow-head memory.
During training, the contribution of this semantic conditioning path is multiplied by a curriculum weight that linearly increases to its final value.

\subsection{Loss Definitions}
\label{app:loss_definitions}

We provide the explicit definitions of the loss terms used to train \textsc{SignaVox}.
For a target response \(Y_t\), let \(B_\ell\) denote the \(\ell\)-th \(K\)-frame ground-truth motion block, and let \(C_\ell\) be the decoder conditioning memory for that block.
Each block is decomposed into anatomical components:
\begin{equation}\small
B_\ell =
\left[
B_\ell^{\mathrm{body}},
B_\ell^{\mathrm{face}},
B_\ell^{\mathrm{hand}}
\right].
\end{equation}

\paragraph{Flow matching loss.}
For each anatomical component \(a\in\{\mathrm{body},\mathrm{face},\mathrm{hand}\}\), we sample Gaussian noise \(x_0^a\sim\mathcal{N}(0,I)\), take the target block component \(x_1^a=B_\ell^a\), and sample \(\tau\sim\mathcal{U}(0,1)\).
The interpolation point is
\begin{equation}\small
x_\tau^a = (1-\tau)x_0^a+\tau x_1^a.
\end{equation}
The component-wise flow matching loss is
\begin{equation}\small
\mathcal{L}_{\mathrm{FM}}^{a}
=
\mathbb{E}_{\ell,x_0^a,\tau}
\left[
\left\|
v_{\theta}^{a}(x_\tau^a,\tau;C_\ell)
-
(x_1^a-x_0^a)
\right\|_2^2
\right].
\end{equation}
The total motion generation loss is
\begin{equation}\small
\mathcal{L}_{\mathrm{FM}}
=
\mathcal{L}_{\mathrm{FM}}^{\mathrm{body}}
+
\mathcal{L}_{\mathrm{FM}}^{\mathrm{face}}
+
\lambda_{\mathrm{hand}}
\mathcal{L}_{\mathrm{FM}}^{\mathrm{hand}}.
\end{equation}

\paragraph{Boundary loss.}
For each valid block \(\ell\), the boundary head predicts sentence-end and turn-end probabilities:
\begin{equation}\small
p_\ell^{\mathrm{sent}}=\sigma(z_\ell^{\mathrm{sent}}),
\qquad
p_\ell^{\mathrm{turn}}=\sigma(z_\ell^{\mathrm{turn}}),
\end{equation}
with binary targets \(y_\ell^{\mathrm{sent}},y_\ell^{\mathrm{turn}}\in\{0,1\}\).
No-boundary corresponds to \(y_\ell^{\mathrm{sent}}=0\) and \(y_\ell^{\mathrm{turn}}=0\).

Let \(\mathcal{V}\) be the set of valid, non-padding blocks.
We use weighted binary cross entropy for sentence-end and turn-end prediction:
\begin{equation}\small
\mathcal{L}_{\mathrm{BCE}}^{\mathrm{sent}}
=
-\frac{1}{|\mathcal{V}|}
\sum_{\ell\in\mathcal{V}}
\left[
\alpha_{\mathrm{sent}}y_\ell^{\mathrm{sent}}\log p_\ell^{\mathrm{sent}}
+
(1-y_\ell^{\mathrm{sent}})\log(1-p_\ell^{\mathrm{sent}})
\right],
\end{equation}
\begin{equation}\small
\mathcal{L}_{\mathrm{BCE}}^{\mathrm{turn}}
=
-\frac{1}{|\mathcal{V}|}
\sum_{\ell\in\mathcal{V}}
\left[
\alpha_{\mathrm{turn}}y_\ell^{\mathrm{turn}}\log p_\ell^{\mathrm{turn}}
+
(1-y_\ell^{\mathrm{turn}})\log(1-p_\ell^{\mathrm{turn}})
\right].
\end{equation}
Here, \(\alpha_{\mathrm{sent}}\) and \(\alpha_{\mathrm{turn}}\) are positive-class weights.

To calibrate the predicted boundary frequency, we add a boundary-rate loss:
\begin{equation}\small
\mathcal{L}_{\mathrm{rate}}
=
(\bar{p}^{\mathrm{sent}}-\bar{y}^{\mathrm{sent}})^2
+
(\bar{p}^{\mathrm{turn}}-\bar{y}^{\mathrm{turn}})^2,
\end{equation}
where
\begin{equation}\small
\bar{p}^{\mathrm{sent}}
=
\frac{1}{|\mathcal{V}|}
\sum_{\ell\in\mathcal{V}}
p_\ell^{\mathrm{sent}},
\qquad
\bar{y}^{\mathrm{sent}}
=
\frac{1}{|\mathcal{V}|}
\sum_{\ell\in\mathcal{V}}
y_\ell^{\mathrm{sent}},
\end{equation}
and \(\bar{p}^{\mathrm{turn}}\), \(\bar{y}^{\mathrm{turn}}\) are defined analogously.
The full boundary loss is
\begin{equation}\small
\mathcal{L}_{\mathrm{bdry}}
=
\mathcal{L}_{\mathrm{BCE}}^{\mathrm{sent}}
+
\mathcal{L}_{\mathrm{BCE}}^{\mathrm{turn}}
+
\lambda_{\mathrm{rate}}
\mathcal{L}_{\mathrm{rate}}.
\end{equation}

\paragraph{Pre-motion CTC planning loss.}
Let \(G=(g_1,\ldots,g_J)\) be the target gloss sequence.
The pre-motion planning head predicts a block-level gloss posterior sequence \(P_{\mathrm{plan}}\) before observing the corresponding target motion blocks.
Using CTC, the planning loss marginalizes over all monotonic alignments \(\pi\) that collapse to \(G\):
\begin{equation}\small
\mathcal{L}_{\mathrm{plan}}
=
-\log
\sum_{\pi\in\mathcal{A}(G)}
\prod_{\ell=1}^{L}
P_{\mathrm{plan}}(\pi_\ell \mid C_\ell),
\end{equation}
where \(\mathcal{A}(G)\) is the set of CTC alignments whose collapse equals \(G\).

\paragraph{Post-motion CTC loss.}
We also apply a weaker CTC loss to the hidden states after teacher-forced target motion is observed.
Let \(H_\ell^{\mathrm{post}}\) denote the post-motion hidden representation for block \(\ell\), and let \(P_{\mathrm{post}}\) be the corresponding gloss posterior sequence.
The post-motion CTC loss is
\begin{equation}\small
\mathcal{L}_{\mathrm{post}}
=
-\log
\sum_{\pi\in\mathcal{A}(G)}
\prod_{\ell=1}^{L}
P_{\mathrm{post}}(\pi_\ell \mid H_\ell^{\mathrm{post}}).
\end{equation}

\paragraph{Landmark gloss loss.}
For a subset of gloss labels, we use approximate landmark block indices indicating where a gloss is centered in the response.
Let \(\mathcal{Q}\) be the set of valid landmark annotations, where each item \((j,q_j)\) contains the gloss index \(j\) and its landmark block index \(q_j\).
The landmark loss is
\begin{equation}\small
\mathcal{L}_{\mathrm{lm}}
=
-\frac{1}{|\mathcal{Q}|}
\sum_{(j,q_j)\in\mathcal{Q}}
\log
P_{\mathrm{post}}(g_j \mid H_{q_j}^{\mathrm{post}}).
\end{equation}
If no valid landmark annotation is available, this term is masked out.

%% file: suppl/Additional_Experiments.tex
\section{Additional Experiments Results}
\label{sec:suppl_experiments}
\subsection{Frame Selection}
\input{tables/evaluation_frame_selection_gloss_level}
We evaluate how effectively the proposed frame selection pipeline identifies core articulation frames.
To this end, we compare the selected isolated sign segments with gloss-level ground-truth segments from ASLLRP using DTW-MPJPE and DTW-MPVPE.
We also report FGD to measure the distributional similarity of motion features, and Length Ratio to quantify the temporal mismatch between the selected segment and the target gloss-level segment.
The results are shown in Table~\ref{tab:frame_selection2}.

The raw isolated clips show large errors and a severe length mismatch, indicating that they contain substantial redundant frames such as preparation, retraction, and non-signing intervals.
Applying motion-based selection reduces both motion error and temporal mismatch, while VideoLLM-only selection further improves all metrics.
Our full pipeline achieves the best results across DTW-MPJPE, DTW-MPVPE, and FGD, and substantially reduces the Length Ratio compared with the raw input.
These results suggest that coarse motion-based localization and VideoLLM-based refinement are complementary, enabling more accurate selection of core articulation frames.

\subsection{Isolated-to-Continuous Signing}

\noindent \textbf{Ablation Studies of Training Components.}
Table~\ref{tab:suppl_braid_ablation} ablates the training components of \emph{BRAID}.
Without the proposed training stabilizers, the model shows a large vertex-level error despite a reasonable joint-level error, suggesting unstable reconstruction of fine-grained motion.
Min-SNR weighting and EMA substantially reduce DTW-MPVPE, indicating that noise-level reweighting and parameter averaging stabilize diffusion training.
Inpainting supervision alone improves DTW-MPJPE, but is less effective for vertex-level accuracy, suggesting that boundary-focused supervision benefits trajectory alignment but requires stable diffusion training to improve detailed motion.
The full configuration achieves the lowest DTW-MPJPE and DTW-MPVPE.

Table~\ref{tab:hand_ablation_results} reports the effect of varying the depth of the hand-specific MLP in the part-wise motion head. Performance improves as the depth increases up to \( \times 4 \), which achieves the lowest DTW-MPJPE and DTW-MPVPE in both hand and overall metrics. This suggests that additional hand-specific capacity is beneficial for modeling fine-grained manual articulation. However, increasing the depth further to \( \times 5 \) degrades performance, indicating that excessive depth does not provide additional benefit and may reduce prediction stability. Based on these results, we adopt the \( \times 4 \) configuration.

% \noindent \textbf{Effect of Inpainting Range}. \textcolor{red}{해당 부분 제목 일단 바꿔야됨.} 여기서 inference 단에서 하는거랑 Training 단에서 하는거 range 조절해서 하는거 그거 실험 돌려서 실험.

\noindent \textbf{Qualitative Results.}
\begin{figure}[t]
    \centering
    \includegraphics[width=\linewidth]{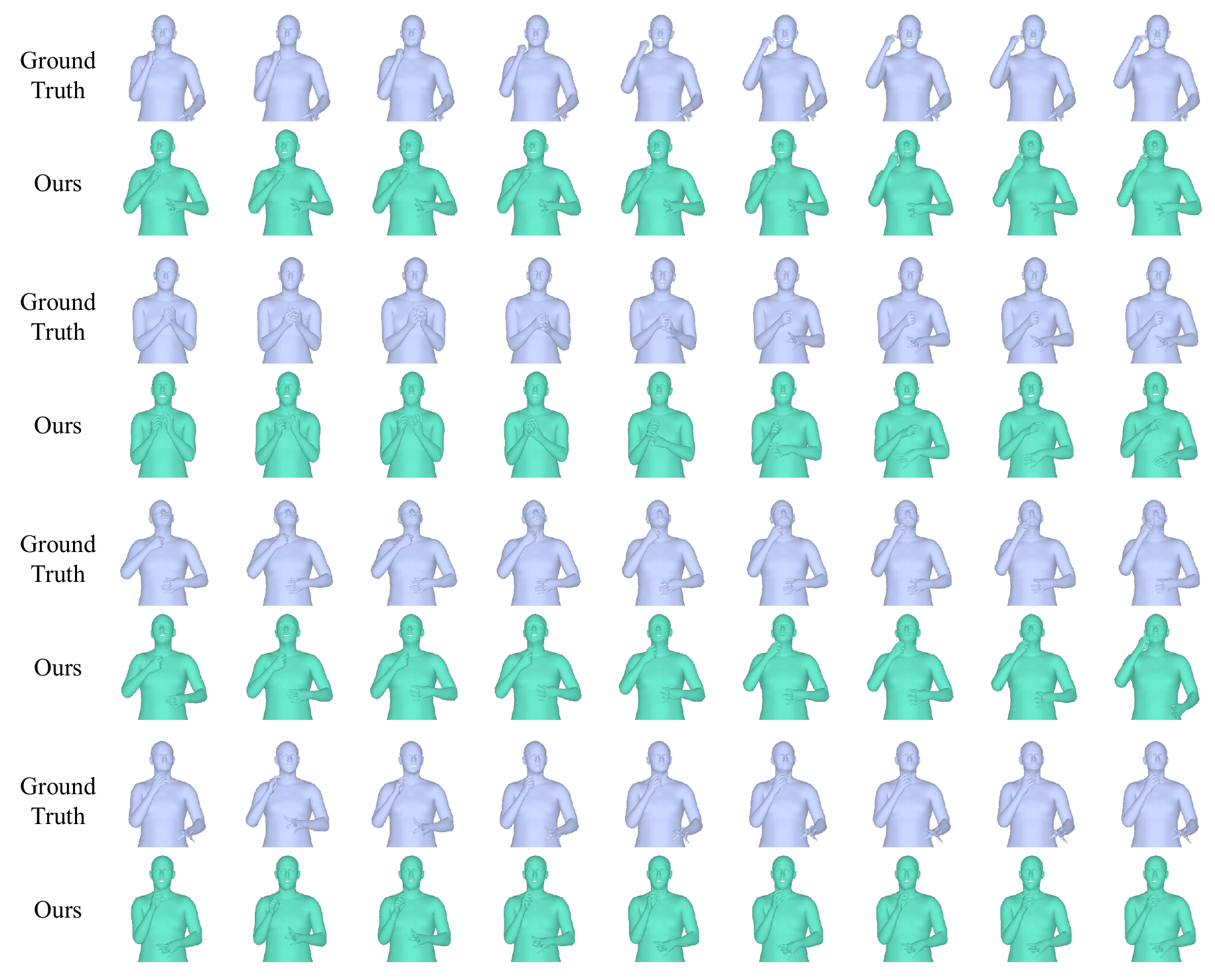}
    \caption{Qualitative results of \emph{BRAID}. We compare the synthesized motions of our model (green) with the ground truth sequences (blue), demonstrating that our method accurately generates realistic and highly aligned sign language gestures.}
    \label{fig:suppl_braid_qual_results}
\end{figure}
\namefig{}~\ref{fig:suppl_braid_qual_results} shows qualitative comparisons between the generated motions from \emph{BRAID} and the ground-truth continuous signing sequences. Overall, the generated motions closely follow the reference sequences in terms of body posture, arm configuration, and hand trajectory. Across different examples, our model preserves the relative position of the dominant hand with respect to the torso and face, while also producing temporally coherent transitions between neighboring frames. These results suggest that \emph{BRAID} can effectively refine the boundary region between isolated signs and recover continuous signing motions that are well aligned with the ground truth.

At the same time, we observe minor artifacts in some cases, particularly around the non-dominant hand. For example, the overall arm position remains consistent with the reference, but fine-grained hand shape or finger articulation can be slightly distorted. This indicates that while \emph{BRAID} captures the global motion structure and co-articulatory transition well, accurately preserving detailed hand geometry remains a challenging aspect of 3D sign motion generation.

\input{tables/suppl_braid_ablation}

\subsection{Spoken Language to Gloss}
\label{sec:suppl_experiments_t2g}
\noindent \textbf{Qualitative Results.}
\input{tables/suppl_t2g_qual}
\namefig{}~\ref{fig:qual_translator} presents qualitative examples of spoken-language-to-gloss translation. The model generally preserves the core meaning of the spoken sentence while producing compact gloss sequences. In the first example, it preserves the wh-question structure but realizes the reference \texttt{DO-DO} as \texttt{WHAT}, showing that plausible gloss alternatives may differ from the reference convention. In the second example, it omits the discourse-dependent marker \texttt{IX-honorific} while retaining the main meaning of ``my mother is still working.'' In the third example, the model correctly captures the conditional structure and event semantics, but generates \texttt{FIRE IX-1p} instead of the reference \texttt{FIRED IX-1p}. This indicates that the translator may favor a surface-aligned verbal form when the reference uses a convention-specific gloss form. Overall, the results suggest that our translator captures propositional meaning reliably, while fine-grained gloss conventions and context-dependent markers remain challenging.

\subsection{Sign-to-Sign Conversation}
\subsubsection{Retrieval-Based Semantic Evaluation Method}
We further evaluate whether sentence-level sign motions can recover their corresponding semantic annotations through the proposed hybrid retrieval protocol. This experiment is conducted on ground-truth assistant test motions as queries, and serves as an upper-bound calibration of the retrieval-based evaluator. Specifically, it measures how often the evaluator retrieves the corresponding reference sentence from the \textsc{SignaVox-U} memory and how closely the selected hybrid candidate matches the reference gloss annotation.

\paragraph{Evaluation metrics.}
We report retrieval accuracy using Recall@\(k\) and mean reciprocal rank (MRR). Recall@\(k\) measures whether the ground-truth sentence appears within the top-\(k\) retrieved candidates, and MRR measures the average reciprocal rank of the ground-truth sentence. We report these as R@1, R@5, R@10, and MRR. To evaluate semantic recoverability at the gloss level, we compare the gloss annotation of the final hybrid retrieval result with the reference gloss sequence. We report token-level F1, BLEU-4, and chrF, denoted as Gloss F1, Gloss BLEU-4, and Gloss chrF, respectively.

\paragraph{Results.}
Table~\ref{tab:hybrid_retrieval_upper_bound} shows the upper-bound calibration results on ground-truth assistant test motions. The evaluator retrieves the exact reference sentence at rank 1 for 41.5\% of the queries, and the recall increases to 58.4\% and 64.4\% within the top-5 and top-10 candidates, respectively. The MRR of 0.495 indicates that the corresponding reference sentence is often ranked near the top of the retrieval list. The final hybrid retrieval output also achieves 0.479 Gloss F1, 0.305 Gloss BLEU-4, and 0.498 Gloss chrF, showing that the retrieved candidates preserve a meaningful amount of gloss-level semantic information. These results provide a calibrated reference point for applying the same evaluator to generated \textsc{SignaVox} sentence motions.

\subsubsection{Qualitative Results}
\begin{figure}
    \centering
    \includegraphics[width=\linewidth]{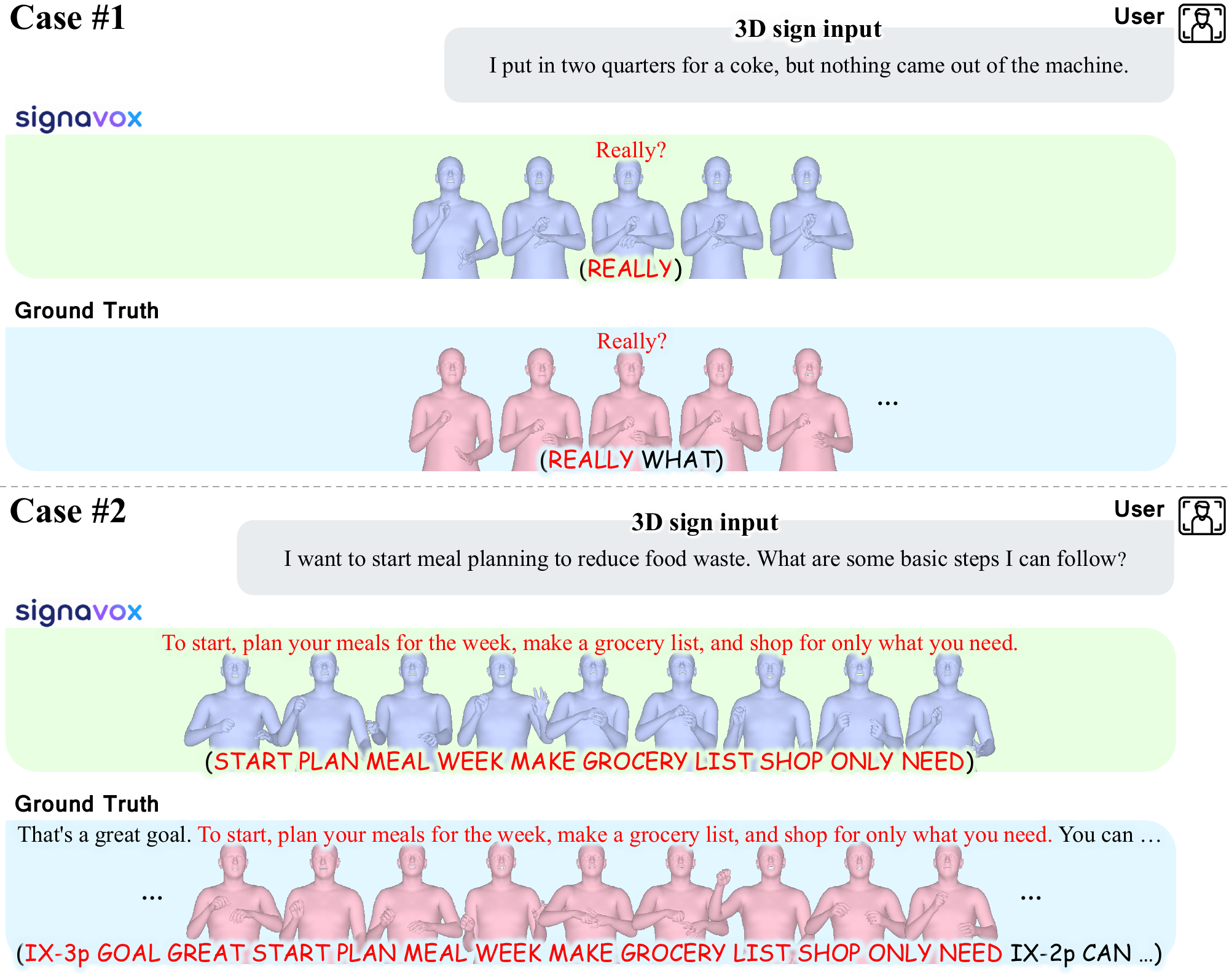}
    \caption{Qualitative results of the \textsc{SignaVox} conversational model. We compare the generated 3D sign responses (blue) with the ground truth (pink) based on the user's input. Note that while the actual user input is provided as 3D sign features, it is displayed here as spoken language text for better readability. Additionally, the glosses corresponding to the \textsc{SignaVox} outputs are predicted by our retrieval model (in Sec~\ref{sec:retrieval_based_translation}).}
    \label{fig:signavox_qualitative}
\end{figure}

\namefig{}~\ref{fig:signavox_qualitative} presents qualitative examples of \textsc{SignaVox}. For interpretability, we visualize the glosses and text retrieved by our semantic proxy evaluation model above the generated motions; these annotations are not directly produced by \textsc{SignaVox}. In Case~\#1, the model generates a concise response corresponding to \texttt{REALLY}, which matches the initial reaction in the ground-truth response. The generated motion also exhibits a similar upper-body pose and hand configuration, suggesting that the model can produce appropriate short conversational feedback from the preceding sign context. In Case~\#2, the model generates a longer informative response to a meal-planning question. Although the output is not identical to the reference, it covers the same main semantic content, including planning meals, making a grocery list, and shopping only for necessary items. The motion sequence remains coherent across multiple signs and visually follows a similar gesture progression to the ground truth. These results indicate that \textsc{SignaVox} can generate contextually appropriate sign responses at different response lengths, from short reactions to longer instruction-like utterances.

%% file: tables/evaluation_frame_selection_gloss_level.tex
% \newcommand{\Lbranch}{\rule[0.1ex]{0.5pt}{1.2ex}\rule[0.1ex]{0.7em}{0.5pt}\,}

% \begin{table}[t]
%     \centering
%     \resizebox{\columnwidth}{!}{%
%     \setlength{\aboverulesep}{0pt}
%     \setlength{\belowrulesep}{0pt}
%     \renewcommand{\arraystretch}{1.3}
%     \begin{tabular}{lcccc}
%     \specialrule{0.8pt}{1pt}{1pt}
%     & \makecell{DTW\\MPJPE $\downarrow$}
%     & \makecell{DTW\\MPVPE $\downarrow$}
%     & FGD $\downarrow$
%     & \makecell{Length\\Ratio} \\
%     \specialrule{0.5pt}{1pt}{1pt}

%     Raw
%     & 0.1070 & 0.0710 & 22.12 & 18.10 \\
%     Motion-only
%     & 0.1005 & 0.0675 & 19.89 & 12.71 \\
%     VideoLLM-only
%     & 0.0931 & 0.0635 & 17.03 & 7.13 \\
%     \Lbranch Ours
%     & \textbf{0.0921} & \textbf{0.0629} & \textbf{16.2866} & 5.76 \\

%     \specialrule{0.8pt}{1pt}{1pt}
%     \end{tabular}
%     }
%     \caption{Ablation on frame selection.}
%     \label{tab:frame_selection2}
% \end{table}

\begin{wrapfigure}{r}{0.5\textwidth} % r: 우측 배치, 0.5\textwidth: 표의 너비
    \centering
    \vspace{-10pt} % 상단 여백 조절
     \captionof{table}{Evaluation of articulation-frame selection against ASLLRP gloss-level ground truth.}
    \resizebox{\linewidth}{!}{%
    \setlength{\aboverulesep}{0pt}
    \setlength{\belowrulesep}{0pt}
    \renewcommand{\arraystretch}{1.3}
    \begin{tabular}{lcccc}
    \specialrule{1pt}{1pt}{1pt}
    & \makecell{DTW\\MPJPE $\downarrow$}
    & \makecell{DTW\\MPVPE $\downarrow$}
    & FGD $\downarrow$
    & \makecell{Length\\Ratio} \\
    \specialrule{0.5pt}{1pt}{1pt}

    Raw
    & 0.1070 & 0.0710 & 22.12 & 18.10 \\
    Motion-only
    & 0.1005 & 0.0675 & 19.89 & 12.71 \\
    VideoLLM-only
    & 0.0931 & 0.0635 & 17.03 & 7.13 \\
    Ours % \Lbranch 대신 일반적인 기호 사용 가능
    & \textbf{0.0921} & \textbf{0.0629} & \textbf{16.29} & \textbf{5.76} \\

    \specialrule{0.8pt}{1pt}{1pt}
    \end{tabular}
    }
   
    \label{tab:frame_selection2}
    \vspace{-10pt} % 하단 여백 조절
\end{wrapfigure}

%% file: tables/suppl_braid_ablation.tex
% \begin{table}[t]
% \centering
% \footnotesize
% \setlength{\tabcolsep}{3.8pt}
% \renewcommand{\arraystretch}{1.0}

% \begin{tabular}{@{}c c c !{\vrule width 0.5pt} c c@{}}
% \specialrule{1pt}{1pt}{1pt}
% Min-SNR & Inpaint & EMA & DTW-MPJPE $\downarrow$ & DTW-MPVPE $\downarrow$ \\
% \specialrule{0.5pt}{1pt}{1pt}
%               &               &               & 0.0698     &  0.4685   \\
% \mytablecheck& &      &  0.0708     &  0.0476    \\
% &\mytablecheck&      &  0.0670     &  0.0881    \\
% &    &  \mytablecheck    &  0.0716     &  0.0493    \\
%  &  \mytablecheck   &  \mytablecheck    &  0.0695     &  0.0472    \\
% \mytablecheck &  &  \mytablecheck    &0.0670&  0.0462  \\
% \mytablecheck & \mytablecheck & \mytablecheck & \textbf{0.0653} & \textbf{0.0439} \\
% \specialrule{0.8pt}{1pt}{1pt}
% \end{tabular}

% \captionsetup{skip=4pt}
% \caption{Ablation study on \emph{BRAID} training components.}
% \label{tab:suppl_braid_ablation}
% \end{table}

\begin{table*}[t]
\centering
% --- 첫 번째 테이블 (Ablation Study) ---
\begin{minipage}[t]{0.5\textwidth}
    \centering
    \caption{Ablation study on \emph{BRAID} training components.}
    \label{tab:suppl_braid_ablation}
    \footnotesize
    \setlength{\tabcolsep}{3pt}
    % 행 개수가 많으므로 간격을 1.15로 설정
    \renewcommand{\arraystretch}{1.05}
    \begin{tabular}{@{}c c c !{\vrule width 0.5pt} c c@{}}
    \specialrule{1pt}{1pt}{1pt}
    Min-SNR & Inpaint & EMA & DTW-MPJPE $\downarrow$ & DTW-MPVPE $\downarrow$ \\
    \specialrule{0.5pt}{1pt}{1pt}
                  &               &               & 0.0698     &  0.4685   \\
    \mytablecheck &               &               &  0.0708     &  0.0476    \\
                  & \mytablecheck &               &  0.0670     &  0.0881    \\
                  &               &  \mytablecheck    &  0.0716     &  0.0493    \\
                  &  \mytablecheck   &  \mytablecheck    &  0.0695     &  0.0472    \\
    \mytablecheck &               &  \mytablecheck    & 0.0670      &  0.0462  \\
    \mytablecheck & \mytablecheck & \mytablecheck & \textbf{0.0653} & \textbf{0.0439} \\
    \specialrule{0.8pt}{1pt}{1pt}
    \end{tabular}
\end{minipage}
\hfill % 두 테이블 사이의 간격을 최대한 벌림
% --- 두 번째 테이블 (Hand Configurations) ---
\begin{minipage}[t]{0.48\textwidth}
    \centering
    \caption{Quantitative results with different hand depth.}
    \label{tab:hand_ablation_results}
    \setlength{\tabcolsep}{2pt} 
    % 행 개수가 적으므로 간격을 1.48로 높여 첫 번째 테이블과 높이를 맞춤
    \renewcommand{\arraystretch}{1.48}
    \resizebox{0.96\linewidth}{!}{%
    \begin{tabular}{l ccc cccc}
    \specialrule{1pt}{1pt}{1pt}
    \multirow{2}{*}{} & \multicolumn{3}{c}{DTW-MPJPE $\downarrow$} & \multicolumn{4}{c}{DTW-MPVPE $\downarrow$} \\
    \cmidrule(lr){2-4} \cmidrule(lr){5-8}
     & Body & Hand & Overall  & Body & Hand & Face & Overall \\
    \specialrule{0.5pt}{1pt}{1pt}
    $\times 1$ & 0.0414 & 0.14544 &0.0693& 0.0336 & 0.0178  &0.0017 & 0.0470 \\
    $\times 2$ &0.0427& 0.1500 &0.0702& 0.0341 & 0.0185  &0.0017& 0.0478 \\
    $\times 3$ & 0.0421 & 0.14679 &0.0691& 0.0336& 0.0189  &0.0017& 0.0470 \\
    \rowcolor{hansayellow!20}$\times 4$ & \textbf{0.0376} & \textbf{0.1384} &\textbf{0.0653}& \textbf{0.0309}& \textbf{0.0171}  &\textbf{0.0017}& \textbf{0.0439} \\
    % \rowcolor{hansayellow!20} \textbf{$\times 4$} & \textbf{0.0376}  \textbf{0.1384} & \textbf{0.0309} & &\textbf{0.0171} & \textbf{0.0017} & \\
    $\times 5$ & 0.0419 & 0.14632 &0.0691 & 0.0335 & 0.0193  &0.0017& 0.0469 \\
    % \specialrule{0.5pt}{1pt}{1pt}
    \specialrule{0.8pt}{1pt}{1pt}
    \end{tabular}%
    }
\end{minipage}
\end{table*}

% \begin{table}[t]
% \centering
% \footnotesize
% \setlength{\tabcolsep}{4.5pt}
% \renewcommand{\arraystretch}{1.0}
% \begin{tabular}{cccccc}
% \toprule
% \multicolumn{4}{c}{Settings} & \multicolumn{2}{c}{Metrics} \\
% \cmidrule(r){1-4} \cmidrule(l){5-6}
% D\_gloss & min snr & inpaint & EMA & DTW-MPJPE & DTW-MPVPE \\
% \midrule
% \mytablecheck &               &               &               &           &           \\
% \mytablecheck & \mytablecheck &               &               &           &           \\
% \mytablecheck &               & \mytablecheck &               &           &           \\
% \mytablecheck & \mytablecheck & \mytablecheck & \mytablecheck &           &           \\
% \bottomrule
% \end{tabular}
% \caption{Braid ablation.}
% \label{tab:braid_ablation}
% \end{table}

%% file: tables/suppl_t2g_qual.tex
\begin{figure}[t]
\centering
\small

\newcommand{\exampleline}{
    \par\vspace{2pt}
    \noindent\textcolor{black!25}{\rule{\linewidth}{0.4pt}}
    \vspace{2pt}
}

\begin{tcolorbox}[
    colback=gray!3,
    colframe=black!35,
    boxrule=0.4pt,
    arc=2pt,
    left=6pt,
    right=6pt,
    top=5pt,
    bottom=5pt,
    width=0.96\linewidth
]
\textbf{Spoken sentence:} Do you usually talk with Jessica and Donna? What? Why?

\vspace{2pt}
\textbf{GT:} \texttt{IX-2p TEND CHAT WITH JESSICA AND DONNA DO-DO WHY}

\vspace{2pt}
\textbf{Ours:} \texttt{IX-2p TEND CHAT WITH JESSICA AND DONNA WHAT WHY}

\exampleline

% \textbf{Spoken sentence:} If it happens to rain I won’t have an umbrella.

% \vspace{2pt}
% \textbf{GT:} \texttt{HAPPEN RAIN IX-1p NOT HAVE UMBRELLA}

% \vspace{2pt}
% \textbf{Ours:} \texttt{HAPPEN RAIN IX-1p NOT HAVE UMBRELLA}

% \exampleline

\textbf{Spoken sentence:} My mother is still working.

\vspace{2pt}
\textbf{GT:} \texttt{IX-honorific POSS-1p MOTHER STILL WORK}

\vspace{2pt}
\textbf{Ours:} \texttt{POSS-1p MOTHER STILL WORK}

\exampleline

\textbf{Spoken sentence:} If I arrive to work late, my boss will fire me.

\vspace{2pt}
\textbf{GT:} \texttt{IF IX-1p ARRIVE WORK LATE POSS-1p BOSS FUTURE FIRED IX-1p}

\vspace{2pt}
\textbf{Ours:} \texttt{IF IX-1p ARRIVE WORK LATE POSS-1p BOSS FUTURE FIRE IX-1p}

\end{tcolorbox}

\caption{
Qualitative examples of spoken-to-gloss translation.
}
\label{fig:qual_translator}
\end{figure}

%% file: suppl/LLM_prompt.tex
\section{LLM Prompt}
\label{sec:suppl_llm_prompt}
\subsection{VideoLLM Prompt for Articulation Frame Selection}
\label{sec:videollm_prompt}
The VideoLLM prompts used for articulation-frame selection are shown in \namefig{}~\ref{fig:stage0}, \ref{fig:stage1}, and \ref{fig:stage2}.
\input{tables/videollm_system_prompt}
\subsection{GPT Evaluation Prompt for Spoken to Gloss Translation}
We evaluate the quality of spoken-to-gloss translation using GPT-5.2~\cite{openai_gpt52}. Beyond automatic metrics, we also employ LLM-based evaluation to assess aspects that are difficult to measure with surface-level matching alone.
Specifically, the semantic score measures how well the generated ASL gloss preserves the meaning of the original spoken language utterance. The structure score evaluates how well the generated gloss follows the ASL grammatical ordering of Time--Topic--Comment. Both scores are rated on a five-point scale, and the evaluation prompt is shown in \namefig{}~\ref{fig:gpt_eval}.
\input{tables/gpt_eval_prompt}

\subsection{Spoken to Gloss Translation System Prompt}
The system prompt used for spoken-language-to-gloss translation is shown in \namefig{}~\ref{fig:gloss_rules_full}.

\clearpage
\input{tables/t2g_system_prompt}

%% file: tables/videollm_system_prompt.tex
\begin{figure}
    \begin{AIbox}{Single-Frame Selection for Fingerspelled Letters}
        \parbox[t]{0.99\textwidth}{
            \textbf{System Prompt} \\
            \scriptsize
            \begin{alltt}
                    You are given a temporally trimmed sign language video clip of a single fingerspelled alphabet letter. \\
                    \\ 
                    Select exactly one frame that best preserves the canonical handshape and visibility of the target letter. \\
                    Do not prefer preparation frames. \\
                    Do not prefer transition frames. \\
                    Do not prefer blurred motion frames unless no clearer articulation exists. \\
                    \\
                    Guidelines: \\
                    - Prefer the frame where the handshape is most canonical and visually stable. \\
                    - Avoid preparation, transition, and release frames if a clearer articulation frame exists. \\
                    - If several frames are nearly identical, choose the best single representative frame. \\
                    \\
                    Strict output rules:\\
                    - Output ONLY one integer: frame\_index \\
                    - No explanations \\
                    - No brackets \\
                    - No extra text \\
                    \\
                    Example output: \\
                    12\\
            \end{alltt}
            \normalsize
            \textbf{User Prompt}\\
            \scriptsize
            \begin{alltt}
                The target gloss label is ``\textcolor{blue}{gloss}''.\\
                This gloss is a single fingerspelled alphabet letter. \\
                Choose exactly one frame index for the clearest and most representative articulation of the letter ``\textcolor{blue}{gloss}''.\\
                \\
                Guidelines: \\
                    - Prefer the frame where the handshape is most canonical and visually stable. \\
                    - Avoid preparation, transition, and release frames if a clearer articulation frame exists.\\
                    - If several frames are nearly identical, choose the best single representative frame.\\
                \\
                Return only one integer: \\
                frame\_index\\
                \\
                Each frame contains a visible red label in the bottom-left corner formatted as `frame \#NN'. Use those visible frame indices exactly.
                The valid frame index range is 0 to \textcolor{blue}{frame\_count - 1}.
            \end{alltt}
            }
    \end{AIbox}
    \caption{Prompt used to judge single frame selection. \textcolor{blue}{Blue} text denotes input variables. }
    \label{fig:stage0}
\end{figure}

\begin{figure}
    \begin{AIbox}{Contiguous Span Selection for Core Lexical Articulation}
        \parbox[t]{0.99\textwidth}{
            \textbf{System Prompt} \\
            \scriptsize
            \begin{alltt}
            You are given a temporally trimmed sign language video clip.\\
            \\
            Your task is to identify the contiguous temporal span that best preserves the core lexical articulation of the sign.\\
            \\
            Select the earliest frame where the articulation begins and the latest frame needed to preserve the articulation.\\
            Do not include extra preparation frames before the articulation starts.\\
            Do not include idle or resting frames.\\
            Do not include repeated hold frames after the articulation is completed.\\
            
            Prefer the shortest contiguous span that still preserves the sign's lexical identity.\\
            Avoid selecting only a single frame unless the articulation is truly instantaneous.\\
            \\
            Stop the span immediately once the lexical articulation is completed.\\
            Do not include frames that only maintain the final pose.\\
            \\
            Strict output rules:\\
            - Output ONLY two integers: start\_frame,end\_frame\\
            - No explanations\\
            - No brackets\\
            - No extra text\\
            \\
            Example output:\\
            12,16\\
            \end{alltt}

            \normalsize
            \textbf{User Prompt} \\
            \scriptsize
            \begin{alltt}
            The target gloss label is ``\textcolor{blue}{gloss}''.\\
            \\
            Use the gloss label only as semantic context. Base your decision on the visible articulation in the video.\\
            Select the start and end frame indices of the core lexical articulation for the sign ``\textcolor{blue}{gloss}''.\\
            \\
            Guidelines:\\
            - Include all frames necessary to preserve the handshape, orientation, and motion progression of the sign.\\
            - Return one contiguous articulation span, not scattered frames.\\
            - If the sign is mostly static, return the shortest span that still preserves the canonical articulation.\\
            - If the sign is defined by motion or direction change, include the full visible articulation phase rather than only the peak frame.\\
            - Stop the span as soon as the lexical articulation is completed.\\
            \\
            Return only two integers separated by a comma:\\
            start\_frame,end\_frame\\
            \\
            Each frame contains a visible red label in the bottom-left corner formatted as 'frame \#NN'. Use those visible frame indices exactly.\\
            The valid frame index range is 0 to \textcolor{blue}{frame\_count - 1}.
            \end{alltt}
        }
    \end{AIbox}
    \caption{Prompt for selecting the start and end boundaries of the core articulation in general sign language videos. \textcolor{blue}{Blue} text denotes input variables.}
    \label{fig:stage1}
\end{figure}

\begin{figure}[!ht]
    \begin{AIbox}{Boundary Refinement for Proposed Lexical Articulation Span}
        \parbox[t]{0.99\textwidth}{
            \textbf{System Prompt} \\
            \scriptsize
            \begin{alltt}
                    You are reviewing a first-pass temporal span for a sign language video clip. \\
                    \\
                    Your job is to correct the proposed span only if the boundaries are not tight enough. \\
                    \\
                    Be especially careful about unnecessary frames near the end of the span. \\
                    If the final part only maintains the completed sign, repeated hold, or resting posture, remove it. \\
                    \\
                    Check carefully: \\
                    - Is the proposed start too early because preparation frames are still included? \\
                    - Is the proposed end too late because repeated hold, resting, or post-articulation frames remain? \\
                    - Only move the end later if meaningful lexical articulation is clearly missing. \\
                    \\
                    Prefer the tightest contiguous span that still preserves the sign's lexical identity. \\
                    \\
                    Strict output rules: \\
                    - Output ONLY two integers: start\_frame,end\_frame \\
                    - No explanations \\
                    - No brackets \\
                    - No extra text \\
                    \\
                    Example output: \\
                    12,16 \\
            \end{alltt}

            \normalsize
            \textbf{User Prompt} \\
            \scriptsize
            \begin{alltt}
                    The target gloss label is "\textcolor{blue}{gloss}". \\
                    \\
                    A first-pass model predicted this articulation span: \\
                    start\_frame=\textcolor{blue}{candidate\_start}, end\_frame=\textcolor{blue}{candidate\_end} \\
                    \\
                    You are now shown a candidate-focused clip covering frames \textcolor{blue}{focus\_start} to \textcolor{blue}{focus\_end}. \\
                    The visible frame labels are still the original global frame indices from the full trimmed clip. \\
                    \\
                    Review this focused clip and tighten the proposal if needed. \\
                    \\
                    Your main goal is to remove unnecessary late frames. \\
                    Prefer trimming borderline late frames unless they clearly contribute new lexical information. \\
                    \\
                    Questions to check: \\
                    - Does the start still include preparation that should be removed? \\
                    - Does the end still include repeated hold, resting, or post-articulation frames that should be removed? \\
                    - Does the end include frames that only maintain the completed sign? \\
                    - Only extend the span if lexical articulation is clearly missing. \\
                    \\
                    Return the final corrected span as: \\
                    start\_frame,end\_frame \\
                    \\
                    Each frame contains a visible red label in the bottom-left corner formatted as 'frame \#NN'. Use those visible frame indices exactly. \\
                    The valid frame index range is \textcolor{blue}{focus\_start} to \textcolor{blue}{focus\_end}. \\
            \end{alltt}
        }
    \end{AIbox}
    \caption{Prompt for refining the start and end boundaries of the core articulation in sign language videos based on a candidate span. \textcolor{blue}{Blue} text denotes input variables.}
    \label{fig:stage2}
\end{figure}

%% file: tables/gpt_eval_prompt.tex
\begin{figure}[!ht]
    \begin{AIbox}{GPT Evaluation Prompt}
        \parbox[t]{0.99\textwidth}{
            \textbf{System Prompt} \\
            \scriptsize
            \begin{alltt}
                You are a strict evaluator for English $\rightarrow$ ASL gloss prediction quality.\\
                This is a reference-based evaluation. Compare predicted gloss to reference gloss.\\
                Evaluate with exactly two perspectives.\\
                Return JSON only with this schema:\\
                \{``semantic\_score'': <integer 1..5>, ``ttc\_structure\_score'': <integer 1..5>, \} \\
                Guidelines:\\
                - semantic\_score: 1 means meaning is mostly wrong, 5 means meaning is fully preserved.\\
                - ttc\_structure\_score: 1 means poor ASL Time-Topic-Comment ordering, 5 means strong TTC ordering.\\
                - Judge structure as ASL gloss grammar quality, not English word order.\\
            \end{alltt}

            \normalsize
            \textbf{User Prompt} \\
            \scriptsize
            \begin{alltt}
                English sentence: \\
                \textcolor{blue}{spoken language} \\
                \\
                Reference ASL gloss:\\
                \textcolor{blue}{Reference gloss}\\
                \\
                Predicted ASL gloss:\\
                \textcolor{blue}{Predicted gloss}\\
                
                Evaluate now and return JSON only.
            \end{alltt}
        }
    \end{AIbox}
    \caption{Prompt used to evaluate spoken-to-gloss translation quality with GPT-5.2. \textcolor{blue}{Blue} text denotes input variables.}
    \label{fig:gpt_eval}
\end{figure}

%% file: tables/t2g_system_prompt.tex
\begin{AIbox}[breakable, left=5pt, right=5pt, top=8pt, bottom=8pt]{English-to-ASL Gloss Translation Prompt} 
    \scriptsize
    \begin{alltt}
GRAMMAR RULES (for English \(\rightarrow\) ASL Gloss)

0) Output format constraints (important)
- Output ONLY the gloss tokens in ONE LINE.
- Tokens are \textbf{space-separated}.
- Use \textbf{UPPERCASE} for content signs whenever possible (e.g., GO, SEE, HOUSE, EXCITED).
- Keep the dataset-style tokens when needed:
  - \textbf{Pronouns / deixis:} IX-1p, IX-2p, IX-3p, IX-loc
  - \textbf{Possessives:} POSS-1p, POSS-2p, POSS-3p
  - \textbf{Fingerspelling / names:} fs-A-B-C..., ns-..., ns-fs-A-B-C..., WORD (loan sign)
- Words/expressions in quotes (" ") are often \textbf{names or special notation}.  
  When appropriate, prefer dataset-style name/spelling tokens.
- Avoid adding non-manual markers or grammatical annotations that are not in the simplified gloss set.

1) Prefer the canonical ASL ordering: Time + Topic + Comment (important)
- \textbf{Time}: put time indicators first 
  (YESTERDAY, TODAY, TOMORROW, PASTNIGHT, NOW, RECENT-PAST, FINISH, BEFORE, AFTER).
- \textbf{Topic}: place the main noun/theme early (often the object or scene-setting element).
- \textbf{Comment}: what is said about the topic (typically includes the verb and predicate).

Examples:
- "I went to the library yesterday." \(\rightarrow\) YESTERDAY LIBRARY IX-1p GO-TO
- "My friend, I see them." \(\rightarrow\) FRIEND IX-1p SEE IX-3p

2) Topicalization (Topic/Comment) and adjective ordering
- Topics come first. Any description of the topic (adjectives, attributes) comes before the comment.
- Order descriptive details in a natural ASL-like sequence (often: category + attribute + size/degree).
  Example:
- "I see a big orange cat." \(\rightarrow\) CAT ORANGE BIG IX-1p SEE

3) Visual/causal/sequence organization: sign in the order you "see it"
ASL often prefers ordering that matches visualization:
3.1 Cause \(\rightarrow\) Effect
- Express the \textbf{cause} before the \textbf{effect}.
  Example:
- "I feel calm when I go to the park." \(\rightarrow\) PARK GO-TO FEEL CALM IX-1p
3.2 Real-time sequencing (chronological order)
- Arrange events in the order they happen.
  Example:
- "I'm worried because my brother didn't call me after he left."
  \(\rightarrow\) POSS-1p BROTHER LEAVE CALL-BY-PHONE-1p NOT CONCERN IX-1p
3.3 General \(\rightarrow\) Specific (scene setting)
- Establish the broad setting first, then narrow down to details.
  Example:
- "I am excited after moving to my new house in Virginia."
  \(\rightarrow\) VIRGINIA HOUSE NEW MOVE FINISH EXCITED IX-1p

4) Tense: verbs stay base form; tense is set by time markers
- Do NOT conjugate verbs by tense (EAT covers ate/eats/eating/eaten).
- Use time indicators at the beginning to establish time.
- Use FINISH (or similar) to indicate completed/past events when appropriate.

5) Questions: WH-words tend to go at the end
- Place WHO / WHAT / WHEN / WHERE / WHY / WHICH / HOW at the \textbf{end} of the sentence.
- If emphasis is needed, WH-word may appear at both beginning and end, but default is end.
  Example:
- "What is your name?" \(\rightarrow\) POSS-2p NAME WHAT

6) Copula deletion: do not include "to be"
- English "am/is/are/was/were" is often omitted in ASL gloss.
  Example:
- "He is tall." \(\rightarrow\) IX-3p TALL

7) Negation: NOT/NONE usually follows what it negates
- Negative signs often come after the verb or phrase they negate.
  Example:
- "I don't have any pets." \(\rightarrow\) PET HAVE NOT

8) Indexing and referents (use IX- / POSS- consistently)
- Use IX-1p / IX-2p / IX-3p to represent pronouns (I/you/he-she-they).
- Use POSS-1p / POSS-2p / POSS-3p to represent possessives (my/your/his-her-their).
- Use IX-loc to point to locations when needed (there/here/that place).
- In this simplified gloss style, do NOT add locus indices like ":i/:j"; prefer IX-3p and IX-loc.

9) Directional/agreement verbs (simplified)
- Some meanings are naturally expressed with directional verbs (
  GIVE, TELL, ASK, SEND/MAIL, CALL-BY-PHONE).
- In simplified gloss, keep the verb as a single token (e.g., CALL-BY-PHONE) 
  and represent participants with IX/POSS nearby if needed.

10) Fingerspelling / names / loan signs
- If a proper noun (person/place/organization) has no known lexical sign, use:
  - ns-fs-J-O-H-N / ns-fs-M-A-R-Y for name-like entities (common in datasets)
  - fs-... for general fingerspelling
- Use WORD for common loan signs (dataset style), when appropriate.
- Use fingerspelling (fs-...) to emphasize the exact spelling of a word, 
  especially when you want the viewer to notice the letters clearly.
- For common English words that have no known lexical ASL sign
  - Example: ventriloquism \(\rightarrow\) fs-V-E-N-T-R-I-L-O-Q-U-I-S-M

11) Keep it dataset-like: prioritize learnable, consistent gloss
- Prefer common/high-frequency gloss tokens over rare stylistic paraphrases.
- If multiple ASL word orders are possible, prefer the one that matches:
  1. Time first, then topic, then comment
  2. Cause\(\rightarrow\)Effect
  3. General\(\rightarrow\)Specific
- Avoid adding extra functional markers unless they are represented in the simplified gloss vocabulary.

OUTPUT REQUIREMENT:

Given an English sentence, output \textbf{ONLY} the ASL gloss on a single line.
    \end{alltt}
\end{AIbox}

\captionof{figure}{Full system prompt for converting English sentences to ASL gloss, incorporating all grammatical and formatting rules. \textcolor{blue}{Blue} text indicates input variables.}
\label{fig:gloss_rules_full}

%% file: main.bib
@String(BMVC  = {Brit. Mach. Vis. Conf.})

@String(AAAI  = {AAAI})

@String(TOG   = {ACM Trans. Graph.})

@String(BMVC  =	{BMVC})

@String(TOG   = {ACM TOG})

@book{padden1988:deaf,
  title={Deaf in America: Voices from a culture},
  author={Padden, Carol A and Humphries, Tom L},
  year={1988},
  publisher={Harvard University Press}
}

@article{stokoe1980:sign,
  title={Sign language structure},
  author={Stokoe, William C},
  journal={Annual review of anthropology},
  pages={365--390},
  year={1980},
  publisher={JSTOR}
}

@article{sign_early1,
  title={What you don’t know can hurt you: The risk of language deprivation by impairing sign language development in deaf children},
  author={Hall, Wyatte C},
  journal={Maternal and child health journal},
  volume={21},
  number={5},
  pages={961--965},
  year={2017},
  publisher={Springer}
}

@article{parent,
  title={Chasing the mythical ten percent: Parental hearing status of deaf and hard of hearing students in the United States},
  author={Mitchell, Ross E and Karchmer, Michael A},
  journal={Sign language studies},
  volume={4},
  number={2},
  pages={138--163},
  year={2004},
  publisher={Gallaudet University Press}
}

@article{berger2024parent,
  title={Parent American Sign Language skills correlate with child--but not toddler--ASL vocabulary size},
  author={Berger, Lauren and Pyers, Jennie and Lieberman, Amy and Caselli, Naomi},
  journal={Language Acquisition},
  volume={31},
  number={2},
  pages={85--99},
  year={2024},
  publisher={Taylor \& Francis}
}

@article{morford2011deaf,
  title={When deaf signers read English: Do written words activate their sign translations?},
  author={Morford, Jill P and Wilkinson, Erin and Villwock, Agnes and Pi{\~n}ar, Pilar and Kroll, Judith F},
  journal={Cognition},
  volume={118},
  number={2},
  pages={286--292},
  year={2011},
  publisher={Elsevier}
}

@article{morford2014bilingual,
  title={Bilingual word recognition in deaf and hearing signers: Effects of proficiency and language dominance on cross-language activation},
  author={Morford, Jill P and Kroll, Judith F and Pi{\~n}ar, Pilar and Wilkinson, Erin},
  journal={Second Language Research},
  volume={30},
  number={2},
  pages={251--271},
  year={2014},
  publisher={SAGE Publications Sage UK: London, England}
}

@article{goodwin2023deaf,
  title={Deaf and hearing American Sign Language--English bilinguals: Typical bilingual language development},
  author={Goodwin, Corina and Lillo-Martin, Diane},
  journal={Journal of Deaf Studies and Deaf Education},
  volume={28},
  number={4},
  pages={350--362},
  year={2023},
  publisher={Oxford University Press}
}

@article{lederberg2013language,
  title={Language and literacy development of deaf and hard-of-hearing children: successes and challenges.},
  author={Lederberg, Amy R and Schick, Brenda and Spencer, Patricia E},
  journal={Developmental psychology},
  volume={49},
  number={1},
  pages={15},
  year={2013},
  publisher={American Psychological Association}
}

@article{lederberg2014foundations,
  title={Foundations for literacy: An early literacy intervention for deaf and hard-of-hearing children},
  author={Lederberg, Amy R and Miller, Elizabeth M and Easterbrooks, Susan R and Connor, Carol McDonald},
  journal={Journal of deaf studies and deaf education},
  volume={19},
  number={4},
  pages={438--455},
  year={2014},
  publisher={Oxford University Press UK}
}

@inproceedings{emocav2,
  title={Spectre: Visual speech-informed perceptual 3d facial expression reconstruction from videos},
  author={Filntisis, Panagiotis P and Retsinas, George and Paraperas-Papantoniou, Foivos and Katsamanis, Athanasios and Roussos, Anastasios and Maragos, Petros},
  booktitle={Proceedings of the IEEE/CVF conference on computer vision and pattern recognition},
  pages={5745--5755},
  year={2023}
}

@inproceedings{smplerx,
    title={{SMPLer-X}: Scaling up expressive human pose and shape estimation},
    author={Cai, Zhongang and Yin, Wanqi and Zeng, Ailing and Wei, Chen and Sun, Qingping and Yanjun, Wang and Pang, Hui En and Mei, Haiyi and Zhang, Mingyuan and Zhang, Lei and Loy, Chen Change and Yang, Lei and Liu, Ziwei},
    booktitle={Advances in Neural Information Processing Systems},
    year={2023}
}

@inproceedings{smpl-x,
  title={Expressive body capture: 3d hands, face, and body from a single image},
  author={Pavlakos, Georgios and Choutas, Vasileios and Ghorbani, Nima and Bolkart, Timo and Osman, Ahmed AA and Tzionas, Dimitrios and Black, Michael J},
  booktitle={Proceedings of the IEEE/CVF conference on computer vision and pattern recognition},
  pages={10975--10985},
  year={2019}
}

@article{flame,
  title={Learning a model of facial shape and expression from 4D scans.},
  author={Li, Tianye and Bolkart, Timo and Black, Michael J and Li, Hao and Romero, Javier},
  journal={ACM Trans. Graph.},
  volume={36},
  number={6},
  pages={194--1},
  year={2017}
}

@inproceedings{mano,
  author    = {J. Romero and Dimitrios Tzionas and Michael J. Black},
  year      = {2017},
  title     = {Embodied hands},
  booktitle = {ACM Transactions on Graphics},
  doi       = {10.1145/3130800.3130883},
}

@inproceedings{dynhamr,
  title={Dyn-hamr: Recovering 4d interacting hand motion from a dynamic camera},
  author={Yu, Zhengdi and Zafeiriou, Stefanos and Birdal, Tolga},
  booktitle={Proceedings of the Computer Vision and Pattern Recognition Conference},
  pages={27716--27726},
  year={2025}
}

@inproceedings{bragg2019:sign,
  title={Sign language recognition, generation, and translation: An interdisciplinary perspective},
  author={Bragg, Danielle and Koller, Oscar and Bellard, Mary and Berke, Larwan and Boudreault, Patrick and Braffort, Annelies and Caselli, Naomi and Huenerfauth, Matt and Kacorri, Hernisa and Verhoef, Tessa and others},
  booktitle={Proceedings of the 21st international ACM SIGACCESS conference on computers and accessibility},
  pages={16--31},
  year={2019}
}

@inproceedings{signavatars,
  title={Signavatars: A large-scale 3d sign language holistic motion dataset and benchmark},
  author={Yu, Zhengdi and Huang, Shaoli and Cheng, Yongkang and Birdal, Tolga},
  booktitle={European Conference on Computer Vision},
  pages={1--19},
  year={2024},
  organization={Springer}
}

@inproceedings{wlasl,
  title={Word-level deep sign language recognition from video: A new large-scale dataset and methods comparison},
  author={Li, Dongxu and Rodriguez, Cristian and Yu, Xin and Li, Hongdong},
  booktitle={Proceedings of the IEEE/CVF winter conference on applications of computer vision},
  pages={1459--1469},
  year={2020}
}

@misc{signasl,
  author       = {{Sign ASL}},
  title        = {Sign {ASL}: {A}n {A}merican {S}ign {L}anguage {D}ictionary},
  howpublished = {\url{https://www.signasl.org/}},
  year         = {2026},
  note         = {Accessed: 2026-03-01}
}

@inproceedings{msasl,
    author = {Vaezi Joze, Hamid and Koller, Oscar},
    title = {MS-ASL: A Large-Scale Data Set and Benchmark for Understanding American Sign Language},
    booktitle = {The British Machine Vision Conference (BMVC)},
    year = {2019},
    month = {September}
}

@inproceedings{how2sign,
  title={How2sign: a large-scale multimodal dataset for continuous american sign language},
  author={Duarte, Amanda and Palaskar, Shruti and Ventura, Lucas and Ghadiyaram, Deepti and DeHaan, Kenneth and Metze, Florian and Torres, Jordi and Giro-i-Nieto, Xavier},
  booktitle={Proceedings of the IEEE/CVF conference on computer vision and pattern recognition},
  pages={2735--2744},
  year={2021}
}

@inproceedings{openasl,
  title={Open-domain sign language translation learned from online video},
  author={Shi, Bowen and Brentari, Diane and Shakhnarovich, Gregory and Livescu, Karen},
  booktitle={Proceedings of the 2022 Conference on Empirical Methods in Natural Language Processing},
  pages={6365--6379},
  year={2022}
}

@article{youtubeasl,
  title={Youtube-asl: A large-scale, open-domain american sign language-english parallel corpus},
  author={Uthus, Dave and Tanzer, Garrett and Georg, Manfred},
  journal={Advances in Neural Information Processing Systems},
  volume={36},
  pages={29029--29047},
  year={2023}
}

@article{asllex,
  title={ASL-LEX: A lexical database of American Sign Language},
  author={Caselli, Naomi K and Sehyr, Zed Sevcikova and Cohen-Goldberg, Ariel M and Emmorey, Karen},
  journal={Behavior research methods},
  volume={49},
  number={2},
  pages={784--801},
  year={2017},
  publisher={Springer}
}

@inproceedings{asllvd,
  title={The american sign language lexicon video dataset},
  author={Athitsos, Vassilis and Neidle, Carol and Sclaroff, Stan and Nash, Joan and Stefan, Alexandra and Yuan, Quan and Thangali, Ashwin},
  booktitle={2008 IEEE computer society conference on computer vision and pattern recognition workshops},
  pages={1--8},
  year={2008},
  organization={IEEE}
}

@misc{signingsavvy,
  author       = {Signing Savvy},
  title        = {Signing Savvy: {ASL} Sign Language Video Dictionary},
  year         = {2026},
  url          = {https://www.signingsavvy.com},
  note         = {Accessed: 2026-02-05}
}

@article{signbank,
  title={Building the ASL signbank. Lemmatization principles for ASL},
  author={Hochgesang, Julie and Crasborn, OA and Lillo-Martin, Diane},
  year={2018},
  publisher={Paris: ELDA},
  doi={10.6084/m9.figshare.9741788},
  url={http://aslsignbank.haskins.yale.edu}
}

@inproceedings{spreadthesign,
  title={A multilingual dictionary for sign languages:" spreadthesign"},
  author={Hilzensauer, Marlene and Krammer, Klaudia},
  booktitle={ICERI2015 Proceedings},
  pages={7826--7834},
  year={2015},
  organization={IATED},
  url={https://spreadthesign.com}
}

@inproceedings{slt_based,
  title={Neural sign language translation},
  author={Camgoz, Necati Cihan and Hadfield, Simon and Koller, Oscar and Ney, Hermann and Bowden, Richard},
  booktitle={Proceedings of the IEEE conference on computer vision and pattern recognition},
  pages={7784--7793},
  year={2018}
}

@inproceedings{coarticulation,
  title={Signing at scale: Learning to co-articulate signs for large-scale photo-realistic sign language production},
  author={Saunders, Ben and Camgoz, Necati Cihan and Bowden, Richard},
  booktitle={Proceedings of the IEEE/CVF Conference on Computer Vision and Pattern Recognition},
  pages={5141--5151},
  year={2022}
}

@article{ksl,
  title={Neural sign language translation based on human keypoint estimation},
  author={Ko, Sang-Ki and Kim, Chang Jo and Jung, Hyedong and Cho, Choongsang},
  journal={Applied sciences},
  volume={9},
  number={13},
  pages={2683},
  year={2019},
  publisher={MDPI}
}

@inproceedings{phoenix12,
  title={RWTH-PHOENIX-weather: A large vocabulary sign language recognition and translation corpus.},
  author={Forster, Jens and Schmidt, Christoph and Hoyoux, Thomas and Koller, Oscar and Zelle, Uwe and Piater, Justus H and Ney, Hermann},
  booktitle={LREC},
  volume={9},
  pages={3785--3789},
  year={2012}
}

@inproceedings{phoenix14,
  title={Extensions of the Sign Language Recognition and Translation Corpus RWTH-PHOENIX-Weather.},
  author={Forster, Jens and Schmidt, Christoph and Koller, Oscar and Bellgardt, Martin and Ney, Hermann},
  booktitle={LREC},
  pages={1911--1916},
  year={2014}
}

@inproceedings{csl,
  title={Improving sign language translation with monolingual data by sign back-translation},
  author={Zhou, Hao and Zhou, Wengang and Qi, Weizhen and Pu, Junfu and Li, Houqiang},
  booktitle={Proceedings of the IEEE/CVF conference on computer vision and pattern recognition},
  pages={1316--1325},
  year={2021}
}

@article{multi_youtube,
  title={YouTube-SL-25: A Large-Scale, Open-Domain Multilingual Sign Language Parallel Corpus},
  author={Tanzer, Garrett and Zhang, Biao},
  journal={arXiv preprint arXiv:2407.11144},
  year={2024}
}

@article{twostream_slr,
  title={Two-stream network for sign language recognition and translation},
  author={Chen, Yutong and Zuo, Ronglai and Wei, Fangyun and Wu, Yu and Liu, Shujie and Mak, Brian},
  journal={Advances in Neural Information Processing Systems},
  volume={35},
  pages={17043--17056},
  year={2022}
}

@inproceedings{stmc,
  title={Spatial-temporal multi-cue network for continuous sign language recognition},
  author={Zhou, Hao and Zhou, Wengang and Zhou, Yun and Li, Houqiang},
  booktitle={Proceedings of the AAAI conference on artificial intelligence},
  volume={34},
  number={07},
  pages={13009--13016},
  year={2020}
}

@inproceedings{better_stmc,
  title={Better Sign Language Translation with STMC-Transformer},
  author={Yin, Kayo and Read, Jesse},
  booktitle={Proceedings of the 28th International Conference on Computational Linguistics},
  pages={5975--5989},
  year={2020}
}

@article{kym_slt,
  title={Preprocessing for keypoint-based sign language translation without glosses},
  author={Kim, Youngmin and Baek, Hyeongboo},
  journal={Sensors},
  volume={23},
  number={6},
  pages={3231},
  year={2023},
  publisher={MDPI}
}

@inproceedings{glossfree_slt,
  title={Gloss-free sign language translation: Improving from visual-language pretraining},
  author={Zhou, Benjia and Chen, Zhigang and Clap{\'e}s, Albert and Wan, Jun and Liang, Yanyan and Escalera, Sergio and Lei, Zhen and Zhang, Du},
  booktitle={Proceedings of the IEEE/CVF International Conference on Computer Vision},
  pages={20871--20881},
  year={2023}
}

@inproceedings{glossfree_slt2,
  title={Gloss-Free End-to-End Sign Language Translation},
  author={Lin, Kezhou and Wang, Xiaohan and Zhu, Linchao and Sun, Ke and Zhang, Bang and Yang, Yi},
  booktitle={Proceedings of the 61st Annual Meeting of the Association for Computational Linguistics (Volume 1: Long Papers)},
  pages={12904--12916},
  year={2023}
}

@inproceedings{simple_baseline,
  title={A simple baseline for spoken language to sign language translation with 3d avatars},
  author={Zuo, Ronglai and Wei, Fangyun and Chen, Zenggui and Mak, Brian and Yang, Jiaolong and Tong, Xin},
  booktitle={European Conference on Computer Vision},
  pages={36--54},
  year={2024},
  organization={Springer}
}

@inproceedings{slp,
  title={Sign language production using neural machine translation and generative adversarial networks},
  author={Stoll, Stephanie and Camg{\"o}z, Necati Cihan and Hadfield, Simon and Bowden, Richard},
  booktitle={Proceedings of the 29th British Machine Vision Conference (BMVC 2018)},
  year={2018},
  organization={British Machine Vision Association}
}

@inproceedings{nonverbal,
    title = "Speaking Beyond Language: A Large-Scale Multimodal Dataset for Learning Nonverbal Cues from Video-Grounded Dialogues",
    author = "Kim, Youngmin  and
      Chung, Jiwan  and
      Kim, Jisoo  and
      Lee, Sunghyun  and
      Lee, Sangkyu  and
      Kim, Junhyeok  and
      Yang, Cheoljong  and
      Yu, Youngjae",
    booktitle = "Proceedings of the 63rd Annual Meeting",
    month = jul,
    year = "2025",
    address = "Vienna, Austria",
    publisher = "Association for Computational Linguistics",
    pages = "2247--2265",
    ISBN = "979-8-89176-251-0",
}

@article{yolov3,
  title={Yolov3: An incremental improvement},
  author={Redmon, Joseph and Farhadi, Ali},
  journal={arXiv preprint arXiv:1804.02767},
  year={2018}
}

@software{yolov8,
author = {Jocher, Glenn and Qiu, Jing and Chaurasia, Ayush},
license = {AGPL-3.0},
month = jan,
title = {{Ultralytics YOLO}},
url = {https://github.com/ultralytics/ultralytics},
version = {8.0.0},
year = {2023}
}

@inproceedings{osnet,
  title={Omni-scale feature learning for person re-identification},
  author={Zhou, Kaiyang and Yang, Yongxin and Cavallaro, Andrea and Xiang, Tao},
  booktitle={Proceedings of the IEEE/CVF international conference on computer vision},
  pages={3702--3712},
  year={2019}
}

@book{sub-dtw,
  title={Information retrieval for music and motion},
  author={M{\"u}ller, Meinard},
  year={2007},
  publisher={Springer}
}

@article{slp_keypoint,
  title={Continuous 3d multi-channel sign language production via progressive transformers and mixture density networks},
  author={Saunders, Ben and Camgoz, Necati Cihan and Bowden, Richard},
  journal={International journal of computer vision},
  volume={129},
  number={7},
  pages={2113--2135},
  year={2021},
  publisher={Springer}
}

@inproceedings{progressive_slp,
  title={Progressive transformers for end-to-end sign language production},
  author={Saunders, Ben and Camgoz, Necati Cihan and Bowden, Richard},
  booktitle={European Conference on Computer Vision},
  pages={687--705},
  year={2020},
  organization={Springer}
}

@article{asllrp,
  title={Asl video corpora \& sign bank: Resources available through the american sign language linguistic research project (asllrp)},
  author={Neidle, Carol and Opoku, Augustine and Metaxas, Dimitris},
  journal={arXiv preprint arXiv:2201.07899},
  year={2022}
}

@inproceedings{nms_generation,
  title={Towards ai-driven sign language generation with non-manual markers},
  author={Zhang, Han and Shalev-Arkushin, Rotem and Baltatzis, Vasileios and Gillis, Connor and Laput, Gierad and Kushalnagar, Raja and Quandt, Lorna C and Findlater, Leah and Bedri, Abdelkareem and Lea, Colin},
  booktitle={Proceedings of the 2025 CHI Conference on Human Factors in Computing Systems},
  pages={1--26},
  year={2025}
}

@article{zhang2025:large,
  title={Large Sign Language Models: Toward 3D American Sign Language Translation},
  author={Zhang, Sen and He, Xiaoxiao and Liu, Di and Xia, Zhaoyang and Zhao, Mingyu and Tan, Chaowei and Li, Vivian and Liu, Bo and Metaxas, Dimitris N and Kapadia, Mubbasir},
  journal={arXiv preprint arXiv:2511.08535},
  year={2025}
}

@article{signgpt,
  title={Sign2GPT: Leveraging large language models for gloss-free sign language translation},
  author={Wong, Ryan and Camgoz, Necati Cihan and Bowden, Richard},
  journal={arXiv preprint arXiv:2405.04164},
  year={2024}
}

@inproceedings{signllm,
  title={Signllm: Sign language production large language models},
  author={Fang, Sen and Chen, Chen and Wang, Lei and Zheng, Ce and Sui, Chunyu and Tian, Yapeng},
  booktitle={Proceedings of the IEEE/CVF International Conference on Computer Vision},
  pages={6622--6634},
  year={2025}
}

@inproceedings{llm_good_translator,
  title={Llms are good sign language translators},
  author={Gong, Jia and Foo, Lin Geng and He, Yixuan and Rahmani, Hossein and Liu, Jun},
  booktitle={Proceedings of the IEEE/CVF conference on computer vision and pattern recognition},
  pages={18362--18372},
  year={2024}
}

@article{savagol,
  title={Smoothing and differentiation of data by simplified least squares procedures.},
  author={Savitzky, Abraham and Golay, Marcel JE},
  journal={Analytical chemistry},
  volume={36},
  number={8},
  pages={1627--1639},
  year={1964},
  publisher={ACS Publications}
}

@inproceedings{pseudogloss,
  title={Bridging Sign and Spoken Languages: Pseudo Gloss Generation for Sign Language Translation},
  author={Guo, Jianyuan and Li, Peike and Cohn, Trevor},
  booktitle={The Thirty-ninth Annual Conference on Neural Information Processing Systems}
}

@article{signstream3,
  title={A User's Guide to SignStream{\textregistered} 3},
  author={Neidle, Carol},
  journal={Boston, MA: American Sign Language Linguistic Research Project Report},
  number={16},
  year={2017}
}

@inproceedings{splade,
  title={Splade: Sparse lexical and expansion model for first stage ranking},
  author={Formal, Thibault and Piwowarski, Benjamin and Clinchant, St{\'e}phane},
  booktitle={Proceedings of the 44th International ACM SIGIR Conference on Research and Development in Information Retrieval},
  pages={2288--2292},
  year={2021}
}

@inproceedings{reranker,
  title={C-pack: Packed resources for general chinese embeddings},
  author={Xiao, Shitao and Liu, Zheng and Zhang, Peitian and Muennighoff, Niklas and Lian, Defu and Nie, Jian-Yun},
  booktitle={Proceedings of the 47th international ACM SIGIR conference on research and development in information retrieval},
  pages={641--649},
  year={2024}
}

@inproceedings{vp,
  title={What does clip know about a red circle? visual prompt engineering for vlms},
  author={Shtedritski, Aleksandar and Rupprecht, Christian and Vedaldi, Andrea},
  booktitle={Proceedings of the IEEE/CVF International Conference on Computer Vision},
  pages={11987--11997},
  year={2023}
}

@inproceedings{numberit,
  title={Number it: Temporal grounding videos like flipping manga},
  author={Wu, Yongliang and Hu, Xinting and Sun, Yuyang and Zhou, Yizhou and Zhu, Wenbo and Rao, Fengyun and Schiele, Bernt and Yang, Xu},
  booktitle={Proceedings of the Computer Vision and Pattern Recognition Conference},
  pages={13754--13765},
  year={2025}
}

@article{qwen3vl,
  title={Qwen3-vl technical report},
  author={Bai, Shuai and Cai, Yuxuan and Chen, Ruizhe and Chen, Keqin and Chen, Xionghui and Cheng, Zesen and Deng, Lianghao and Ding, Wei and Gao, Chang and Ge, Chunjiang and others},
  journal={arXiv preprint arXiv:2511.21631},
  year={2025}
}

@inproceedings{signd2c,
  title={Discrete to continuous: Generating smooth transition poses from sign language observations},
  author={Tang, Shengeng and He, Jiayi and Cheng, Lechao and Wu, Jingjing and Guo, Dan and Hong, Richang},
  booktitle={Proceedings of the Computer Vision and Pattern Recognition Conference},
  pages={3481--3491},
  year={2025}
}

@misc{openai_gpt52,
  author = {{OpenAI}},
  title  = {Introducing GPT-5.2},
  year   = {2025},
  url    = {https://openai.com/index/introducing-gpt-5-2/},
  note   = {Accessed: 2026-03-24}
}

@inproceedings{bleu,
  title={Bleu: a method for automatic evaluation of machine translation},
  author={Papineni, Kishore and Roukos, Salim and Ward, Todd and Zhu, Wei-Jing},
  booktitle={Proceedings of the 40th annual meeting of the Association for Computational Linguistics},
  pages={311--318},
  year={2002}
}

@inproceedings{scarebleu,
  title={A call for clarity in reporting BLEU scores},
  author={Post, Matt},
  booktitle={Proceedings of the third conference on machine translation: Research papers},
  pages={186--191},
  year={2018}
}

@inproceedings{chrf,
  title={chrF: character n-gram F-score for automatic MT evaluation},
  author={Popovi{\'c}, Maja},
  booktitle={Proceedings of the tenth workshop on statistical machine translation},
  pages={392--395},
  year={2015}
}

@article{realtalk,
  title={Realtalk: A 21-day real-world dataset for long-term conversation},
  author={Lee, Dong-Ho and Maharana, Adyasha and Pujara, Jay and Ren, Xiang and Barbieri, Francesco},
  journal={arXiv preprint arXiv:2502.13270},
  year={2025}
}

@misc{everydayconversations,
  author = {Hugging Face},
  title = {Everyday Conversations for LLMs},
  year = {2024},
  howpublished = {\url{https://huggingface.co/datasets/HuggingFaceTB/everyday-conversations-llama3.1-2k}}
}

@inproceedings{dailydialog,
  title={Dailydialog: A manually labelled multi-turn dialogue dataset},
  author={Li, Yanran and Su, Hui and Shen, Xiaoyu and Li, Wenjie and Cao, Ziqiang and Niu, Shuzi},
  booktitle={Proceedings of the Eighth International Joint Conference on Natural Language Processing (Volume 1: Long Papers)},
  pages={986--995},
  year={2017}
}

@book{bm25,
  title={The probabilistic relevance framework: BM25 and beyond},
  author={Robertson, Stephen and Zaragoza, Hugo},
  volume={4},
  year={2009},
  publisher={Now Publishers Inc}
}

@article{vikey,
  title={ViKey: Enhancing Temporal Understanding in Videos via Visual Prompting},
  author={Lee, Yeonkyung and Ju, Dayun and Kim, Youngmin and Kang, Seil and Hwang, Seong Jae},
  journal={arXiv preprint arXiv:2603.23186},
  year={2026}
}

@article{mdm,
  title={Human motion diffusion model},
  author={Tevet, Guy and Raab, Sigal and Gordon, Brian and Shafir, Yonatan and Cohen-Or, Daniel and Bermano, Amit H},
  journal={arXiv preprint arXiv:2209.14916},
  year={2022}
}

@article{ddpm,
  title={Denoising diffusion probabilistic models},
  author={Ho, Jonathan and Jain, Ajay and Abbeel, Pieter},
  journal={Advances in neural information processing systems},
  volume={33},
  pages={6840--6851},
  year={2020}
}

@article{rope,
  title={Roformer: Enhanced transformer with rotary position embedding},
  author={Su, Jianlin and Ahmed, Murtadha and Lu, Yu and Pan, Shengfeng and Bo, Wen and Liu, Yunfeng},
  journal={Neurocomputing},
  volume={568},
  pages={127063},
  year={2024},
  publisher={Elsevier}
}

@inproceedings{minsnr,
  title={Efficient diffusion training via min-snr weighting strategy},
  author={Hang, Tiankai and Gu, Shuyang and Li, Chen and Bao, Jianmin and Chen, Dong and Hu, Han and Geng, Xin and Guo, Baining},
  booktitle={Proceedings of the IEEE/CVF international conference on computer vision},
  pages={7441--7451},
  year={2023}
}

@article{ddim,
  title={Denoising diffusion implicit models},
  author={Song, Jiaming and Meng, Chenlin and Ermon, Stefano},
  journal={arXiv preprint arXiv:2010.02502},
  year={2020}
}

@misc{gpt4o,
  title        = {Hello GPT-4o},
  author       = {{OpenAI}},
  year         = {2024},
  howpublished = {\url{https://openai.com/index/hello-gpt-4o/}},
  note         = {Accessed: 2026-04-22}
}

@inproceedings{g2pddm,
  title={G2p-ddm: Generating sign pose sequence from gloss sequence with discrete diffusion model},
  author={Xie, Pan and Zhang, Qipeng and Taiying, Peng and Tang, Hao and Du, Yao and Li, Zexian},
  booktitle={Proceedings of the AAAI Conference on Artificial Intelligence},
  volume={38},
  number={6},
  pages={6234--6242},
  year={2024}
}

@article{mar,
  title={Autoregressive image generation without vector quantization},
  author={Li, Tianhong and Tian, Yonglong and Li, He and Deng, Mingyang and He, Kaiming},
  journal={Advances in Neural Information Processing Systems},
  volume={37},
  pages={56424--56445},
  year={2024}
}

@article{rag,
  title={Retrieval-augmented generation for knowledge-intensive nlp tasks},
  author={Lewis, Patrick and Perez, Ethan and Piktus, Aleksandra and Petroni, Fabio and Karpukhin, Vladimir and Goyal, Naman and K{\"u}ttler, Heinrich and Lewis, Mike and Yih, Wen-tau and Rockt{\"a}schel, Tim and others},
  journal={Advances in neural information processing systems},
  volume={33},
  pages={9459--9474},
  year={2020}
}

@inproceedings{flowmatching,
  title={Flow Matching for Generative Modeling},
  author={Lipman, Yaron and Chen, Ricky TQ and Ben-Hamu, Heli and Nickel, Maximilian and Le, Matthew},
  booktitle={The Eleventh International Conference on Learning Representations}
}

@inproceedings{slrtp_challenge,
  title={SLRTP2025 Sign Language Production Challenge: Methodology, Results and Future Work},
  author={Walsh, Harry and Fish, Ed and Sincan, Ozge Mercanoglu and Lakhal, Mohamed Ilyes and Bowden, Richard and Fox, Neil and Woll, Bencie and Wu, Kepeng and Li, Zecheng and Zhao, Weichao and others},
  booktitle={Proceedings of the IEEE/CVF Conference on Computer Vision and Pattern Recognition},
  pages={4148--4158},
  year={2025}
}

@article{unisign,
  title={Uni-sign: Toward unified sign language understanding at scale},
  author={Li, Zecheng and Zhou, Wengang and Zhao, Weichao and Wu, Kepeng and Hu, Hezhen and Li, Houqiang},
  journal={arXiv preprint arXiv:2501.15187},
  year={2025}
}

@inproceedings{bsl,
  title={BSL-1K: Scaling up co-articulated sign language recognition using mouthing cues},
  author={Albanie, Samuel and Varol, G{\"u}l and Momeni, Liliane and Afouras, Triantafyllos and Chung, Joon Son and Fox, Neil and Zisserman, Andrew},
  booktitle={European conference on computer vision},
  pages={35--53},
  year={2020},
  organization={Springer}
}

@inproceedings{recognition_coarticulation,
  title={Read and attend: Temporal localisation in sign language videos},
  author={Varol, Gul and Momeni, Liliane and Albanie, Samuel and Afouras, Triantafyllos and Zisserman, Andrew},
  booktitle={Proceedings of the IEEE/CVF Conference on Computer Vision and Pattern Recognition},
  pages={16857--16866},
  year={2021}
}

@inproceedings{recognition_coarticulation2,
  title={Automatic dense annotation of large-vocabulary sign language videos},
  author={Momeni, Liliane and Bull, Hannah and Prajwal, KR and Albanie, Samuel and Varol, G{\"u}l and Zisserman, Andrew},
  booktitle={European Conference on Computer Vision},
  pages={671--690},
  year={2022},
  organization={Springer}
}

@software{spacy,
  author = {Honnibal, Matthew and Montani, Ines and Van Landeghem, Sofie and Boyd, Adriane},
  title = {{spaCy}: Industrial-strength Natural Language Processing in Python},
  year = {2020},
  publisher = {Zenodo},
  doi = {10.5281/zenodo.1212303},
  url = {https://doi.org/10.5281/zenodo.1212303}
}

@article{quantile_loss,
  title={Regression quantiles},
  author={Koenker, Roger and Bassett Jr, Gilbert},
  journal={Econometrica: journal of the Econometric Society},
  pages={33--50},
  year={1978},
  publisher={JSTOR}
}

@inproceedings{slt_ctc,
  title={Sign language transformers: Joint end-to-end sign language recognition and translation},
  author={Camgoz, Necati Cihan and Koller, Oscar and Hadfield, Simon and Bowden, Richard},
  booktitle={Proceedings of the IEEE/CVF conference on computer vision and pattern recognition},
  pages={10023--10033},
  year={2020}
}

@article{conditional_flowmatching,
  title={Improving and generalizing flow-based generative models with minibatch optimal transport},
  author={Tong, Alexander and Fatras, Kilian and Malkin, Nikolay and Huguet, Guillaume and Zhang, Yanlei and Rector-Brooks, Jarrid and Wolf, Guy and Bengio, Yoshua},
  journal={arXiv preprint arXiv:2302.00482},
  year={2023}
}

@inproceedings{ctcloss,
  title={Connectionist temporal classification: labelling unsegmented sequence data with recurrent neural networks},
  author={Graves, Alex and Fern{\'a}ndez, Santiago and Gomez, Faustino and Schmidhuber, J{\"u}rgen},
  booktitle={Proceedings of the 23rd international conference on Machine learning},
  pages={369--376},
  year={2006}
}

@article{fgd,
  title={Speech gesture generation from the trimodal context of text, audio, and speaker identity},
  author={Yoon, Youngwoo and Cha, Bok and Lee, Joo-Haeng and Jang, Minsu and Lee, Jaeyeon and Kim, Jaehong and Lee, Geehyuk},
  journal={ACM Transactions on Graphics (TOG)},
  volume={39},
  number={6},
  pages={1--16},
  year={2020},
  publisher={ACM New York, NY, USA}
}

@inproceedings{remos,
  title={Remos: 3d motion-conditioned reaction synthesis for two-person interactions},
  author={Ghosh, Anindita and Dabral, Rishabh and Golyanik, Vladislav and Theobalt, Christian and Slusallek, Philipp},
  booktitle={European conference on computer vision},
  pages={418--437},
  year={2024},
  organization={Springer}
}
